\documentclass[acmsmall]{acmart}
\usepackage[T1]{fontenc}
\usepackage{amsmath}
\usepackage{graphicx}
\usepackage{amsmath}
\usepackage{comment}
\usepackage[svgnames]{xcolor}

\definecolor{AmirBlue}{HTML}{002EAF}

\definecolor{arman}{HTML}{4bb3a1}

\definecolor{arash}{HTML}{a88d32}

\definecolor{tooba}{HTML}{007FFF}

\usepackage{booktabs}
\usepackage{xcolor}
\usepackage{pifont}
\usepackage{longtable}
\newcommand{\cmark}{\textcolor{green!60!black}{\ding{51}}}

\usepackage{breakcites}
\begin{document}
%
% \title{Human Cognition in Machines: A Survey of World Models}
\title{Human Cognition in Machines: A Unified Perspective of World Models}

% \author{Professor Wang's Group\inst{1,2}}
% \author{Timothy Rupprecht\inst{1} \and Pu Zhao\inst{1} \and Amir Taherin\inst{1} \and Arash Akbari\inst{1} \and Arman Akbari\inst{1} \and Rahul Chowdhury \and Enfu Nan\inst{1} \and Juyi  Lin\inst{1} \and Yumei He\inst{2} \and \\ Yanzhi Wang\inst{1}}

% \author{
% Timothy Rupprecht\inst{1,2,}*$^{,\dagger}$ \and
% Pu Zhao\inst{1,}* \and
% Amir Taherin\inst{1}* \and
% Arash Akbari\inst{1}* \and
% Arman Akbari\inst{1}* \and
% Yumei He\inst{3}* \and
% Tooba Imtiaz\inst{1}* \and
% Sean Duffy\inst{1} \and
% Juyi Lin\inst{1} \and
% Yixiao Chen\inst{1,2} \and
% Rahul Chowdhury\inst{1} \and
% Enfu Nan\inst{1} \and
% Yixin Shen\inst{1,5} \and
% Yifan Cao\inst{1,2} \and
% Haochen Zeng\inst{2} \and
% Chen Wang\inst{2} \and
% Weiwei Chen\inst{2} \and
% Geng Yuan\inst{1,4} \and
% Jennifer Dy\inst{1} \and
% Sarah Ostadabbas\inst{1} \and
% Xuan Zhang\inst{1} \and
% David Kaeli\inst{1} \and
% Edmund Yeh\inst{1} \and \\
% Yanzhi Wang\inst{1,2,}$^{\dagger}$}

\author{Timothy Rupprecht}
\authornote{These authors contributed equally to this research.}
\authornote{These authors are corresponding authors for this research.}
\affiliation{%
  \institution{Northeastern University}
  \city{Boston}
  \state{MA}
  \country{USA}
}
\affiliation{%
  \institution{EmbodyX Inc.}
  \city{San Mateo}
  \state{CA}
  \country{USA}
}
\author{Pu Zhao}
\authornotemark[1]
\affiliation{%
  \institution{Northeastern University}
  \city{Boston}
  \state{MA}
  \country{USA}
}

\author{Amir Taherin}
\authornotemark[1]
\affiliation{%
  \institution{Northeastern University}
  \city{Boston}
  \state{MA}
  \country{USA}
}

\author{Arash Akbari}
\authornotemark[1]
\affiliation{%
  \institution{Northeastern University}
  \city{Boston}
  \state{MA}
  \country{USA}
}

\author{Arman Akbari}
\authornotemark[1]
\affiliation{%
  \institution{Northeastern University}
  \city{Boston}
  \state{MA}
  \country{USA}
}

\author{Yumei He}
\authornotemark[1]
\affiliation{%
  \institution{Tulane University}
  \city{New Orleans}
  \state{LA}
  \country{USA}
}

\author{Tooba Imtiaz}
\authornotemark[1]
\affiliation{%
  \institution{Northeastern University}
  \city{Boston}
  \state{MA}
  \country{USA}
}

\author{Sean Duffy}
\affiliation{%
  \institution{Northeastern University}
  \city{Boston}
  \state{MA}
  \country{USA}
}

\author{Juyi Lin}
\affiliation{%
  \institution{Northeastern University}
  \city{Boston}
  \state{MA}
  \country{USA}
}
\author{Yixiao Chen}
\affiliation{%
  \institution{Northeastern University}
  \city{Boston}
  \state{MA}
  \country{USA}
}
\affiliation{%
  \institution{EmbodyX Inc.}
  \city{San Mateo}
  \state{CA}
  \country{USA}
}

\author{Rahul Chowdhury}
\affiliation{%
  \institution{Northeastern University}
  \city{Boston}
  \state{MA}
  \country{USA}
}
\author{Enfu Nan}
\affiliation{%
  \institution{Northeastern University}
  \city{Boston}
  \state{MA}
  \country{USA}
}
\author{Yixin Shen}
\affiliation{%
  \institution{Northeastern University}
  \city{Boston}
  \state{MA}
  \country{USA}
}
\affiliation{%
  \institution{Cornell University}
  \city{Ithaca}
  \state{NY}
  \country{USA}
}
\author{Yifan Cao}
\affiliation{%
  \institution{Northeastern University}
  \city{Boston}
  \state{MA}
  \country{USA}
}
\affiliation{%
  \institution{EmbodyX Inc.}
  \city{San Mateo}
  \state{CA}
  \country{USA}
}
\author{Haochen Zeng}
\affiliation{%
  \institution{EmbodyX Inc.}
  \city{San Mateo}
  \state{CA}
  \country{USA}
}
\author{Chen Wang}
\affiliation{%
  \institution{EmbodyX Inc.}
  \city{San Mateo}
  \state{CA}
  \country{USA}
}
\author{Weiwei Chen}
\affiliation{%
  \institution{EmbodyX Inc.}
  \city{San Mateo}
  \state{CA}
  \country{USA}
}

\author{Geng Yuan}
\affiliation{%
  \institution{Northeastern University}
  \city{Boston}
  \state{MA}
  \country{USA}
}
\affiliation{%
  \institution{University of Georgia}
  \city{Athens}
  \state{GA}
  \country{USA}
}
\author{Jennifer Dy}
\affiliation{%
  \institution{Northeastern University}
  \city{Boston}
  \state{MA}
  \country{USA}
}
\author{Sarah Ostadabbas}
\affiliation{%
  \institution{Northeastern University}
  \city{Boston}
  \state{MA}
  \country{USA}
}
\author{Xuan Zhang}
\affiliation{%
  \institution{Northeastern University}
  \city{Boston}
  \state{MA}
  \country{USA}
}

\author{David Kaeli}
\affiliation{%
  \institution{Northeastern University}
  \city{Boston}
  \state{MA}
  \country{USA}
}

\author{Edmund Yeh}
\affiliation{%
  \institution{Northeastern University}
  \city{Boston}
  \state{MA}
  \country{USA}
}

\author{Yanzhi Wang}
\authornotemark[2]
\affiliation{%
  \institution{Northeastern University}
  \city{Boston}
  \state{MA}
  \country{USA}
}

%
%\titlerunning{Abbreviated paper title}
% If the paper title is too long for the running head, you can set
% an abbreviated paper title here
%
% \author{First Author\inst{1}\orcidID{0000-1111-2222-3333} \and
% Second Author\inst{2,3}\orcidID{1111-2222-3333-4444} \and
% Third Author\inst{3}\orcidID{2222--3333-4444-5555}}
%
% \authorrunning{F. Author et al.}
% First names are abbreviated in the running head.
% If there are more than two authors, 'et al.' is used.
%
% \institute{Northeastern University, Boston MA 02445, USA \and 
% EmbodyX Inc., San Mateo, CA 94002, USA \and  Tulane University, New Orleans, LA 70118, USA \and University of Georgia, Athens, GA 30602, USA \and Cornell University, Ithaca, NY 14850, USA \\[1.0em]
% {\normalsize *These authors equally contributed}\\
% {\normalsize $^{\dagger}$These authors are corresponding authors}\\[1.0em]
% \textit{A Report from the Physical AI Research (PAIR) Center at Northeastern University}
% }

% \email{lncs@springer.com}\\
% \url{http://www.springer.com/gp/computer-science/lncs} 
% \and
% ABC Institute, Rupert-Karls-University Heidelberg, Heidelberg, Germany\\
% \email{\{abc,lncs\}@uni-heidelberg.de}
%}
%
% \maketitle              % typeset the header of the contribution
%
\begin{abstract}
This report of world models distinguishes prior works by the cognitive functions they innovate. Many works claim an almost \textit{human}-like cognitive capability in their world models. 
To evaluate these claims requires a proper grounding in first principles from human and machine cognition theory. 
In moving towards \textit{human}-like world models we present a conceptual unified framework for world models that fully incorporates all the cognitive functions (i.e., memory, perception, language, reasoning, imagining, motivation, and metacognition) and identify gaps in existing research as a guide for future states of the art. 
In particular, we find that motivation (especially intrinsic motivation) and metacognition remain drastically under-researched, and we propose concrete directions to address these gaps informed by active inference and global workspace theory. 
We also introduce epistemic world models, a new category encompassing agent frameworks for scientific discovery that operate over structured knowledge. 
Our taxonomy, applied to video, embodied, and epistemic world models, suggests research directions where prior taxonomies have not.

% This comprehensive report of world models distinguishes prior works by the cognitive function they innovate.
% Many works claim an almost ``human-like'' cognitive capability in their world models.
% To evaluate these claims requires a proper grounding in first principles in Cognition Architecture Theory (CAT).
% We synthesize our survey into a conceptual unified framework for world models that fully incorporates all the cognitive functions associated with CAT and identify gaps in the research as a guide for future states of the art.
% These unified world models holistically incorporate the component parts of cognition: memory, perception, language, reasoning, imagining, motivation, and metacognition.
% Furthermore, our taxonomy rooted in CAT reveals solutions to the revealed research gaps where other taxonomies have failed to instigate innovation.
% \keywords{World Models \and  \and  \and }
\keywordsname{: World Models, Vision Language Action Models, World Action Models, Machine Cognition}

\end{abstract}

\maketitle              % typeset the header of the contribution
\renewcommand{\shortauthors}{Rupprecht et al.}

% here
\section{Introduction}

%  Kenneth Craik was first to describe world models used in human cognition as the mental models that represent an external reality capable of predicting the future~\cite{craik1967nature}. 
% \textcolor{red}{
World models have become a central abstraction for building intelligent machines. At their core, world models learn internal representations of an external environment and use them to predict how that environment may evolve over time~\cite{craik1967nature}.
Contemporary world models learn an environment's spatial and temporal characteristics for representation and generation~\cite{ha2018world}. 
More recently, the definition has expanded to include a model's capability to predict how an environment will evolve under counterfactual actions~\cite{lecun2022path}. 
% \textcolor{red}{
Across video generation, embodied agents, and language-based reasoning systems, world models are increasingly expected not only to represent the world, but also to support imagination, decision-making, and adaptive behavior.
% }
 
% \textcolor{red}{
However, this rapid expansion has also created conceptual ambiguity. Many recent world models are described using cognitive or anthropomorphic language. They are said to understand scenes, imagine futures, reason about actions, remember past states, or behave in increasingly human-like ways~\cite{ibrahim2025thinking,xiao2025humanizing,kim2025humanoid,BeingH,Kim2025towards}. Such descriptions are useful when they point to concrete computational capabilities, but they can also obscure what a model actually does. As a result, the field lacks a shared vocabulary for distinguishing which cognitive functions are present, which are absent, and which remain aspirational for world models.
% }

% \textcolor{red}{
This is why a human cognition perspective is necessary. When world models are evaluated, compared, or advertised in cognitive terms, the comparison should be grounded in a principled account of cognition. Cognitive Architecture Theory (CAT), especially the functional decomposition proposed by Newell, provides a foundation by identifying key components of cognition, including memory, perception, language, reasoning, imagination, motivation, and metacognition~\cite{newell1994unified}. This perspective allows us to move beyond broad claims of human-like intelligence and ask more precise questions: What cognitive functions does a given world model implement? How are these functions represented computationally? Which functions are underdeveloped in current systems? And how might progress in one domain inform another?
% }

%\textcolor{blue}{Is it a desirable trait for a world model's cognition to be \textit{human}-like?
%What does it mean for a world model's cognition to be \textit{human}-like?
%We do not make a prescription regarding the first question as it is context dependent.
%There are examples where \textit{human}-like cognitive functionality benefits world models (e.g. enabling hierarchical or recursive problem solving), and there are examples where \textit{unhuman}-like cognitive functionality benefits world models (e.g. training simulations).
%We set out to answer the second question by defining what a \textit{human}-like cognitive framework would look like within contemporaneous world model research.
%We offer our perspective on a unified framework for world models that incorporates all cognitive functions as a conceptual road map for future research.} 

% \textcolor{red}{
In this survey, we use human cognition as an organizing lens for understanding contemporary world models. Rather than treating world models as a single technical category, we analyze them according to the cognitive functions they emulate, approximate, or neglect. 
% }
%Our report is among the first to categorize contemporaneous world model research by the cognitive functions they emulate and innovate as identified by cognitive scientist Allen Newell (memory, perception, language, reasoning, imagining, motivation, and metacognition)~\cite{newell1994unified}.
From this synthesis, we identify two critically under-researched cognitive components: motivation and metacognition. 
% \textcolor{red}{
Current state-of-the-art world models depend on externally specified reward functions, hand-designed objectives, or task-specific prompts, rather than intrinsic mechanisms for curiosity, uncertainty reduction, or self-directed goal formation. Likewise, few of them possess metacognitive mechanisms that monitor their own uncertainty, evaluate the reliability of their internal predictions, or regulate reasoning and action based on self-assessment. These missing components limit the extent to which current world models can support autonomous, adaptive, and robust intelligence.
% }
%Current state-of-the-art world models lack intrinsic motivation mechanisms beyond hand-crafted reward signals, and none demonstrate genuine metacognitive capabilities such as monitoring and control on the top of those functionalities. Addressing these gaps is essential if the field's aspiration toward \textit{human}-like world models is to move beyond rhetoric.

Our report spans three domains of contemporary world model research, including 1) video world models (Sec.~\ref{sec:video_wm}) generate future visual states conditioned on observations and actions, where maintaining spatial consistency and long-horizon temporal coherence remain central challenges. 
2) embodied world models (Sec.~\ref{sec:embodied_wm}) extend these demands to physical settings, requiring perception of contact geometry, memory of persistent environments, and reasoning over force propagation to guide real-world task execution. 
Beyond these established domains, we propose a new category we call {\bf epistemic world models} (Sec.~\ref{sec:collab}), in which the environment is not a physical scene but a structured knowledge space interacted with by an agent framework.
Epistemic world models are contrasted with what we refer to as latent world models that learn spatial and temporal dynamics over a learned latent state space, comprising most prior world model works.
In the epistemic setting, agents with an VLM or LLM backbone are already world models themselves~\cite{ge2024worldgpt,gu2024your}, but when combined with an agent harness and a human-in-the-loop to provide additional reasoning and motivation, an agent updates its world state within a global workspace defined through easy-to-interpret language. 
Epistemic world models also provide early instantiations of the metacognitive mechanisms that latent world models currently lack, making them both a distinct research domain and a source of solutions for the gaps our taxonomy reveals.

The contributions of our report are as follows:
\begin{enumerate}
    \item We are among the first to provide a comprehensive review of contemporary world models grounded in human-machine cognition.
    \item We propose a unified world model as a conceptual road-map for incorporating all the component parts of cognitive architecture for robust world representation and generation.
    \item We identify motivation and metacognition as critical but underexplored components of current world models and discuss research directions.
    \item We introduce epistemic world models as a new category of world model in which agents represent, update, and reason over structured knowledge spaces.
\end{enumerate}

\section{Background}\label{sec:bg}
\begin{figure*}[t!]
    \centering
    \includegraphics[width=\linewidth]{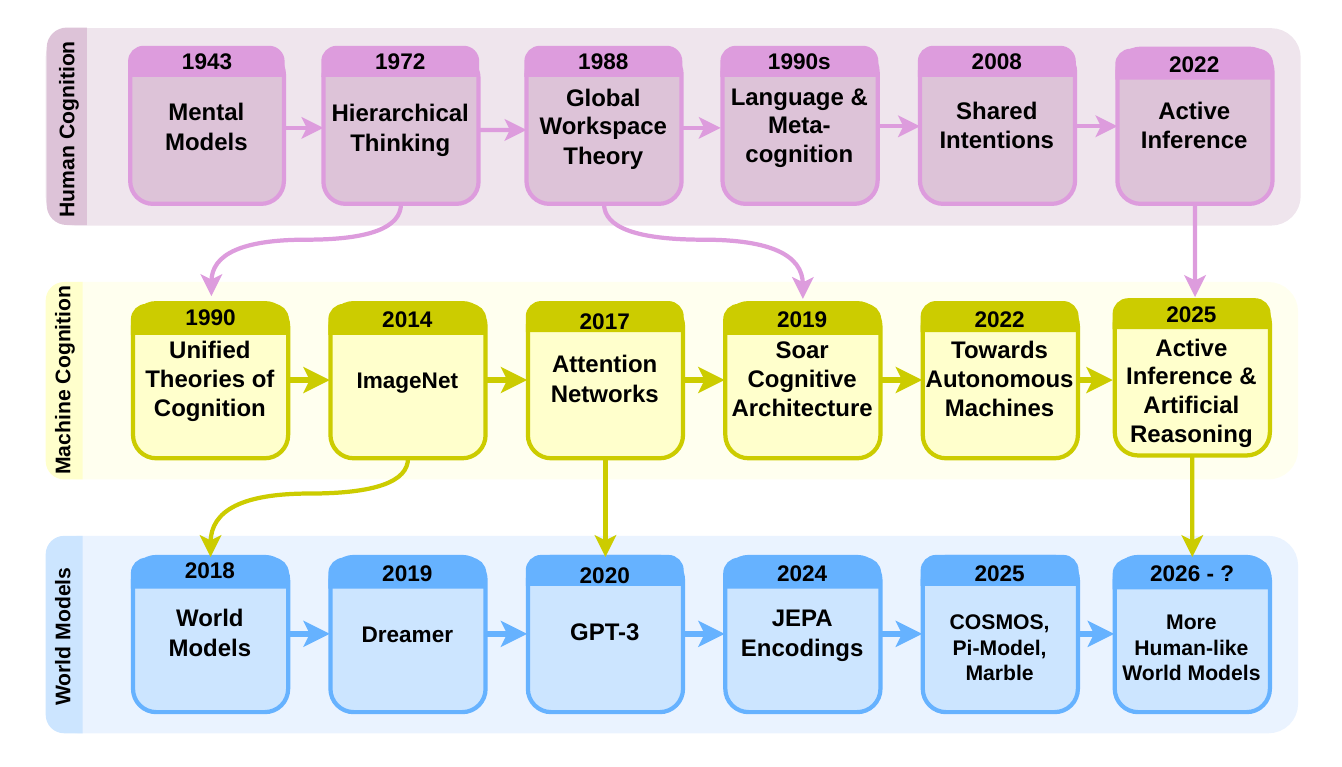}
    % \vspace{-0.1cm}
    % \hrule
    % \vspace{-0.1cm}
    \caption{Our survey studies the convergence of three different but inter-related fields: human cognition, machine cognition, and world models.}
    \label{fig:history}
\end{figure*}
% \textcolor{red}{(Add more citations in this section, for example,  the following 1) and 2) categories. )}

Our report spans these three interrelated research tracks as summarized in Figure~\ref{fig:history}.
Previous world model surveys create a coarse dichotomy in their taxonomies.
They classify world models as 1) world representations~\cite{hansen2024td,assran2023self,ge2024worldgpt,gao2025adaworld,bheemaiah2025knowledge}, and 2) world generators~\cite{huang2024safedreamer,wang2025adawm,jiang2025irl,wu2026pragmatic,xiang2025pan}.
In Figure~\ref{fig:taxonomy} our finer dichotomy of world models is shown with exemplary works that innovate on their primary cognitive function. 
Our taxonomy draws on lessons from the fields of human and machine cognition that we will review in this section.
While our taxonomy is not explicit in prior work, it emerges naturally when aligning model capabilities with longstanding~\cite{newell1994unified,laird2019soar,anderson1997act,meyer2001executive,langley1991design} and recent~\cite{friston2016active,ritter2019act,pezzulo2024active} cognitive architectures.
We discuss the research tracks from Figure~\ref{fig:history} now, first by reviewing world model research in Sec.~\ref{sec:bg:wm}, then by reviewing human and machine cognition in Sec.~\ref{sec:bg:cat}.

\subsection{World Models}\label{sec:bg:wm}
The seminal work on \textit{contemporary} world Models is from Ha et al. (2018)~\cite{ha2018world} proposing a framework enabling dreamer architectures~\cite{ha2018world}. 
However, world models have been in use long before this framework solidified with antecedents in control theory~\cite{bryson2018applied}, machine cognition~\cite{newell1995gps,langley1991design,newell1994unified,anderson1997act}, and early neural network theory~\cite{nguyen1990neural}.
Early world models were trained with Reinforcement Learning (RL) to learn action policies for direct robotic motor control~\cite{ha2018world,wu2023daydreamer}. 
The scope has since expanded to include multi-modal video world models capable of generating vibrant visuals~\cite{marbleworldblog2026}, embodied WMs for mapping and locomotion~\cite{intelligence2025pi_,huang2026pointworld}, and as we will argue, should also expand to world models used within agent frameworks~\cite{gottweis2025towards,gottweis2026accelerating,OpenAI_Prism_2026,shao2025omnisci}. 

A review of recent surveys of state-of-the-art World Models~\cite{ding2025understanding,li2025comprehensive,yue2025simulating,long2025survey,lin2025exploring,liu2025generative,xu2026specialist,dong2026learning} shows that in order to support planning and decision-making, especially in embodied settings~\cite{long2025survey,li2025comprehensive,hou2026world}, world models consistently function as simulators that 1) represent current world structure and 2) predict future world dynamics~\cite{ding2025understanding}.
Recent works also survey advances in video world models~\cite{yue2025simulating}, embodiment~\cite{li2025comprehensive}, temporal–spatial modeling~\cite{liu2025generative}, and physical realism~\cite{lin2025exploring}, all highlighting challenges in long-horizon consistency, computational efficiency, and alignment to real-world physics.
The available surveys also focus on domain application~\cite{long2025survey,yue2025simulating,li2025comprehensive,lin2025exploring,dong2026learning}, architectural or input modality distinctions~\cite{liu2025generative}, or abstract taxonomies~\cite{ding2025understanding,xu2026specialist}.
For a benchmark of video world models aligned with physical laws see recent work PhyGround~\cite{lin2026phyground} with its public leader board~\cite{lin2026phyground_lb}.

As demonstrated in Table~\ref{tab:survey_comparison}, we are among the first to systematically distinguish recent states of the art by the cognitive functions they primarily innovate (further discussed in Sec.~\ref{sec:bg:taxonomy}).
Surveys near-universally discuss the foundational models and works from state-of-the-art teams at META AI with Yann LeCun~\cite{hansen2025hierarchical,garrido2026learning,yuan2026inference}, the Alibaba group~\cite{wan2025wan}, Cosmos from Nvidia~\cite{kim2026cosmos}, Berkeley University's  Sergey Levine's group~\cite{intelligence2025pi_} and Stanford University's Fei-Fei Li's group~\cite{yang2025cambrian,wang2026vagen,huang2026pointworld}.
Additional states-of-the-art innovate on architecture by using Vision-Action~\cite{li2026causal} and Vision-Action-Language models~\cite{team2026advancing}, auto-regression models~\cite{gao2025adaworld}, and diffusion models~\cite{zhang2025epona,li2025dawm}.
Nevertheless, a framework for a Unified World Model remains elusive to researchers. 
A divide remains in the designs for general-purpose and domain-specific world models~\cite{xu2026specialist}.

\begin{table*}[]
\centering
\scriptsize
\noindent\makebox[\textwidth]{%
\setlength{\tabcolsep}{3.0pt}
\renewcommand{\arraystretch}{1.05}
\begin{tabular}{l l c c c c c c p{4.5cm}}
\toprule
 \textbf{Survey} & \textbf{Year} & \rotatebox{0}{\textbf{Video}} & \rotatebox{0}{\textbf{Embodied}} & \rotatebox{0}{\textbf{Simulation}} & \rotatebox{0}{\textbf{Phys. Align.}} & \rotatebox{0}{\textbf{Epistemic}} & \rotatebox{0}{\textbf{CAT}} & \textbf{Primary Distinction} \\
\midrule
Ding et al.~\cite{ding2025understanding} & 2025 & \cmark & \cmark &  &  &  &  & World understanding vs.\ prediction \\
Li et al.~\cite{li2025comprehensive} & 2025 &  & \cmark & \cmark &  &  &  & Embodied AI and simulators \\
Yue et al.~\cite{yue2025simulating} & 2025 & \cmark &  & \cmark &  &  &  & Roadmap for visual world simulation \\
Long et al.~\cite{long2025survey} & 2025 &  & \cmark & \cmark &  &  &  & Simulators for embodied intelligence \\
Lin et al.~\cite{lin2025exploring} & 2025 & \cmark &  &  & \cmark &  &  & Physics alignment in video gen. \\
Liu et al.~\cite{liu2025generative} & 2025 & \cmark & \cmark &  &  &  &  & Architectural and input distinctions \\
Xu et al.~\cite{xu2026specialist} & 2026 & \cmark & \cmark &  &  &  &  & Specialist-to-generalist progression \\
Dong et al.~\cite{dong2026learning} & 2026 & \cmark & \cmark &  &  &  &  & Learning to model the world in AI \\
\midrule
\textbf{Ours} & \textbf{2026} & \cmark & \cmark & \cmark & \cmark & \cmark & \cmark & Cognitive architecture theory \\
\bottomrule
\end{tabular}
}
\caption{%\textcolor{blue}{
Comparison of world model survey scope. A check (\cmark) indicates that the survey explicitly covers the corresponding domain. To the best of our knowledge, our survey is among the first to examine world models from a human-cognition perspective by relating recent advances to the cognitive functions they emulate or extend, using Cognitive Architecture Theory (CAT) as an organizing framework.}%}

\label{tab:survey_comparison}
\end{table*}

\subsection{Cognitive Architecture Theory}\label{sec:bg:cat}

The capabilities of world models regularly draw comparisons to the abilities of their human counterparts~\cite{Kim2025towards,xiao2025humanizing,wu2026visual}.
This contributes to a widely held public belief that LLMs, world models, and similar generative AI approach human-like reasoning capabilities, are conscious, or herald imminent artificial general intelligence~\cite{kang2025identifying,colombatto2024folk,SARIKAYA2025101021}.
Without proper grounding in first-principles thinking from Cognitive Architecture Theory, these comparisons are misleading and often overstate the capabilities of our non-human counter-parts.

% \textcolor{blue}{
Any study of cognition, human or otherwise, begins with the work of Kenneth Craik who was a pioneer of cognitive sciences, and amongst the first to describe the mental models humans create of the world. 
In his seminal work~\cite{craik1967nature}, he theorized that humans use these models of the world to predict future states of the world, much like the world models we discuss in this report.
Posthumously, his work compared humans to a servomechanism that performs discrete tasks from taking in sensory input, making a decision, and then following through on the decision~\cite{craik1947theory,craik1948theory}.
More recently, Grush argues this same sensorimotor loop in humans, identified by Craik, creates neural circuitry within the brain modeling the world and this can be emulated in motor control~\cite{grush2004emulation}.
Sensorimotor engagement with the world confirms or rejects model expectations of the sensory feedback to enhance and process sensory information of future events.
This thinking will be used in both human and machine cognitive architectures.%}

\subsubsection{Cognitive Architecture in Humans.} To say a world model achieves \textit{Human}-like capabilities in cognition requires a first-principle investigation of Cognitive Architecture Theory in humans.
Human cognition includes component parts such as motor skills, adaptive learning, perception, symbolic reasoning with language, and memory~\cite{craik1967nature,donald1993origins,cognition2000,grush2004emulation,jaynes2013origin}. 
Human cognition also includes metacognition, or consciousness, and refers to a human's ability to self-monitor on more complex tasks~\cite{baars1993cognitive,koriat2006metacognition,rosenthal2000consciousness}.
% \textcolor{blue}{
This simply put can be described as \textit{thinking about thinking}, \textit{reasoning about reasoning}. This metacognition in humans can best be visualized through Baar's theater analogy used to describe his Global Workspace Theory as an explanation for human consciousness.
In Global Workspace Theory, there exist component parts of cognition such as memory, language, and sensory processing, and these correspond to the actors, and props on stage with a director offstage giving feedback during rehearsal. 
A spotlight moves from component to component as a human's conscious thinking invokes memories and sensory processing one at a time.
Despite the spotlight's illusion, all the component parts of the theater remain present, and interact even when the spotlight moves on to focus attention on something else on the stage.%}

Language as symbolic reasoning is relevant to our understanding of world models because both these and humans share the input and output modality of textual language.
Early and modern experts on the development of human language have argued that language is a tool in response to external factors like natural selection~\cite{darwin1872descent,pinker1990natural}, and is used to influence the external world~\cite{buhler1934sprachtheorie,lieberman2006toward}.
An emerging view argues that language evolved from humans’ ability to share intentions (i.e. humans' ability to share goals, to share attention or to share common ground)~\cite{tomasello2005understanding}.
Human language exists in reference to a world model shared between a party of humans.

From a review of first principles regarding how both human cognition and consciousness have evolved in humans, we assert the role of language is paramount.
We look to the works of evolutionary biologist Terrance Deacon~\cite{deacon1998symbolic} and psychologists Julian Jaynes~\cite{jaynes2013origin} and Merlin Donald~\cite{donald1993origins}. 
Deacon succinctly asserts that in humans ``symbolic thought does not come innately built in, but develops by internalizing the symbolic process that underlies language.''
Similarly, Jaynes asserts that ``consciousness becomes embedded in language.''

\subsubsection{Cognitive Architecture in Machines.} 
Machine cognition can be understood as a functional analogue of human cognition~\cite{craik1947theory,craik1948theory,newell1994unified}.
There have been a variety of machine cognitive architectures proposed over the years~\cite{langley1991design,anderson1997act,meyer2001executive,ritter2019act,laird2019soar,friston2025active}. 
ICARUS utilizes hierarchical skills and concepts~\cite{langley1991design,langley2006unified}. ACT-R uses Bayesian-style activation in an Adaptive Control of Thought—Rational loop~\cite{anderson1997act,ritter2019act}. 
EPIC leverages perception–action timing, making it well-suited for human-computer interaction in embodied settings~\cite{meyer2001executive}. 
Soar is a recursive cognitive architecture that models intelligent behavior through a symbolic State–Operator–Result (or S-O-R) cycle, in which reasoning proceeds through the selection and application of operators to representations of the world~\cite{laird2019soar}. 
% \textcolor{blue}{
Across all these cognitive loops, an agent/world model in discrete steps observes the environment, encodes and stores state information, forms symbolic or semantic abstractions, reasons over possible transitions, simulates alternative futures, and selects actions according to goals or preferences.%}

Cognitive Scientist Allen Newell laid the groundwork for the Soar Cognitive Architecture by first asserting a unified list of cognitive components seen in both human and machine cognition.
We list Newell's component part of the function of cognition in Table~\ref{tab:cognition_human_wm}.
Laird synthesized Newell's taxonomy to implement with Newell the Soar machine cognitive architecture~\cite{laird2019soar}.
S-O-R was proposed as $R = f_{\mathrm{rules}}(S_t, O_t)$ where $R$ is $S_{t+1}$.
In our world model notation, this is respectively,
\begin{align}\label{eq:basic_wm}
    z_{t+1} =  W_{\theta}(z_t, a_t)
\end{align}
The Soar system is capable of being applied in worlds that require visual and symbolic representative  reasoning~\cite{boggs2025towards}, implying compatibility with our state-of-the-art world models.
Soar provides a unified framework for perception, decision-making, and learning via production rules operating over a shared working memory similar to the Global Workspace Theory proposed for human cognition~\cite{baars1993cognitive}. 
Soar's use in world models was validated empirically in recent work~\cite{yuan2025nl2gensym}.

% \textcolor{blue}{
Recently, Yann LeCun outlined a machine cognition framework for autonomous machines~\cite{lecun2022path}.
A key argument LeCun makes is that ``common sense'' within machines can emerge as a hierarchy of low level models to high level models allowing the machine to ``fill in the blanks'' from incomplete world observations.
This is exactly how JEPA operates, and is reminiscent of the recursive S-O-R loop from machine cognition framework Soar~\cite{laird2019soar} and ``levels of processing'' theory of Fergus Craik~\cite{craik1972levels}.%}

\subsection{Our Taxonomy}\label{sec:bg:taxonomy}

\begin{table*}[t]
\centering
\scriptsize
\setlength{\tabcolsep}{5pt}
\renewcommand{\arraystretch}{1.15}
\begin{tabular}{p{1.85cm} p{5cm} p{6cm}}
\toprule
\textbf{Function~\cite{newell1994unified}} & \textbf{Human \& Machine Cognition} & \textbf{World Model Analogue} \\
\midrule
\textbf{\textit{Perception}} 
& Senses the current world. 
& Encodes observations into a world state $z_t$. \\

\textbf{\textit{Memory}} 
& Saves past experience and context. 
& Stores past states $z_{0:t}$ of context, maps, or external memory. \\

\textbf{\textit{Language}} 
& Uses symbols to communicate abstract meaning. 
& Tokenizes language, goals, states, and reasoning. \\

\textbf{\textit{Reasoning}} 
& Problem solving and action selection with inference 
& Infers $z_{t+1}$ or $a_{t}$ from other cognitive components. \\

\textbf{\textit{Imagination}} 
& Simulates possible and/or hypothetical futures. 
& Rolls out generated, simulated, or hypothetical $z_{t+1}$ or $a_{t}$. \\

\textbf{\textit{Motivation}} 
% & Selects behavior according to reward signal. 
% & Chooses actions using extrinsic or intrinsic rewards. \\
 & Behavior from extrinsic or intrinsic rewards.
 & Criteria for selecting $z_{t+1}$ or $a_{t}$. \\
\hline
\textbf{\textit{Metacognition}} 
& Operates over own representations, predictions, and decisions. 
& Self-reflection, self-evaluation, and self-control of all other cognitive components \\
\bottomrule
\end{tabular}
\caption{We select Newell's~\cite{newell1994unified} component parts of human and machine cognition to ground our framework. 
Alternative architectures offer different component decompositions but Newell's are unified across both human and machine cognition with plausible world model analogues applicable to all our explored domains.
}
\label{tab:cognition_human_wm}
\end{table*}

We select Newell's component parts of Human and Machine cognition to ground our framework. 
His prominence in early~\cite{newell1995gps} and contemporaneous machine cognition~\cite{newell1994unified,laird2019soar} makes the decision defensible, but we admit other sets of component parts found throughout machine cognition would likely also be defensible~\cite{hollan2000distributed,friston2010free,laird2025proposal}.
We unpack the core cognitive capabilities discussed in Tab.~\ref{tab:cognition_human_wm} and use these component parts as a framework to taxonomize world model research.
We survey works since Ha et al. (2018) that are highly relevant, appearing in top conferences or surveys, or are found in repositories hosted on GitHub~\cite{huang2025awesomeworldmodels}.
Newell's original list of the functional parts of human and machine cognition were \textbf{\textit{perception}}, \textbf{\textit{memory}}, \textbf{\textit{language}}, \textbf{\textit{reasoning}}, \textbf{\textit{imagination}}, and \textbf{\textit{motivation}}.
% \textcolor{blue}{
Newell's assertion that \textbf{\textit{reasoning}} extends to action selection~\cite{newell1994unified}, allows us to consider direct policy networks like Vision-Language-Action models in Sec.~\ref{sec:embodied_wm} which rely on implicit world models to transition world states through action selection alone.%}

\begin{figure*}[t!]
    \centering
    \includegraphics[width=\linewidth]{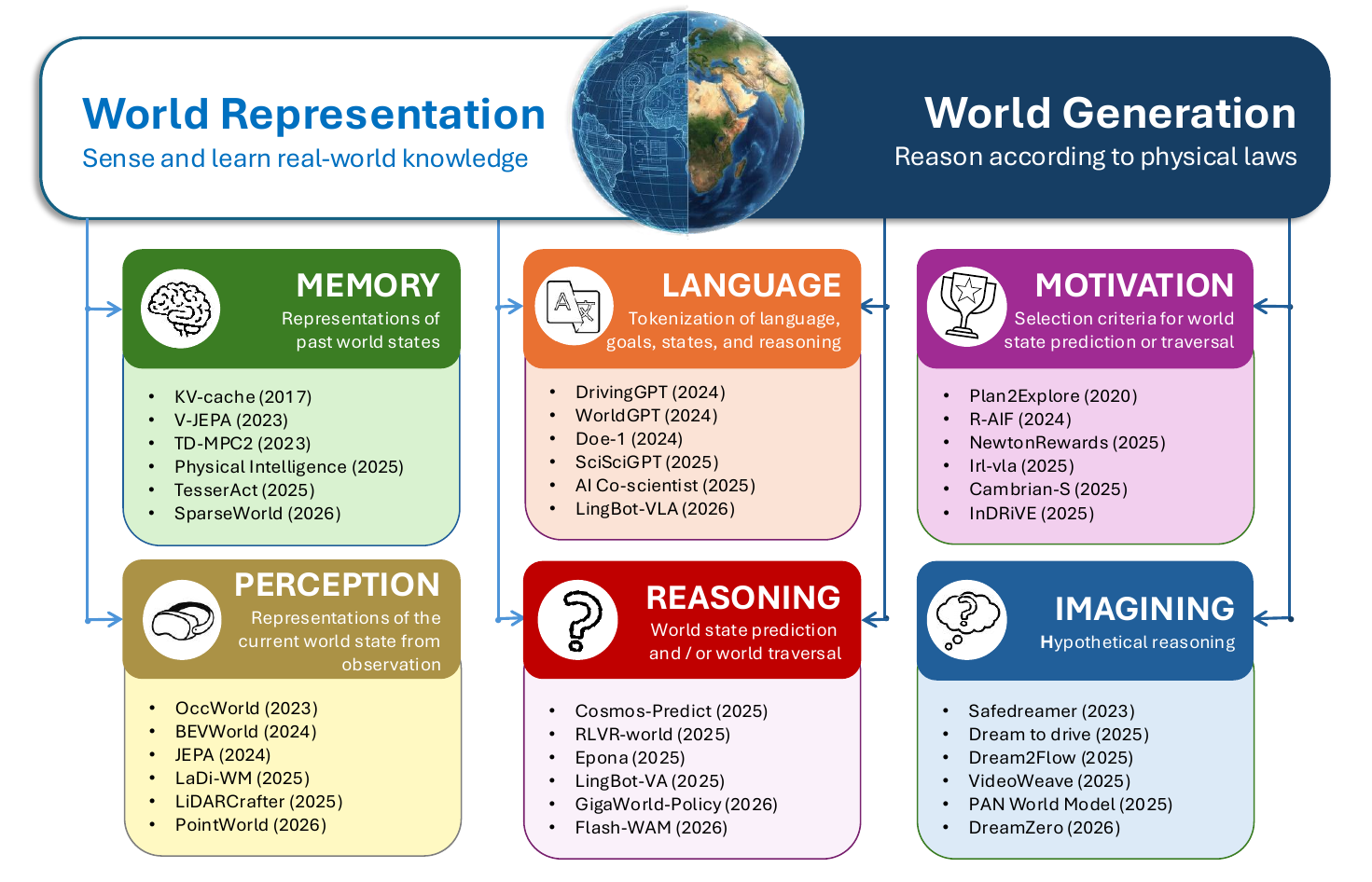}
    % \vspace{-0.1cm}
    % \hrule
    % \vspace{-0.1cm}
    \caption{The taxonomy of world models under cognitive architecture~\cite{newell1994unified}.}
    \label{fig:taxonomy}
\end{figure*}

% \textcolor{blue}{
Due to its paramount role in regulating cognitive functions, within our taxonomy's component parts list we include \textbf{\textit{metacognition}}, which refers to a world model's capacity to reflect on, evaluate, and control each component part of its cognitive processes.
%}  
\textbf{\textit{Metacognition}} is the recursive application of the other cognitive functions to themselves, such as, reasoning about one's own reasoning, monitoring one's own perception, evaluating one's own memory retrieval. In world models, this may be implemented through uncertainty estimation, self-evaluation, error detection, reflection, model selection, re-planning, or control mechanisms that decide when to revise representations, seek additional information, simulate alternatives, or defer action~\cite{baars1993cognitive,rosenthal2000consciousness,koriat2006metacognition}. 
 %Because it operates on the same functional categories Newell identified, rather than introducing a new category of its own, its inclusion completes the taxonomy rather than extending it.
  We can map these cognitive functions to the two traditional functions attributed to world models: world representation, and world generation.
  The first two items in Tab.~\ref{tab:cognition_human_wm} correspond to typical world model tasks in world representation, and the last two correspond to world prediction or generation.
  The middle two are found across both representation and generation tasks.
  This taxonomy is depicted in Figure~\ref{fig:taxonomy}.

% \textcolor{blue}{
In many previous world model works most of the components of cognition appear.
For example H-JEPA as an autonomous machine~\cite{lecun2022path} is proposed with capabilities to encode observations of the current world state (\textbf{\textit{perception}}), to learn this encoding with a latent predictive objective (\textbf{\textit{motivation}}), and stores queryable world states for later use (\textbf{\textit{memory}}). 
In follow-up works JEPA uses text tokens as an input (\textbf{\textit{language}}), and can be paired with an action policy network (\textbf{\textit{reasoning}})~\cite{le2025text,sun2026vla}.
However, this set of capabilities is not distinguishable from most other works when considering all applications of a work. 
Any taxonomy requires sorting decisions that are contestable at the margin. 
A review organized by application must decide whether a given robotics paper is 'manipulation,' 'locomotion,' or 'mapping' when its method plausibly serves all three.
In our case, rather than sorting works based on a cognitive function's presence alone, our sorting criterion is explicit and applies uniformly: we judge a work as innovative on a cognitive function when the work's stated contribution operates on that function; works claiming contributions on several are placed under several.
Then we sort our review by these innovations in world model cognition.
This makes our assignments inspectable and, where a reader disagrees, precisely locatable.

Using I-JEPA as an example~\cite{assran2023self}, the authors state three research contributions: 1) I-JEPA learns strong representations from input observations by predicting embeddings, not pixels, (we label a \textbf{\textit{perception}} and \textbf{\textit{motivation}} innovation) 2) by
using a simpler model with less rigid inductive bias,
I-JEPA is applicable to a wider set of tasks (we label a non-cognitive, evaluative consideration) and 3) JEPA is scalable and efficient (we label a non-cognitive, practical consideration).
Subsequently, we would taxonomize the I-JEPA work as innovating upon \textbf{\textit{perception}} and \textbf{\textit{motivation}}.
% In another example, augmenting JEPA's inputs to include camera-aligned LiDAR depth, a novel input for JEPA-based models, then we would consider that a \textbf{\textit{perception}} innovation~\cite{cornelissen2026mumo}.
We also present papers we deem exemplary of the cognitive innovations we survey in Figure~\ref{fig:taxonomy}.
Our comprehensive categorization of world model research can be found across Tables~\ref{tab:vwm_models},~\ref{tab:embodied_wm} and~\ref{tab:human_ai_collab}.
% }

Throughout our review, it is important to note that machine cognition functionally emulates, and does not mechanistically emulate, these component parts of human cognition. 
For example, when we say a world model has ``memory,'' we mean it maintains state representations across time steps, not that it has anything resembling episodic recall or the biological substrates of human memory. 
Machine cognition emulates human cognition in \textbf{\textit{memory}}, \textbf{\textit{perception}}, and some symbolic \textbf{\textit{reasoning}} tasks through \textbf{\textit{language}}, and \textbf{\textit{imagination}} with sim-to-real design paradigms.
It is ambiguous if machine cognition fully emulates true human-like \textbf{\textit{reasoning}} and machine cognition almost universally fails to emulate intrinsic \textbf{\textit{motivation}}. 
Machine intelligence has demonstrated little capability in emulating \textbf{\textit{metacognition}}.

\begin{figure*}[t!]
    \centering
    \includegraphics[width=\linewidth]{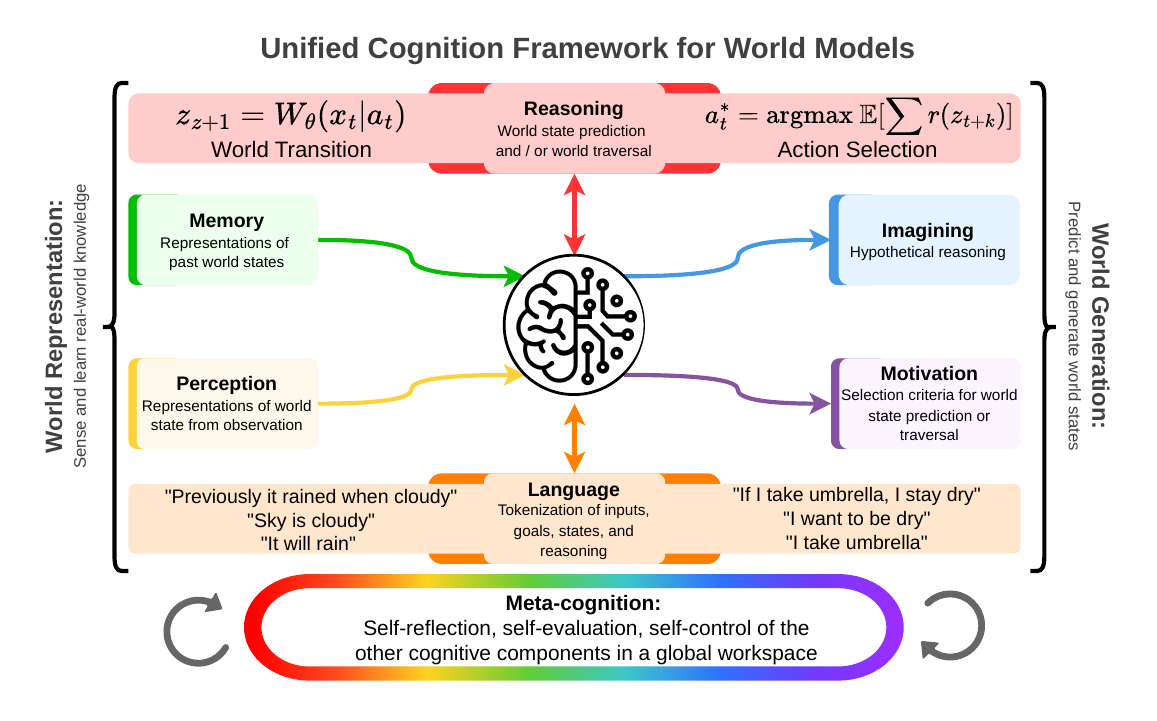}
    % \vspace{-1.5cm}
    % \hrule
    % \vspace{-0.1cm}
    \caption{Our Unified World Model derived from human-machine cognition~\cite{baars1993cognitive,newell1994unified}. This serves as a conceptual road-map for world model research. We also want to highlight that the component of \textit{language} mimics \textit{metacognition} in that it can be used to operate over each other component of cognition.}
    \label{fig:catbrain}
\end{figure*}

\section{Unified Cognition Framework for World Models} 
\label{sec:unified_wm}
% \textcolor{green}{Pu: maybe we need more figures and tables to summarize these papers in each section. These are all texts which is hard to read. We need more figures and tables to help readers  catch the key point in each section. } responded to, i think? - t.r.

We report on world models in the context of video world models (discussed further in Sec.~\ref{sec:video_wm}), embodied world models (discussed further in Sec.~\ref{sec:embodied_wm}), and what we define as epistemic world models, or world models used by agents for scientific discovery (discussed further in Sec.~\ref{sec:unified:cat} and~\ref{sec:collab}). 
We sub-categorize each section according to the component parts of machine and human cognition listed in Sec.~\ref{sec:bg:cat}.
We review exemplars for contemporary works that innovate on specific functions of cognition as shown in Figure~\ref{fig:taxonomy}.
We continue to taxonomize works across a more comprehensive body of research available for all to see in Tables~\ref{tab:vwm_models},~\ref{tab:embodied_wm}, and~\ref{tab:human_ai_collab}. 
A trend emerges across all three tables revealing a research gap regarding innovations targeting \textbf{\textit{metacognition}}.
Furthermore, upon closer inspection, innovations belonging to the \textbf{\textit{motivation}} column near always implement a form of \textit{extrinsic} motivation, external to the world model itself.
This differs from \textit{intrinsic} motivation as it is reliant on hand-crafted rewards, or actor-critic models trained with reinforcement learning.
As we stated in the introduction, if researchers want to claim that a world model has \textit{human}-like cognition, then we would expect to find all the components from Sec.~\ref{sec:bg:cat} of cognition unified within that world model framework.

We propose our conceptual unified world model framework that serves as a conceptual road map for this report and future world model research through the research gaps we identify.
% \textcolor{red}{(NOTE: added for Professor Wang's standard taxonomy)} 
World models sense and learn from real-world knowledge, predict and generate world states, reason and control according to physical laws implicitly, and can do all of this with or without a machine agent or a human-in-the-loop.
To accomplish this we propose a unified world model that holistically incorporates every component part of the CAT concepts discussed in Sec.~\ref{sec:bg:cat} into one conceptual unified framework as seen in Figure~\ref{fig:catbrain}.
% For the remaining sections of this work, it will serve as a conceptual road map for organizing other works.

As a design paradigm, our conceptual unified framework for world models calls for standardizing best practices in both world model representation, and world model generation.
% Our Unified Framework for world models includes,
% \begin{enumerate}
%     \item Multimodal perception  and~\ref{sec:worldrep:lang})
%     \item Efficient state representation with sufficiently long context windows (discussed further in Sec.~\ref{sec:worldrep:long})
%     \item Dreaming for hypothetical reasoning (discussed further in Sec.~\ref{sec:worldgen:imag})
%      \item Proper motivation during training and inference (discussed further in Sec.~\ref{sec:unified:wild})
%     \item Metacognitive capabilities (discussed further in Sec.~\ref{sec:unified:wild})
% \end{enumerate}
To fully span the functions of cognition emulated, Unified World Models encourage researchers to:
\begin{enumerate}
    % \item Use multi-modal inputs for \textbf{\textit{perception}} (discussed further in Sec.~\ref{sec:worldrep:multi}),
    \item Encode \textbf{\textit{perception}} of world representations from sensory observations with multi-modal state spaces (discussed further in Sec.~\ref{sec:worldrep:multi}),
    % \item Represent the world using latent state-spaces as \textbf{\textit{memory}} to enable downstream cognitive functions such as \textbf{\textit{imagination}}, \textbf{\textit{motivation}}, and \textbf{\textit{reasoning}} (discussed further in Sec.~\ref{sec:worldrep:long}),
    \item Leverage \textbf{\textit{memory}} for robust world models with state-spaces that encode multiple temporal scales, or store previous queryable states for efficient access (discussed further in Sec.~\ref{sec:worldrep:long}),
    \item Include \textbf{\textit{language}} tokens as an input, intermediate reasoning space, or output to facilitate human-in-the-loop cooperation (discussed further in Sec.~\ref{sec:worldrep:lang}),
    \item Enable \textbf{\textit{imagination}} as hypothetical reasoning during inference and sim-to-real transfer learning during training when data is scarce (discussed further in Sec.~\ref{sec:worldgen:imag}),
    \item Perform \textbf{\textit{reasoning}} with domain-specific models in Figs~\ref{fig:vwm_architecturesa} to~\ref{fig:collab_architectures} (discussed further in Sec.~\ref{sec:worldgen:reason}),
    \item Provide reward signals to world models for \textbf{\textit{motivation}} using state-based rewards that make salient robust measurements like active inference (discussed further in Sec.~\ref{sec:unified:cat}),
    \item Utilize \textbf{\textit{metacognition}} through global workspaces enabling self-reflection, self-evaluation, and self-control (discussed further in Sec.~\ref{sec:unified:cat}).
\end{enumerate}
None of these suggestions conflict with each other, but some are setting specific or conditional.
In Sec.~\ref{sec:unified:cat}, we will argue that our proposed research directions regarding \textbf{\textit{motivation}} and \textbf{\textit{metacognition}} fill a real research gap our taxonomy has revealed.

Reviewing video world models in Sec.~\ref{sec:video_wm} shows us the importance of creating world representations that enforce spatial consistency and meet memory constraints, using solutions like KV-cache, while still enforcing longer temporal consistency.
We show state-of-the-art video world model architectures in Figures~\ref{fig:vwm_architecturesa} to~\ref{fig:vwm_architecturesc}.
When we review embodied world models in Sec.~\ref{sec:embodied_wm} we see the importance of leveraging multi-modal inputs to make precise locomotion possible.
We also see in Sec.~\ref{sec:embodied_wm} that when training data is scarce, sim-to-real training can overcome this scarcity when world models learn latent representations that are traversable and remain consistent under domain shift to real-world applications (discussed more in Sec.~\ref{sec:worldrep}).
We show state-of-the-art embodied world model architectures in Figure~\ref{fig:emb_architecturesa} and~\ref{fig:emb_architecturesb}.
We propose in Sec.~\ref{sec:unified:cat} a new category of world models called \textit{epistemic world models} that we review in Sec.~\ref{sec:collab}.
% Epistemic world models represent structured knowledge and the scientific process itself, whereas latent world models learn state transition dynamics in video and embodied settings.
This domain includes agent frameworks for human-in-the-loop scientific discovery which serve as inspiration for overcoming one of the research gaps that we have observed in the latent world model's \textbf{\textit{metacognition}} capability.
We show state-of-the-art epistemic world model architectures in Figure~\ref{fig:collab_architectures}.

Latent world models learn state transition dynamics in video and embodied settings while agent frameworks are not typically framed that way.
However, epistemic world models in an agent framework with a Global Workspace representation of the world (defined in Sec.~\ref{sec:bg:cat}) perfectly aligns with Soar's State Operation Result loop~\cite{laird2019soar}, and we argue, even the latent world model definition as LLMs and VLMs are already considered to be world models~\cite{ge2024worldgpt,gu2024your}. 
A Global Workspace is not a learned latent encoding like JEPA~\cite{assran2023self}.
Instead, a Global Workspace is a \textbf{\textit{language}} based collection of all past prompts and responses including tool-call outputs.
% In the epistemic world model setting, the initial world is a static world representing structured knowledge, but as the agent acts by augmenting its context through multimodal RAG queries, performing analysis of existing literature, and encoding tool inputs and outputs into a global workspace (i.e., a chat history), creates a changing world, or in other words, a dynamic state space, meeting the traditional definition of the world model. 
In the epistemic world model setting, the initial world is static and represents structured knowledge from prompts, and context supplied by the user. 
However, as the agent interacts with this world—through multimodal Retrieval-Augmented Generation (RAG) queries, analyzing existing literature, and encoding inputs and outputs of tool-calls into a global workspace (e.g., a chat history)—it creates a changing environment. 
This transformation leads to a dynamic state space that aligns with the traditional definition of a world model.
Working within this agentic framework with a global workspace implementation allows models like Gemini Co-scientist~\cite{gottweis2025towards,gottweis2026accelerating}, an LLM and world model itself, to propose novel trajectories through an evolving state space to perform actual scientific discovery.

\subsection{World Representation}\label{sec:worldrep}
World representation refers to how spatio-temporal world knowledge is sensed and encoded ideally enabling downstream world model tasks such as world prediction over time and action selection within the world. 
World models use spatio-temporal world representations aligning with physical laws (discussed in both Sec.~\ref{sec4:generation} and~\ref{sec:embodied:world-gen}), and can use language representations to enable self-reflection, self-evaluation, and self-control (discussed further in Sec.~\ref{sec6:video:generation}).
In our unified framework, the functional components of world models for representation correspond to the cognitive functions of \textbf{\textit{memory}}, \textbf{\textit{perception}}, and \textbf{\textit{language}}.
Together they encode spatio-temporal representations of the world.

\subsubsection{\textbf{Multi-modal Perception.}}\label{sec:worldrep:multi}
The state-of-the-art perception innovations in world model research are multi-modal~\cite{zhen2025tesseract,huang2026pointworld,stereoworld,liang2026lidarcrafter,zheng2024occworld,zhang2024bevworld}. 
This synchronizes with human-like perception which is multi-modal due to our innate need to integrate information across at least five senses used to represent the world~\cite{brandt2024human}.
Additional exemplary works are shown in Figure~\ref{fig:taxonomy}, with more examples discussed in Sec.~\ref{sec:video_wm} to~\ref{sec:collab}.

Perception is often where world model framework's begin. We will start by considering an input of multi-modal observations $o_t^{i}$ for images, $o_t^{\ell}$ for \textit{language}, $o_t^{a}$ for audio, and so on.
We define $\phi_\theta(\cdot)$ as our multi-modal world model for representation, used to create $z_t$, our world's latent representation at step $t$, or in other words,
% \begin{align}\label{eq:our_latent}
%     z_t = \phi_\theta(o_t) \quad \text{s.t.} \quad \dim(z_t) \leq B, \;\;
%     I(o_{1:t}; z_t) \leq C
% \end{align}
\begin{align}\label{eq:our_latent}
    z_t = \phi_\theta(o_t^{i},\, o_t^{\ell},\, o_t^{a}, \dots ) \quad \text{s.t.} \quad \dim(z_t) \leq B, \;\; I(o_{t}; z_t) \leq C
\end{align}
After input encoding, we have latent variable $z_t$ subject to the constraints for physically fitting $z_t$ within a \textit{memory} budget $B$, while also keeping mutual information below $C$, or a sufficient mutual information between the observations and latent space at step $t$.
At this stage, alignment with physical laws is implicitly instilled through training data and handcrafted reward signals (the latter is discussed more later on).
To achieve sufficient world representation for downstream tasks, such as action selection for state transition performed by operator $\mathbb{T}$, we must select an ideal encoder $\phi^{*}_\theta()$ from among state-of-the-art world encoders $\mathcal{F}$, or in other words,
\begin{align}\label{eq:ideal_rep}
\phi_\theta^{*} = \arg\min_{\phi_\theta \in \mathcal{F}}
I\left(o_{t}; \phi_\theta(o_{t})\right)
\quad
\text{s.t.}
\quad
\mathbb{E}\left[R(\mathbb{T}(\phi_\theta(o_{t})))\right] \geq \rho
\end{align}
The above selects an ideal encoder to compress the current observation while preserving enough information for a downstream tasks that produce a sufficient reward for training and inference.
$\mathbb{T}$ may be a generative world model $W_\theta$ from Eq.~\eqref{eq:basic_wm}, a direct policy network, or an MPC planner model. 
This sufficiency value $\rho$ is task dependent, and selected by the researcher conditioned on expected results from competing state-of-the-art representation strategies found in $\mathcal{F}$.

% Or with Soar-styled recursion~\cite{laird2019soar} (enabling some degree of self-monitoring \textit{metacognition}) with world-encoding latent state persistence, 
% \begin{align}\label{eq:our_recursive_latent}
%     z_t = \phi_\theta(z_{t-1}, \mathbf{o}_t^{v},\, \mathbf{o}_t^{\ell},\, \mathbf{o}_t^{a}, \dots ) \quad \text{s.t.} \quad \dim(z_t) \leq B, \;\;
%     I(\mathbf{o}_{1:t}; z_t) \leq C
% \end{align}

Encoders for world representation often are both multi-modal and multi-scalar (spatially, or as discussed in the next section, temporally).
Both features complement each other in practice, and innovative \textit{perception} works rarely breakdown into one or the other~\cite{zhen2025tesseract,huang2026pointworld,stereoworld,liang2026lidarcrafter,zheng2024occworld,zhang2024bevworld}.
TesserAct and PointWorld utilize RGB, Depth, and Normal readings to create a point-cloud representation of the world~\cite{zhen2025tesseract,huang2026pointworld}.
StereoWorld, much like \textit{human}-like perception, utilizes disparity and epipolar constraints on two camera feeds to add a depth scale to 2D appearance~\cite{stereoworld}.
Some embodied world models integrate encoded LiDAR measurements, $o_t^{\text{LiDAR}}$, into the perceptual pipeline and demonstrate that fusing complementary spatial modalities improves robustness and downstream planning performance, particularly in navigation and autonomous systems~\cite{liang2026lidarcrafter,zheng2024occworld}. 
Similarly, BEVworld fuses camera and LiDAR readings to create a Birds Eye View occupancy grid for world representation~\cite{zhang2024bevworld}.
Across these works multi-modal \textbf{\textit{perception}} serves as a mechanism for reducing ambiguity in $z_t$ by grounding latent representations in multiple correlated spatial observation streams.

% Recent systems combining Global Workspace (GW) architectures with world models, such as GW-Dreamer~\cite{maytie2025multimodal}, provide evidence that global information sharing with multimodal latent spaces can improve both sample efficiency and generalization from simulated dreaming. 
% From the perspective of cognitive architecture theory, this approach aligns with the role of \textit{perception} as self-reflection in metacognition. 
% as a gateway to a shared workspace, where diverse sensory inputs are fused into a coherent representation to support downstream reasoning, memory, and action selection.

\subsubsection{\textbf{Multi-scalar Memory}}\label{sec:worldrep:long}
World representation models, world generation models, and unified frameworks performing both representation and generation are meant to learn the state transition dynamics of a world; this means \textit{temporal} as well as \textit{spatial} characteristics.
This reveals a vulnerability in Eq.~\eqref{eq:ideal_rep} we left unaddressed and will correct now.
Our ideal encoder from a single time step $\phi_\theta^{*}(o_{t}))$ captures no state transition dynamics to embed spatio-temporal features in our state space $z_t$.
Furthermore, looking to our constraint, maximizing an expected scalar or low-dimensional reward from a single observed state is likely to lead to reward collapse~\cite{de2018multi} and propagate a credit assignment problem in long horizon tasks with delayed rewards~\cite{rupprecht2022survey}.
We have two options for addressing this, we can create a richer reward signal by enabling \textit{imagination} as hypothetical reasoning over a future time horizon H (discussed more in Sec.~\ref{sec:worldgen:imag}, or as we show now, we can add short-term memory to our observations as $o_{\leq t}$. 
The later adds state transition dynamics to $z_t$ as a restating of Eq.~\ref{eq:our_latent} such that,
\begin{align}\label{eq:our_dynamic_latent}
    z_t = \phi_\theta(\textbf{o}_{\leq t}^{v},\, \textbf{o}_{\leq t}^{\ell},\, \textbf{o}_{\leq t}^{a}, \dots ) \quad \text{s.t.} \quad \dim(z_t) \leq B, \;\; I(\mathbf{o}_{1:t}; z_t) \leq C
\end{align}
This updates our selection criteria for an ideal encoder for world representation re-expressed through Eq.~\eqref{eq:ideal_rep} as,
\begin{align}\label{eq:ideal_dynamic_rep}
\phi_\theta^{*} = \arg\min_{\phi_\theta \in \mathcal{F}}
I\left(\textbf{o}_{\leq t}; \phi_\theta(\textbf{o}_{\leq t})\right)
\quad
\text{s.t.}
\quad
\mathbb{E}\left[R(\mathbb{T}(\phi_\theta(\textbf{o}_{\leq t})))\right] \geq \rho
\end{align}
In practical terms this is the difference between a researcher using V-JEPA~\cite{bardes2023v} over I-JEPA~\cite{assran2023self} when the world is represented through video frames as a sliding window of short-term \textit{memory}.

Short-term \textit{memory} within world model perception makes that perception more \textit{human}-like.
Human perception at the mechanistic level in the human brain utilizes dual streams for spatial and temporal representations of the world~\cite{zachariou2014ventral}.
The field of computer vision already emulates this human spatio-temporal perception by proposing dual stream architectures with one for spatial representations and a second for temporal representations implying short-term memory~\cite{simonyan2014two,ebrahimpour2019ventral}.
World models use this approach now~\cite{zheng2024occworld,liang2026lidarcrafter,zhen2025tesseract,huang2026pointworld}.
We see this spatio-temporal perception in use by OccWorld  and LiDAR crafter which use LiDAR to create an evolving map used by downstream reasoning tasks~\cite{zheng2024occworld,liang2026lidarcrafter}, and in TesserAct utilizing optical flow as one of its multiple input modalities~\cite{zhen2025tesseract}, as is the case in PointWorld which uses 3D point flows to perceive temporal changes in the world~\cite{huang2026pointworld}.

Long-term memory is also used by many of these works to create queryable maps of past world encodings saving compute at inference~\cite{mosaicmem,dang2026sparseworld}.
Mosaicmem, a video diffusion world model, encodes 2d observations as 3d observations leveraging latent memories of previous time-steps' 2d observations to create patch-level and scene level encodings~\cite{mosaicmem}.
SparseWorld's perception is similarly improved due to the presence of queryable memory~\cite{dang2026sparseworld}.

Using memory to instill spatio-temporal characteristics also allows downstream tasks such as sim-to-real ``dreaming'' during training and hypothetical reasoning during inference, which will be discussed in Sec.~\ref{sec:worldgen}. 
This perspective aligns with prior work on traversable latent spaces~\cite{gumbsch2023learning,burchi2025accurate,wang2025genie,bruce2024genie,hansen2025hierarchical}, and suggests that long-context memory in world models is inherently tied to the capacity for spatio-temporal structured latent prediction.
For instance, augmenting Dreamer-style~\cite{hafner2019dream} agents with traversable spatial latent representations enables improved sim-to-real transfer, highlighting the importance of jointly encoding spatial structure alongside temporal dynamics~\cite{burchi2025accurate}. 
Similarly, temporal hierarchy learning introduces multiple scales of abstraction in \textit{perception}, allowing agents to reason over both short-term transitions and long-horizon dependencies~\cite{gumbsch2023learning} again paralleling human-cognition with multiple layers and abstract thinking~\cite{craik1972levels,lecun2022path}. 
Hierarchical world models extend \textit{human}-like multi-layered thinking to control: Hansen et al.~\cite{hansen2025hierarchical} couple a high-level world model that reasons over abstract targets with a low-level model that produces motor commands. 

Finally, world model memory can also be learned through architectural enhancements to world model frameworks.
Autoregressive transformers as with Large Language Models (LLMs) and Vision Language Models (VLMs) use memory in their reasoning towards output tokens corresponding to state or action tokens when adapted as Vision Language Action (VLAs) models~\cite{intelligence2025pi_,wu2026pragmatic,team2025gigabrain}.
Autoregressive diffusion models from Video World Models behave the same way, where every token in the output sequence is conditioned on the previous tokens in the output prompt, and tokens from the input prompt~\cite{casualforcing,rewardforcing,selfforcing}.
Additionally, works have used architectural modules to simulate \textit{memory}, as is the case with Transformer State-Space Models (TSSMs) for bi-directional diffusion models to learn long-range memory dependencies and predict future observations more accurately, and more efficiently than their autoregressive alternatives~\cite{burchi2025accurate}.
We discuss more in Sec.~\ref{sec:video:trends} the usage of memory in architectural reasoning and the usage of memory implementations advances like KV-cache that lead to efficient and state-of-the-art world models~\cite{vaswani2017attention,nguyen2025attention}.
% This is achieved through either short-term memory, or long term queryable memory of previous spatial features.

Adding memory to world models does not just make world models more \textit{human}-like, it is what makes their practical use possible.
The encoding and learning of spatio-temporal features enhances the cognitive function of all of the other component parts of cognition as we saw with \textit{perception} and \textit{language} but also \textit{motivation}, \textit{imagination} and other downstream \textit{reasoning} tasks such as state transition prediction, or action selection.
Additional examples of \textit{memory} innovations are summarized in Fig.~\ref{fig:taxonomy} and discussed in Sec.~\ref{sec:video_wm}--\ref{sec:collab}. 
Unified world models seek an encoder $\phi_\theta(\cdot)$ that produces representations supporting sufficiently long temporal context, as required by Eq.~\eqref{eq:our_dynamic_latent} and Eq.~\eqref{eq:ideal_dynamic_rep}.

\subsubsection{\textbf{Language.}}\label{sec:worldrep:lang}
In world model settings, language can be used to represent the world, but also be used in generative models for conditioned world prediction or action selection.
Large Language Models (LLMs) \cite{zhao2024fully,zhao-etal-2024-pruning,shen2024search,shen2024numerical,zhan2024rethinking,shen2025sparse} and Vision-Language Models (VLMs) \cite{zhao2025open,xu2021vlm,shen2025draftattention,shen2025fastcar,shen2025efficient}, are innately world models themselves~\cite{ge2024worldgpt,gu2024your}. 
How robust or \textit{human}-like LLMs or VLMs perform alone as world models requires more research.
Empirically, LLMs show insensitivity to meaning in comprehension tasks when compared to human counterparts~\cite{dentella2024testing}.
LLMs still fail to reason as humans do in the tasks we ask humans to perform~\cite{ibrahim2025thinking,pate2026replicating}.
However, we know that latent spaces learned by language models converge to similar embeddings across differing modal inputs~\cite{huh2024platonic} and visual embeddings as well as activation patterns are aligned with the human brain~\cite{doerig2025high, schrimpf2021neural}.
Generally, when researchers align an LLM's \textit{language} cognitive function to humans' \textit{language} cognitive function the LLM accuracy improves~\cite{leviathan2025prompt}.
Leviathan et al. does so by repeating LLM prompts to emulate a human's innate ability to hold both the beginning and end of a spoken sentence in their mind at once something an autoregressive LLM is innately unable to do.

\paragraph{Metacognition via Language.}
As discussed in Sec.~\ref{sec:bg:cat} \textit{language} plays a paramount role in human cognition~\cite{tomasello2005understanding}, with some arguing it is language that is directly responsible for human metacognition~\cite{deacon1998symbolic,jaynes2013origin}.
Empirically one can see from our Figure~\ref{fig:catbrain} that language allows a machine to operate over all the other components of cognition.
This is a key element of metacognition, but it remains to be seen if language is the ideal operator of all other components of cognition in machines.
Still when we explore epistemic world models, we repeatedly find language at the center of existing global workspaces used in agent frameworks.
We discuss this more in the Section~\ref{sec:unified:meta}.

\subsubsection{\textbf{A Unified World Model Approach to Representation.}}
Foundational representative world models include Cosmos-Predict~\cite{cosmos-predict1,ali2025world}, LingBot-WM~\cite{team2026advancing}, GIGA-World~\cite{team2025gigaworld}, RLVR-World~\cite{wu2026rlvr}, and finally JEPA~\cite{assran2023self}, which all learn to represent knowledge of the world implicitly aligning with physical laws through training data.
Joint-Embedding Predictive Architectures (JEPA) provides a canonical instantiation of the representation component within our unified world model framework~\cite{assran2023self,vo2024ti,assran2025v,destrade2025value,sun2026vla,wang2026drive}. 
Rather than reconstructing raw sensory inputs, JEPA learns representations by predicting target embeddings from context embeddings in a shared latent space~\cite{assran2023self}, thereby directly optimizing for a more robust predictive structure.
This predictive latent-space formulation naturally supports multiple cognitive functions. 
JEPA encoders can integrate \textbf{\textit{perception}} across multiple modalities~\cite{cornelissen2026mumo} in embodied settings~\cite{wang2026drive}, incorporate \textbf{\textit{language}} as inputs~\cite{vo2024ti}, or to serve as intermediate representations for downstream \textbf{\textit{reasoning}}~\cite{sun2026vla}. 
Crucially, when combined with latent dynamics models, these spatio-temporal representations enable long-horizon \textbf{\textit{memory}} through compact latent world-encodings, and as we will see in the next section, enable hypothetical action rollouts~\cite{assran2025v,destrade2025value}. 
In this sense, state-of-the-art world representation models like JEPA act as $\phi_\theta^*$ creating spatio-temporal features $z_t$ as is the case in our Eq.~\eqref{eq:our_dynamic_latent}.
These JEPA encodings are not merely compressive, but structured to support prediction over imagined futures, aligning directly with the requirements imposed by our reward-constrained information bottleneck formulation in Eq.~\eqref{eq:ideal_dynamic_rep}.
We will see how world models in Sec.~\ref{sec:video_wm} to~\ref{sec:collab} sense and learn world knowledge in each domain application.

\subsection{World Model Prediction and Generation}\label{sec:worldgen}
While world representation defines how spatio-temporal world knowledge is sensed and encoded, world model prediction and generation refers to how the world models make predictions over time and select action traversals within the world. 
World models generate 3D scenes (discussed further in Sec.~\ref{sec4:generation}), generate controlled real-world scenes (discussed further in both Sec.~\ref{sec4:generation} and~\ref{sec:embodied:world-gen}), select actions for state traversal (discussed in Sec.~\ref{sec:embodied:world-gen}) and can reason over Global Workspaces encoded by language (discussed further in Sec.~\ref{sec6:video:generation}).
In our unified framework, the functional components of world models for generation correspond to the cognitive functions of \textbf{\textit{reasoning}}, \textbf{\textit{imagination}}, \textbf{\textit{language}} and \textbf{\textit{motivation}}.
Together they enable structured planning, hypothetical simulation, and decision-making over future trajectories. 
% Given a latent world state $z_t$, the objective is to approximate the optimal action $a_t^*$ defined in Eq.~\eqref{eq:action} by constructing and evaluating sequences of future latent states $\{z_{t+k}\}_{k=1}^{H}$. 
% Thus, world model generation can be understood as a search over latent trajectories induced by $W_\theta$.
\subsubsection{\textbf{Reasoning.}}\label{sec:worldgen:reason} 
Within our taxonomy world model reasoning is defined both as the structured prediction of future world states and as the selection of action rollouts. 
The transition model $W_\theta$ governs the evolution of latent states and when appropriate is conditioned on an action as was the case in Eq.~\eqref{eq:basic_wm}.
From here, the reward function $R(z)$ evaluates the desirability of future latent states and trajectories. 
We treat $W_\theta(z_t, a_t)$ as a parameterized world transition function, though it may in practice represent a stochastic process.

In a search for the optimal action $a_t^*$ at time step $t$, the reward signal is considered over the time horizon $H$, 
\begin{align} \label{eq:action}
a^*_{t:t+H} = \arg\max_{a_{t:t+H}} \mathbb{E}\left[\sum_{k=0}^{H} R(z_{t+k})\right] 
\end{align}
Thus, action selection can be understood as inference over spatio-temporal latent trajectories, where candidate futures are simulated and evaluated under the reward functional.
The predicted $z_{t+1}$ is obtained by $W_\theta$, our unified World Model for world model generation guided by principles from Cognitive Architecture Theory.

We identify eight recurring design paradigms for how world model reasoning occurs, shown in across Figures~\ref{fig:vwm_architecturesa} to~\ref{fig:collab_architectures}, that span the spectrum from purely discriminative action selection to full generative world simulation. 
At one extreme lie Vision-Language-Action models (VLAs), such as $\pi_{0.5}$~\cite{intelligence2025pi_}, LingBot-VLA~\cite{wu2026pragmatic}, and other VLA methods that map observations directly to actions through a VLM backbone augmented with a flow-matching action expert~\cite{hu2026ar}, bypassing explicit future-state prediction entirely~\cite{zhuns,destrade2025value,jiang2025irl,lin2025vote,sun2026vla,liu2026mmada}.
At the other extreme, some world models only represent the world~\cite{vo2024ti}, and use a separate value~\cite{destrade2025value} or policy models for action or future state generation~\cite{cosmos-predict1,li2026causal}.
These World Action Models leverage training data and are not conditioned on any action.
% Some world models lack the ability to represent the world and only make decisions  like with VLAs~\cite{intelligence2504pi0,zhuns,destrade2025value,sun2026vla,jiang2025irl}.
This decoupling of representation and action generation grants maximum flexibility in imagination but introduces a two-system overhead, in which errors in world prediction and errors in action selection compound independently.
World Action Models~\cite{ye2026world,li2026causal,ye2026gigaworld} tend to be more robust and work better with longer and noisier time horizons.
Additional methods include autoregressive transformer~\cite{chen2025drivinggpt,zhan2024exploring,liu2025toward,zhan2024rethinking,zhuang2025video} and diffusion models~\cite{zhang2025epona,yu2024fastervd,huang2025ladi,zhan2024fast,shen2024lazydit,zhu2025astra,selfforcing}.
Motus claims to be a unified world model because it successfully unites generating latent action, future video states, and agent actions~\cite{bi2026motus}.

\subsubsection{\textbf{Imagination.}}\label{sec:worldgen:imag}
During world model training, \textbf{\textit{imagination}} manifests as dreaming~\cite{ha2018world}, referring to a world model's capability to learn to represent world dynamics through unsupervised training.
If these learned spatio-temporal latent representations are traversable and remain consistent under domain shift, then an action-policy model can be trained with these latent spaces and deployed in a real-world domain~\cite{ha2018world,durante2026videoweave,maytie2025multimodal}.
In real-world applications, training data can be prohibitively scarce and expensive to create; sim-to-real training overcomes this problem~\cite{rupprecht2022survey}.
Figures~\ref{fig:vwm_architecturesc} through~\ref{fig:emb_architecturesb} show architectures that support hypothetical reasoning in inference, and dreaming in training both in the video world model and embodied world model domains.

During inference, \textbf{\textit{imagination}} refers to hypothetical reasoning, planning, or simulating future states of the world over a set of possible action rollouts to find an optimal action~\cite{xiang2025pan,xing2025critiques}.
This is only possible after applying lessons from Sec.~\ref{sec:worldrep}, where Eq.~\eqref{eq:ideal_dynamic_rep} creates a traversable world-encoding latent space.
The optimal action, or $a_t^*$, maximizes the expected reward signal over the time horizon of the possible action rollout.
For action-planning we assume a decomposition of the world model into a transition function and an observable reward functional. 
When our unified framework for representation successfully encodes spatio-temporal representations upstream in Eq.~\eqref{eq:ideal_dynamic_rep} we are ensured that imagined trajectories that remain consistent over extended horizons~\cite{durante2026videoweave,gumbsch2023learning,gao2025adaworld,wu2026visual,xiang2025pan}. 

% \subsubsection{\textbf{A Unified World Model Approach to Generation}}

% In our taxonomy of world models, \textbf{\textit{reasoning}} is defined as the structured prediction of future world states in Eq.~\eqref{eq.worldgen} and the selection of action rollouts for the optimization objective in Eq.~\eqref{eq:action}. 
% Recent work highlights the importance of maintaining narrative and logical consistency across the generated trajectories, particularly in long-horizon settings where compounding errors can degrade performance~\cite{xing2025critiques,li2026causal}. 
% Hierarchical approaches further decompose reasoning into multi-scale planning processes, enabling agents to coordinate short-term actions with long-term goals~\cite{gumbsch2023learning}. 
% Recently, World Action Models propose an end-to-end generative model for proposing actions and predicting future states of the world during training~\cite{ye2026world,li2026causal,ye2026gigaworld,yuan2026fastwamworldactionmodels,yuan2026fastwamworldactionmodels,akbari2026flash}.
% Dream-zero proposes using imagination as a planner~\cite{ye2026world}.
% Gigaworld-Policy uses imagination for supervision during training~\cite{ye2026gigaworld}, and Lingbot-VA uses dreaming as an iterative denoising step~\cite{li2026causal}.
% Often, World Action Models use dreaming as a regularizer during training, and do not use dreaming states during inference due to latency concerns~\cite{ye2026gigaworld}.

% \subsection{Unified world models in the Wild}\label{sec:unified:wild}
\subsection{Cognitive Architecture in Unified World Models} \label{sec:unified:cat}
Our taxonomy categorizes the surveyed works according to the cognitive functions. 
Many works incorporate several cognitive functions in their designs and sometimes innovate along multiple cognitive axes, which is why some works appear in multiple sections or have several check marks in Tables~\ref{tab:vwm_models},~\ref{tab:embodied_wm}, and~\ref{tab:human_ai_collab}.
Analyzing the innovation trends presented in our report through Tables~\ref{tab:vwm_models},~\ref{tab:embodied_wm} and~\ref{tab:human_ai_collab} show that a research gap exists regarding the cognitive functions of \textbf{\textit{motivation}}, and \textbf{\textit{metacognition}}.
The utilization of our taxonomy makes this research gap clear and may not have become apparent otherwise.
It is clear that machine and human cognition continue to be intertwined.

An exciting new theory of human cognition, Active Inference~\cite{friston2010free,friston2016active}, explains human cognition as a system that minimizes surprise about sensory inputs by continuously updating beliefs about the world and taking actions that make the world match those beliefs.
Active Inference has been extended to machine cognition~\cite{pezzulo2024active,friston2025active} as a theory that novelly \textbf{\textit{motivates}} learning in machines and informs our unified world model Framework (discussed further in Sec.~\ref{sec:unified:cat}).
world model works have already begun to incorporate Active Inference~\cite{maytie2025multimodal}.

% \subsubsection{Future Work}
% Across our report of state-of-the-art world models, we have identified two innovation gaps in research regarding two components of our Unified world models: \textbf{\textit{motivation}} and \textbf{\textit{metacognition}}.
% This is elucidated in part by evaluative benchmarks from recent works~\cite{chatlatanagulchai2025use,kim2025humanoid,pagan2025computational,puyin2026quantiphy,song2025evaluating,wang2025enactevaluatingembodiedcognition}.
% But also, one only needs to look at Tables~\ref{tab:vwm_models}, ~\ref{tab:embodied_wm}, and~\ref{tab:human_ai_collab} to see the lack of innovations regarding \textbf{\textit{motivation}} and \textbf{\textit{metacognition}}.
% To overcome these gaps, regarding \textbf{\textit{metacognition}}, researchers should look to the agent frameworks used in human-AI collaboration for Scientific Discovery, as seen in Figure~\ref{fig:collab_architectures}; and regarding \textbf{\textit{motivation}}, researchers should look to advances from human-machine cognitive architecture theory regarding Active Inference~\cite{friston2016active,pezzulo2024active,friston2025active,maytie2025multimodal}.
% toobaimt: rewritten for clarity
Across our report on state-of-the-art world models, we have identified two significant research gaps regarding two components of our unified world models: \textbf{\textit{motivation}} and \textbf{\textit{metacognition}}. This issue is, in part, elucidated by evaluative benchmarks from recent studies \cite{chatlatanagulchai2025use, kim2025humanoid, pagan2025computational, puyin2026quantiphy, song2025evaluating, wang2025enactevaluatingembodiedcognition}. Furthermore, Tables \ref{tab:vwm_models}, \ref{tab:embodied_wm}, and \ref{tab:human_ai_collab} clearly demonstrate the lack of innovation in the areas of \textbf{\textit{motivation}} and \textbf{\textit{metacognition}}.
To address these gaps in \textbf{\textit{metacognition}}, we observe the importance of language in human cognition may be an indication of its importance in machine cognition, and suggest that researchers focus on the agent frameworks, as illustrated in Figure \ref{fig:collab_architectures} as what we call epistemic world models. In terms of \textbf{\textit{motivation}}, advancements in human-machine cognitive architecture theory, particularly regarding Active Inference \cite{friston2010free,friston2016active, pezzulo2024active, friston2025active, maytie2025multimodal}, should be explored.

\subsubsection{\textbf{Motivation.}}
Reinforcement Learning (RL) techniques instill external motivation into world models through reward signals, using a reward function defined external to the world model itself.
Enabling \textbf{\textit{imagination}} or hypothetical reasoning, as described by Eq.~\eqref{eq:action}, requires an observable reward functional $R(z_t)$~\cite{hansen2024td,wu2026rlvr,hao2025neural,khanzada2025indrive,maytie2025multimodal,nachkov2025dream}.
Multi-task learning allows the pursuit of multiple goals~\cite{maytie2025multimodal} and can help align models to the physical world when trained with an appropriate reward signal~\cite{le2025gravity}.
Researchers construct RL agents as a potential solution to artificial general intelligence, assuming that their reward signal is sufficient~\cite{silver2021reward}.
However, often RL provides training mechanisms for world models using hand-crafted reward signals that do not generalize~\cite{rupprecht2022survey} and rewards that are misaligned with the intended operational goal~\cite{amodei2016concrete,hadfield2017off}.
As such, they require additional interventions such as Reinforcement Learning from Human Feedback in some applications~\cite{puyin2026quantiphy,lin2025exploring}.

\paragraph{Intrinsic Motivation.} Promising research directions exist, such as maximizing influence over future states~\cite{klyubin2005empowerment} or minimizing energy (i.e. surprise)~\cite{friston2016active,pezzulo2024active,friston2025active} to explore with training mechanisms decoupled from an externally defined and action-based reward signal.
Both maximizing the influence of future states and active inference are compatible with the state-based reward function described in Eq.~\eqref{eq:action}.
Active Inference formalizes ``minimizing surprise'' as the minimization of variational free energy, a tractable upper bound on negative log model evidence, which decomposes into accuracy and complexity and extends to action selection through expected free energy that balances information gain and value. Concretely, given a latent world model with states $z_t$, observations $o_t$, and actions $a_t$, the generative model is defined as
\begin{equation}
p_\theta(o_{1:T}, z_{1:T} \mid a_{1:T}) = \prod_{t=1}^T p_\theta(o_t \mid z_t)\, p_\theta(z_t \mid z_{t-1}, a_{t-1}).
\end{equation}
Inference proceeds by introducing an approximate posterior $q_\phi(z_{1:T} \mid o_{1:T})$ and minimizing the variational free energy
\begin{equation}
\mathcal{F} = \mathbb{E}_{q_\phi(z_{1:T})} \left[ \log q_\phi(z_{1:T}) - \log p_\theta(o_{1:T}, z_{1:T}) \right],
\end{equation}
which upper bounds the surprise $-\log p_\theta(o_{1:T})$. This objective decomposes as
\begin{equation}
\mathcal{F} = D_{\mathrm{KL}}\!\left(q_\phi(z_{1:T}) \,\|\, p_\theta(z_{1:T})\right)
- \mathbb{E}_{q_\phi}\!\left[\log p_\theta(o_{1:T} \mid z_{1:T})\right],
\end{equation}
corresponding to complexity and accuracy terms, respectively. For action selection, policies $\pi$ are evaluated by minimizing the expected free energy
\begin{equation}
\mathcal{G}(\pi) = \mathbb{E}_{q_\phi}\!\left[ \underbrace{D_{\mathrm{KL}}\!\left(q_\phi(z_{t+1:T} \mid \pi) \,\|\, p_\theta(z_{t+1:T})\right)}_{\text{information gain}} 
- \underbrace{\log p(o_{t+1:T})}_{\text{value}} \right],
\end{equation}
encouraging trajectories that both reduce uncertainty in latent dynamics and achieve preferred outcomes.
As such, world model research has already begun to incorporate Active Inference.

Using intrinsic motivation for world models is drastically unrealized, but there are some recent works using intrinsic motivation within RL that validate this research direction~\cite{pathak2017curiosity,berseth2019smirl,shen2025quartdepth,dung2024r,pazem2025free,yeganeh2025deep,yang2025cambrian,arshid2025toward,kashirskiy2026sus,rome2026learning,khanzada2025indrive,sridhar2024nomad,sekar2020planning}.
Among these works, Active Inference is also being deployed in embodied settings to promote exploration during training~\cite{arshid2025toward,rome2026learning}, and in one case, even control swarms of unmanned autonomous vehicles~\cite{arshid2025toward}.
Seven of these are explicit world model works~\cite{dung2024r,yeganeh2025deep,yang2025cambrian,rome2026learning,khanzada2025indrive,sridhar2024nomad,sekar2020planning}.
Three explicitly use Active Inference as a reward signal for world models~\cite{dung2024r,yeganeh2025deep,yang2025cambrian}.
Cambrian-S explicitly references prediction error as a proxy for measuring surprise but falls short of implementing Active Inference's full exploratory learning objective~\cite{yang2025cambrian}.

It is also worth noting that researchers are ultimately the ones selecting architectures, reward criteria, and training sets within world model research.
Peer review evaluates objectively if the subsequent empirical results are novel and notable.
It is researchers who breathe life into world models by aligning experiment designs with broader research trends.
It is here researchers lend their human cognition in machine cognition with a single example being the selection of $B$, $C$, and $\rho$ as thresholds found in the constraints described in Eq.~\eqref{eq:our_latent} to~\eqref{eq:ideal_dynamic_rep} of our unified world model framework.
This is further demonstrated in the global workspaces of epistemic world models used by agent frameworks for AI-Human Collaboration~\cite{gottweis2025towards,gottweis2026accelerating,shao2025sciscigpt,shao2025omnisci} shown in Figure~\ref{fig:collab_architectures} and discussed in Sec.~\ref{sec:collab}.

\subsubsection{\textbf{Metacognition.}}\label{sec:unified:meta} Soar's usage of sub-tasks and recursive operator decomposition mimics human metacognition by breaking bigger, more abstract tasks into smaller, more discrete tasks~\cite{craik1972levels, laird2019soar}.
For example, if an agent with a world model is navigating a problem and not experiencing progress towards a goal, sub-tasks allow the debugging of this lack of progress by operating over memory, perception, and other components of cognition to propose alternative routes to the goal; demonstrating a minimal degree of self-reflection, self-evaluation and self-control.
While world models' capability to align with \textit{human}-like metacognition is theoretically possible, it still represents a largely unsolved implementation problem~\cite{johnson2025imagining,porkebski2025there,pate2026replicating}.

\paragraph{Metacognition via Global Workspaces.}
Exploring metacognition through the use of agent systems capable of self-reflection, self-evaluation, and self-control shows some progress towards the goal of implementing truly metacognitive capabilities within world models~\cite{bilal2025meta,wang2026metamind}.
Figure~\ref{fig:collab_architectures} shows how these agent frameworks provide a world model that enables some amount of metacognition through the use of a Global Workspace~\cite{baars1993cognitive} accessible by agents and human collaborators.
Altogether, the central limitation identified in world models is the absence of explicit metacognitive mechanisms in latent world models, particularly in embodied and video domains seen in Sec.~\ref{sec:video_wm} and~\ref{sec:embodied_wm}. 
While these systems learn latent state representations $z_t$ and transition dynamics $z_{t+1} = W_\theta(z_t, a_t)$, they lack a mechanism for monitoring, evaluating, or controlling the routing internal computations. 

For latent world models in settings that cannot fully replicate a truly Global Workspace due to context size constraints, we see research begin to approximate it using a mixture-of-experts (MoE) paradigm with a world-aware router function.
A router function selecting an expert exhibits some behavior indicative of metacognition as self-control.
The MoE implementation functions similarly to specialists within multi-agent frameworks and is a growing trend in world models~\cite{wu2026mixture,tang2025moe,jang2026test}.
A world-aware router could be extended to select the amount of loops to be perform in a looped transformer LLM~\cite{yang2024looped} indicating self-control over an amount of reasoning performed (i.e. reasoning about reasoning).
Similarly, to improve self-reflection, increasing input modalities (i.e., increasing the amount of specialists) to augment \textbf{\textit{perception}} from ``first person'' video alone to include an additional ``third person'' video source.
For a fully self-evaluating metacognitive process, using imagination over an MoE implementation, or integrating a router function that can decide to re-reason through a problem, would function similarly to an evaluation specialist operating in a Global Workspace, as discussed in Sec.~\ref{sec:collab}.

Drawing from Cognitive Architecture Theory~\cite{newell1994unified} and the Global Workspace Theory (GWT)~\cite{baars1993cognitive}, metacognition can be interpreted as the ability to self-reflect, self-evaluate, and self-control. 
Notably, while such mechanisms are largely missing from latent world models, they emerge naturally in epistemic world models, where shared workspaces are instantiated through external artifacts such as documents, tool outputs, and execution traces. 
These systems approximate the self-reflection, self-evaluation, and self-control, albeit over distributed and externalized state representations, often as interpretable language, rather than internal latent variables. 
As  we will discuss in Sec.~\ref{sec:collab}, recent agentic systems for scientific discovery provide early instantiations of such global workspaces, suggesting a viable pathway toward incorporating metacognitive capabilities into unified world models using interpretable language.
The typical architecture for such an agentic system is shown in Figure~\ref{fig:collab_architectures}.
\section{Video World Models} \label{sec:video_wm}

Video world models aim to learn compact representations of visual environments together with their temporal evolution. A common abstraction is a latent dynamical system:
\begin{equation}
    z_{t+1} = f_{\theta}(z_t, a_t), \quad x_t = g_{\phi}(z_t),
    \label{eqn:latent}
\end{equation}
where $z_t \in \mathbb{R}^d$ denotes a latent state encoding scene geometry, object configurations, physical factors, and temporal context; $a_t$ denotes external inputs, actions or control signals; $f_{\theta}$ models state transitions; and $g_{\phi}$ maps latent states back to observations. Under this view, video is generated through state evolution and rendering, rather than direct frame-wise synthesis.

Recent advances have positioned video world models as powerful engines for simulating the visual world in 2D space~\cite{zhang2026videorepa,jang2026self,bardes2024revisiting,durante2026videoweave,helios,garrido2026learning,yang2025cambrian,xiang2026geometry,le2025gravity,yuan2026inference,gumbsch2023learning,stereoworld,zhao2026phyworld}, 3D space~\cite{marbleworldblog2026,feifeiblog2026,mosaicmem}, and hybrid geometric settings~\cite{bagchi2026walk,dharmarajan2025dream2flow,burchi2025accurate,gao2025adaworld}. As summarized in Table~\ref{tab:vwm_models}, these models differ not only in their geometric assumptions, but also in the cognitive functions they operationalize, namely perception, memory, reasoning, imagination, motivation, and metacognition.

The defining characteristic of modern video world models is their ability to predict future visual states conditioned on current observations and, more crucially, on latent or explicit actions. Contemporary systems such as the open-source Wan~\cite{wan2025wan}, alongside models such as Sora~\cite{sora_technical_report} and Kling~\cite{team2025kling}, do not merely synthesize pixels; with proper latent representations and training techniques, they can learn the underlying physical, causal, and temporal laws governing their simulated environments. Overall, a world model must go beyond visually plausible synthesis; it should support persistent state representation, causal transition modeling, controllable intervention, and physically grounded prediction. This distinction marks the transition from passive text-to-video generation toward interactive, controllable, and agent-facing environments.

\begin{figure}[t!]
    \centering
    \includegraphics[width=\linewidth]{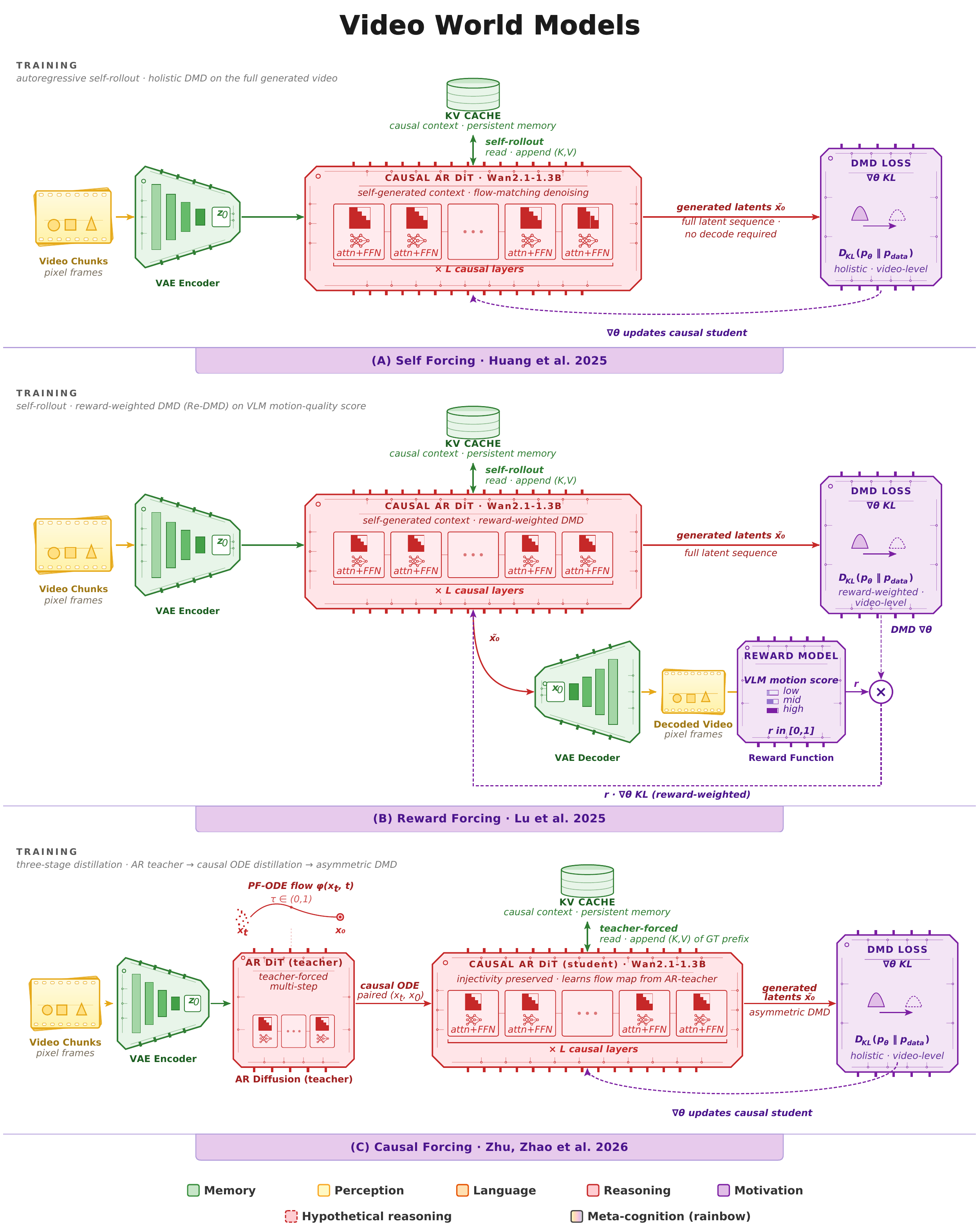}
    \vspace{0.1cm}
    \hrule
    \vspace{0.1cm}
    \caption{
    Representative architectural paradigms in video world models, including autoregressive causal rollout, bidirectional or masked generation, and promptable or action-conditioned prediction.}
    \label{fig:vwm_architecturesa}
\end{figure}

\begin{figure}[t!]
    \centering
    \includegraphics[width=\linewidth]{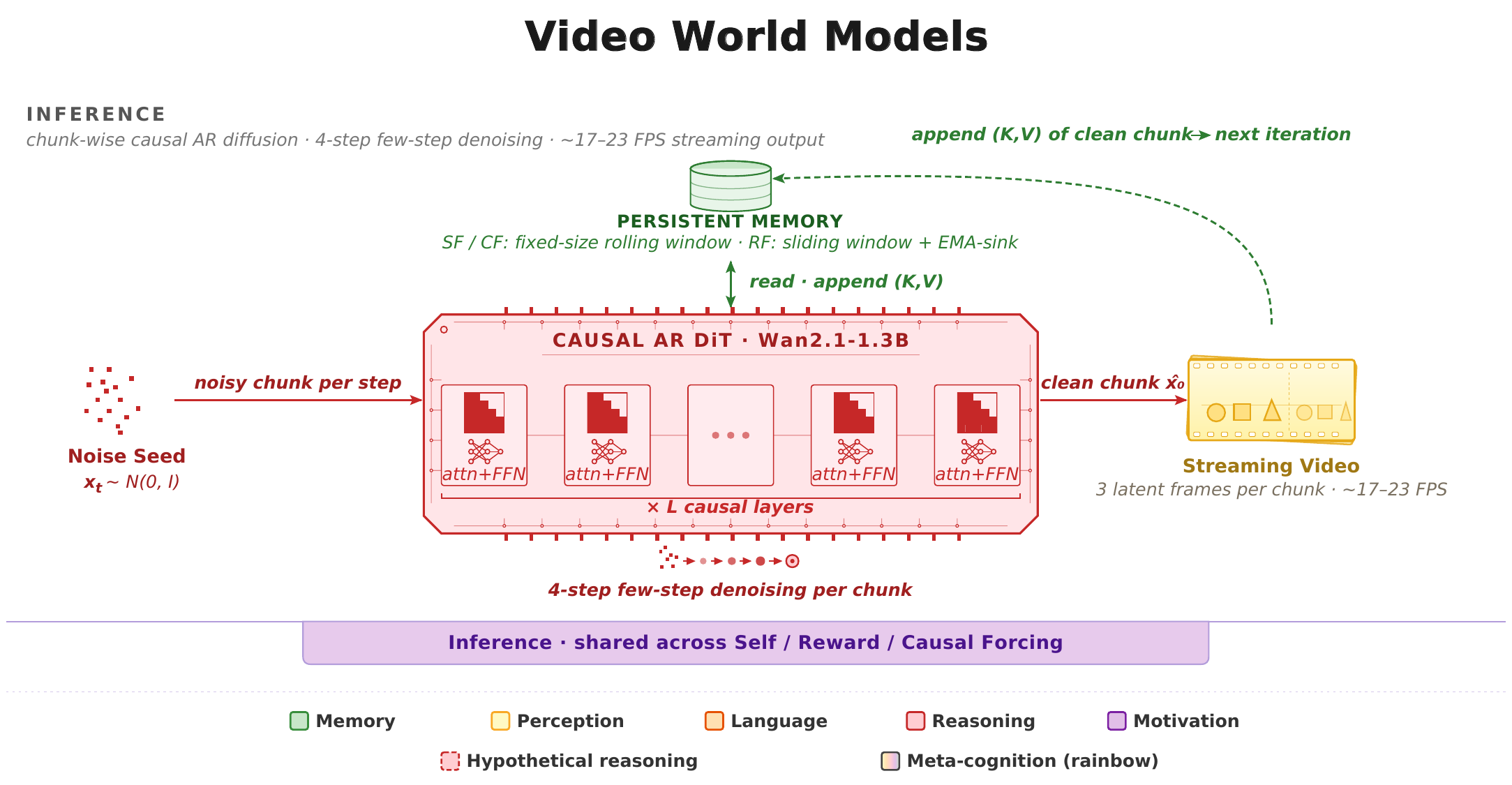}
    % \vspace{-0.1cm}
    \hrule
    % \vspace{-0.1cm}
    \caption{
    Representative architectural paradigms in video world models, including autoregressive causal rollout, bidirectional or masked generation, and promptable or action-conditioned prediction.}
    \label{fig:vwm_architecturesb}
\end{figure}

\begin{figure}[]
    \centering
    \includegraphics[width=\linewidth]{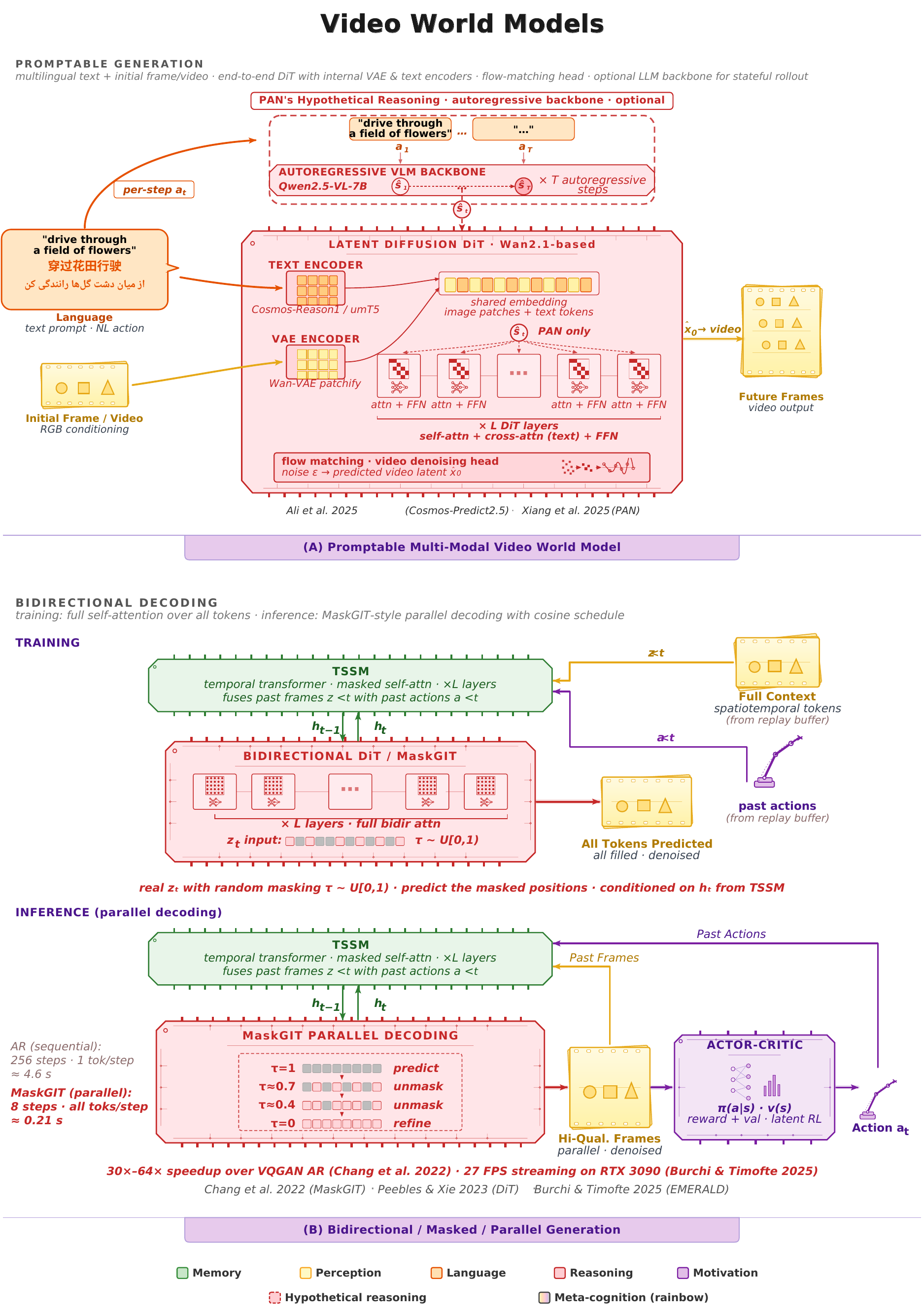}
    \vspace{-1cm}
    % \hrule
    % \vspace{-0.1cm}
    \caption{
    Representative architectural paradigms in video world models, including autoregressive causal rollout, bidirectional or masked generation, and promptable or action-conditioned prediction.}
    \label{fig:vwm_architecturesc}
\end{figure}

In Sec.~\ref{sec4:representation}, we review how video world models construct internal representations of the world, focusing on perception and memory. In Sec.~\ref{sec4:generation}, we discuss how these representations are used for prediction and generation, emphasizing reasoning, imagination, and motivation. 
Sec.~\ref{sec:video:trends} summarizes key trends and limitations of epistemic world models.
Figures~\ref{fig:vwm_architecturesa}~\ref{fig:vwm_architecturesb} and~\ref{fig:vwm_architecturesc} illustrate representative architectures encountered across 2D, 3D, and hybrid video world models.

% \begin{table*}[]
\begingroup
\centering
\scriptsize
\setlength{\tabcolsep}{3.0pt}
\renewcommand{\arraystretch}{1.05}
\begin{longtable}{l l c c c c c c c c p{4.5cm}}
\caption{Comparison of recent video world models according to geometry and cognitive functions. A check (\cmark) indicates that the work explicitly contributes to the corresponding cognitive function in the context of video world models.}
\label{tab:vwm_models} \\
% \centering
% \scriptsize
% \noindent\makebox[\textwidth]{%
% \setlength{\tabcolsep}{3.0pt}
% \renewcommand{\arraystretch}{1.05}
% \begin{tabular}{l l c c c c c c c c p{4.5cm}}
% \toprule
%  \textbf{Work} & \textbf{Year} & \textbf{Geometry} & \rotatebox{90}{\textbf{Mem.}} & \rotatebox{90}{\textbf{Perc.}} & \rotatebox{90}{\textbf{Lang.}} & \rotatebox{90}{\textbf{Reas.}} & \rotatebox{90}{\textbf{Imag.}} & \rotatebox{90}{\textbf{Moti.}} & \rotatebox{90}{\textbf{Meta.}} & \textbf{Description} \\
% \midrule
\toprule
\textbf{Work} & \textbf{Year} & \textbf{Geometry} & 
\rotatebox{90}{\textbf{Mem.}} & 
\rotatebox{90}{\textbf{Perc.}} & 
\rotatebox{90}{\textbf{Lang.}} & 
\rotatebox{90}{\textbf{Reas.}} & 
\rotatebox{90}{\textbf{Imag.}} & 
\rotatebox{90}{\textbf{Moti.}} & 
\rotatebox{90}{\textbf{Meta.}} & 
\textbf{Description} \\
\midrule
\endfirsthead
\multicolumn{11}{c}{\tablename\ \thetable{} -- \textit{continued from previous page}} \\
\toprule
\textbf{Work} & \textbf{Year} & \textbf{Geometry} & 
\rotatebox{90}{\textbf{Mem.}} & 
\rotatebox{90}{\textbf{Perc.}} & 
\rotatebox{90}{\textbf{Lang.}} & 
\rotatebox{90}{\textbf{Reas.}} & 
\rotatebox{90}{\textbf{Imag.}} & 
\rotatebox{90}{\textbf{Moti.}} & 
\rotatebox{90}{\textbf{Meta.}} & 
\textbf{Description} \\
\midrule
\endhead
\midrule
\multicolumn{11}{r}{\textit{continued on next page}} \\
\endfoot

\bottomrule
\endlastfoot

\multicolumn{11}{l}{\textit{Representation: memory and/or perception}} \\
\midrule
V-JEPA~\cite{bardes2024revisiting} & 2024 & 2D & \cmark & \cmark &  & \cmark &  &  &  & Self-supervised video learning by predicting masked spatio-temporal regions in latent space. \\
Gumbsch et al.~\cite{gumbsch2023learning} & 2024 & 2D+temporal & \cmark & &  & \cmark & \cmark & \cmark &  & Learn when the world meaningfully changes via discrete latent dynamics, then build a high-level model that skips between change points. \\
VideoREPA~\cite{zhang2026videorepa} & 2025 & 2D &  & & & \cmark &  & \cmark &  & Physics-aware video generation via relational distillation. \\
Cosmos-Predict2.5~\cite{ali2025world} & 2025 & 2D & \cmark &  &  & \cmark &  & \cmark &  & Unified flow-matching video generator trained on clips for Physical AI simulation, data augmentation, and policy evaluation. \\
VideoWeave~\cite{durante2026videoweave} & 2026 & 2D & \cmark &  &  &  &  &  &  & Splice short captioned videos into synthetic long videos to cheaply train better video-language models. \\
Helios~\cite{helios} & 2026 & 2D & \cmark &  &  & \cmark &  & \cmark &  & 14B model running real-time on one H100 via context compression and drift-aware training. \\
LingBot-World~\cite{team2026advancing} & 2026 & 2D & \cmark &  & \cmark & \cmark &  \cmark &  &  & Block-causal video generator with sub-second latent rollout for agent training \\

\midrule
\multicolumn{11}{l}{\textit{Geometry-aware perception and spatial memory}} \\
\midrule

LTX~\cite{hacohen2024ltx} & 2024 & 2D &  & \cmark &  & \cmark &  & \cmark &  & Uses video diffusion in a real-time open-source world model. \\
Sparse World~\cite{dang2026sparseworld} & 2025 & 4D & \cmark & \cmark &  &  &  \cmark &  & \cmark & Range-Adaptive Perception module learns queries modulated by the ego vehicle with temporal-spatial associations to enable extended-range perception. Some self-control over learning during training. \\
% Marble world model~\cite{marbleworldblog2026,feifeiblog2026} & 2025 & 3D &  & \cmark &  & \cmark &  &  &  & Multimodal 3D world generator. \\
MosaicMem~\cite{mosaicmem} & 2026 & 3D & \cmark & &  & \cmark &  &  &  & Hybrid 3D-patch + latent memory for video world models. \\
StereoWorld~\cite{stereoworld} & 2026 & 2D+depth &  & \cmark &  &  &  \cmark &  &  & Generate stereo video natively via camera-aware RoPE and epipolar-constrained attention, grounding geometry from disparity. \\
ViewRope~\cite{xiang2026geometry}  & 2026 & 2D+depth & \cmark & \cmark &  & \cmark &  &  &  & Replace pixel-grid positional embeddings with camera-ray-based RoPE so video world models stay 3D-consistent across viewpoints \\

\midrule
\multicolumn{11}{l}{\textit{Reasoning and action-conditioned prediction}} \\
\midrule

DIAMOND~\cite{alonso2024diffusion} & 2024 & 2D &  & \cmark &  & \cmark & \cmark &  &  & Conditions on previous latent state and actions to produce future frames. \\
Dream2Flow~\cite{dharmarajan2025dream2flow} & 2025 & Hybrid &   &  &  & \cmark & \cmark & \cmark &  & Generate human-interaction videos, extract 3D object trajectories, then have robots track those trajectories to manipulate. \\
Garrido et al.~\cite{garrido2026learning} & 2026 & 2D &  & \cmark &  & \cmark &  &  &  & Learn action-conditioned world models from unlabeled in-the-wild videos by inferring continuous latent actions via inverse dynamics. \\
EgoWM~\cite{bagchi2026walk} & 2026 & Hybrid &  &  &  & \cmark & \cmark &  &  & Fine-tune pretrained video diffusion models with lightweight action conditioning to get cross-embodiment egocentric world models. \\
EMERALD~\cite{burchi2025accurate} & 2026 & Hybrid & \cmark &  &  & \cmark &  & \cmark &  & MaskGIT-based parallel token prediction in spatial latent space for accurate yet efficient model-based RL world models. \\

\midrule
\multicolumn{11}{l}{\textit{Imagination / embodied world generation}} \\
\midrule

AdaWorld~\cite{gao2025adaworld} & 2025 & Hybrid & \cmark & \cmark &  &  & \cmark &  &  & Pretrain world models with self-supervised continuous latent actions from video, then cheaply adapt to new environments by mapping real actions to latent ones. \\
GigaWorld-0~\cite{team2025gigaworld} & 2025 & 3D &  & \cmark &  & \cmark & \cmark &  &  & World model data engine integrating video generation and 3DGS with physics simulation \\
TesserAct~\cite{zhen2025tesseract} & 2025 & 3D &  & \cmark &  & \cmark & \cmark &  &  & 4D embodied world model converting generated RGB-D-Normal video to point clouds for action prediction \\
Genie~\cite{bruce2024genie} & 2024 & 2D & \cmark &  &  & \cmark & \cmark & \cmark &  & Unsupervised training on unlabeled internet video for conditional spatio-temporal world representations enabling simulation training with RL and downstream policy models. \\
\midrule
\multicolumn{11}{l}{\textit{Motivation / reward- or surprise-guided alignment}} \\
\midrule

Le et al.~\cite{le2025gravity} & 2025 & 2D &  & \cmark &  &  &  & \cmark &  \cmark & Post-train video diffusion with verifiable Newtonian rewards to enforce physically correct motion. Self-evaluation during training. \\
WMReward~\cite{yuan2026inference} & 2025 & 2D &  & \cmark &  & \cmark &  & \cmark & \cmark & Post-train an action-conditioned world model for zero-shot robot planning. Self-reflection, Self-control, self-evaluation at inference. \\
Cambrian-S~\cite{yang2025cambrian} & 2025 & 2D & \cmark & \cmark &  & \cmark &  & \cmark &  & Define spatial supersensing hierarchy, benchmark it, and use prediction-error as surprise to drive memory/attention in long videos. \\
PhyWorld~\cite{zhao2026phyworld} & 2026 & 2D & \cmark & & & & & \cmark & & Direct Policy Optimization for physically aligned video generation \\
\midrule
\multicolumn{11}{l}{\textit{Metacognition / self-improvement}} \\
\midrule

RLVR-World~\cite{wu2026rlvr} & 2025 & 2D &  &  &  & \cmark & \cmark & \cmark & \cmark & Trains world models with RL using verifiable rewards; enables self-improving world models \\
Jang et al.~\cite{jang2026self} & 2026 & 2D &  & \cmark &  & \cmark &  &  & \cmark & Self-reflection through iteratively denoise-and-re-noise video latents at inference time to self-evaluate for physics/motion artifacts. \\
\end{longtable}
\endgroup
% \end{table*}

\subsection{World Model Representation}\label{sec4:representation}

Effective video world models require representations that capture both the spatial structure of a scene and its temporal evolution. In the language of Cognitive Architecture Theory (CAT), this corresponds primarily to \textbf{\textit{perception}}, which encodes the state space for world interaction, and \textbf{\textit{memory}}, which preserves information needed for temporal coherence, long-horizon prediction, and persistent world understanding.

\subsubsection{\textbf{Perception.}}

Perception in video world models refers to the ability to encode visual observations into latent states that preserve object identity, spatial layout, motion, geometry, and physical regularities. This is especially challenging in generated video, where small perceptual errors can accumulate into temporal artifacts, geometric drift, or physically implausible motion.

Several works address perceptual consistency directly during inference time. Self-Refining Video Sampling~\cite{jang2026self} uses the video generator itself as an inference-time refiner. Rather than relying on an external verifier or additional training, it iteratively denoises and re-noises video latents to reduce artifacts and improve motion consistency. PhysicsMind~\cite{mak2026physicsmind} complements such methods by introducing a benchmark for evaluating physical perception and generation, including both visual question answering (VQA) and video generation tasks.

Other works improve perception during training, with some vision-language models showing that carefully curated multimodal data alone can help align models with physical laws. Order of Chaos~\cite{cao2026order} shows that carefully curated multimodal data can implicitly align vision-language models with physical regularities. Its PhysGame dataset uses question-answer pairs about glitches and visual anomalies in video-game footage, suggesting that simulated environments can provide scalable supervision for physical and perceptual alignment.

Geometry-aware perception has also become increasingly important for maintaining viewpoint consistency and reducing spatial drift in video world models. ViewRope~\cite{xiang2026geometry} replaces standard pixel-grid positional encodings with camera ray-based rotary embeddings, allowing attention to reason over viewpoint geometry. Similarly, StereoWorld~\cite{stereoworld} grounds stereo video generation through disparity and epipolar constraints, while 3D and hybrid models such as MosaicMem~\cite{mosaicmem} and TesserAct~\cite{zhen2025tesseract} explicitly lift visual evidence into spatial representations. Together, these works suggest that robust perception in video world models increasingly requires not only appearance modeling, but also geometry-aware latent structure.

\subsubsection{\textbf{Memory.}}

Memory enables video world models to maintain coherent latent states over time. This includes short-term memory for local temporal consistency, long-term memory for persistent objects and scene layout, and structural spatio-temporal memory for maintaining geometric or semantic continuity across viewpoints, clips, and actions.

Cosmos-Predict2.5~\cite{ali2025world} exemplifies a unified world model representation, learning a latent world model:
\begin{align*}
    z_{t+1} \sim p_\theta(z_{t+1} \mid z_t, a_t, c),
\end{align*}
where the conditioning signal $c$ may correspond to text, images, or video context. Its temporally causal tokenizer supports incremental state updates, while flow matching and post-training improve physical fidelity. This design unifies Text2World, Image2World, and Video2World generation, though its dynamics remain largely implicit and may therefore be difficult to interpret or control.

Other models introduce more explicit memory mechanisms, since encoding a consistent representation of the world for 2D and 3D scene generation requires an effective storage and retrieval strategy. To construct a 3D model of the world, MosaicMem~\cite{mosaicmem} stores and retrieves image patches lifted to 3D using estimated depth and camera poses, enabling spatially consistent retrieval and view-aligned composition. Marble world model~\cite{marbleworldblog2026,feifeiblog2026} instead represents the world through a continuous 3D radiance-like structure, instead of patches or pixels. This large set of semitransparent particles, known as 3D gaussian splats~\cite{kerbl20233d}, is amongst the highest-fidelity representations for scene generation.
In 2-dimensional scene generation, to establish a robust state space encoding for memory, V-JEPA~\cite{bardes2024revisiting} introduces a new space encoding for world representation that is trained solely using a feature prediction objective, completely bypassing the use of pretrained image encoders or other sources of supervision. 
Similarly, VideoREPA~\cite{zhang2026videorepa} refines the model's internal state representations during training by aligning token-level relations across spatial and temporal dimensions with a Token Relation Distillation (TRD) loss. 
In doing so, VideoREPA aligns the spatial and temporal token relations of generative diffusion models with robust representations from self-supervised foundation models. 

Beyond structural coherence, a major challenge for video world models is extending memory over long horizons. Gumbsch et al.~\cite{gumbsch2023learning} address this through hierarchical world models with adaptive temporal abstractions, where discrete latent dynamics identify meaningful change points and preserve stable contexts for long-horizon prediction. In this sense, memory is treated as a hierarchy of stable states and temporally abstract events, instead of simply a longer context window.
VideoWeave~\cite{durante2026videoweave} addresses long-context degradation from a data-centric perspective by splicing short captioned videos into longer synthetic sequences, forcing models to track persistent latent states across narrative transitions. Alternatively, Video-GPT~\cite{zhuang2025video} combines autoregressive memory with diffusion through \emph{Next Clip Diffusion}. This enforces strict causal clip-to-clip dependence using historical clean clips to maintain exceptionally stable internal memory over long-horizon video generation.

Several works focus specifically on converting bidirectional video generators into long-horizon autoregressive world models. Self-Forcing~\cite{selfforcing} addresses the train-test exposure bias by simulating inference conditions during training, performing autoregressive rollouts with rolling KV caching to condition future frames on the model's own self-generated past. Building on this, Causal Forcing~\cite{casualforcing} uses an autoregressive teacher for ODE distillation, encouraging strict causal history mapping.
Reward Forcing~\cite{rewardforcing} takes a structural approach to long-term memory via the \emph{EMA-Sink} mechanism, which maintains fixed-size context tokens for long-term coherence. 
Finally, some works achieve real-time speeds due to their effective and efficient world representation~\cite{hacohen2024ltx,helios}. Helios~\cite{helios} approaches long-horizon memory through a highly optimized deep compression flow. Rather than relying on standard anti-drifting memory heuristics like self-forcing, Helios heavily compresses the historical and noisy context and explicitly simulates drifting during training. Overall, these methods show that memory in video world models is not merely a matter of context length, but of designing latent states that remain stable and causal under repeated rollout.

% \arman{should add these two:
% Propose Temporal Hierarchies for action plannin / predicting future frames~\cite{gumbsch2023learning}.
% Order of Chaos: Real-world physical understanding from glitchy videos.}
% \textcolor{red}{Yes, these are good finds in my opinion.}
% Tooba: Done.

\subsection{World Model Prediction and Generation}\label{sec4:generation}

The generation phase evaluates whether a world model can use its internal state to predict coherent future trajectories. Within the CAT framework, this phase primarily involves \textbf{\textit{reasoning}}, which governs causal and temporal consistency, \textbf{\textit{imagination}}, which enables hypothetical rollout and counterfactual simulation, and \textbf{\textit{motivation}}, which introduces reward- or objective-driven pressure toward physically or task-relevant futures. Figures~\ref{fig:vwm_architecturesa}~\ref{fig:vwm_architecturesb} and~\ref{fig:vwm_architecturesc} shows the various architectures encountered in this review for 2D and 3D scene generation.

\subsubsection{\textbf{Reasoning.}}
Within video world models, \textbf{\textit{reasoning}} governs future-state generation under physical and temporal constraints. Architecturally, these mechanisms align with the paradigms illustrated in Figures~\ref{fig:vwm_architecturesa}~\ref{fig:vwm_architecturesb} and~\ref{fig:vwm_architecturesc}, each corresponding to a different instantiation of the world transition function \( W_\theta \).
Together, these paradigms define a spectrum, from strictly causal reasoning to globally optimized trajectory generation, with action-conditioned models bridging toward embodied control.

\paragraph{Autoregressive causal rollouts.}
Autoregressive world models (Figures~\ref{fig:vwm_architecturesa} and~\ref{fig:emb_architecturesb}) generate future states sequentially:
\begin{equation*}
    z_{t+1} = W_\theta(z_{\leq t}),
\end{equation*}
conditioning each prediction on previously generated states. This enforces causality through masked attention, recurrent state updates, or rolling context windows~\cite{zhu2025astra,casualforcing,durante2026videoweave}. Reasoning in this paradigm emerges as a chain of locally consistent transitions, though compounding error over long horizons remains a central challenge.

\paragraph{Promptable and action-conditioned world models.}
Action-conditioned models (Figure~\ref{fig:vwm_architecturesc}(A)) explicitly control transition dynamics:
\begin{equation*}
    z_{t+1} = W_\theta(z_t, a_t, c),
\end{equation*}
where $a_t$ denotes actions and $c$ may include language prompts, initial observations, goals, or other conditioning signals. These models most closely align with embodied reasoning because they simulate the visual consequences of interventions. EgoWM~\cite{bagchi2026walk}, for example, injects motor commands into pretrained video diffusion backbones for controllable egocentric prediction, while Dream2Flow~\cite{dharmarajan2025dream2flow} extracts 3D object flow from generated videos to ground predicted transitions in physically meaningful transformations.

\paragraph{Bidirectional and masked generation.}
Bidirectional or diffusion-based world models (Figure~\ref{fig:vwm_architecturesc}(B)) instead model the joint distribution over a trajectory:
\begin{equation*}
    z_{1:T} \sim p_\theta(z_{1:T}),
\end{equation*}
and iteratively refine the sequence through denoising or masked token prediction~\cite{peebles2023scalable,alonso2024diffusion,burchi2025accurate}. Here, reasoning is less strictly causal but more globally optimized, allowing the model to revise earlier and later states jointly, correcting inconsistencies across space and time during generation.

\paragraph{Latent actions and geometric constraints.}
Several works improve reasoning by structuring the latent transition space. Garrido et al.~\cite{garrido2026learning} jointly learn inverse and forward dynamics,
\begin{equation*}
    a_t \approx f^{-1}(z_t, z_{t+1}), \quad z_{t+1} = f(z_t, a_t),
\end{equation*}
allowing continuous latent actions to be discovered from uncurated video. ViewRope~\cite{xiang2026geometry} incorporates camera-ray geometry directly into attention, augmenting the transition function \( W_\theta \) with explicit spatial constraints to prevent geometric drift.

\paragraph{Hierarchical and Sim-to-Real reasoning.}
For embodied deployment, predicted trajectories must remain meaningful under domain shift. Hierarchical world models address this by decomposing prediction across temporal scales:
\begin{equation*}
    z_{t+1}^{(l)} = W_\theta^{(l)}(z_t^{(l)}, z_t^{(l+1)}),
\end{equation*}
where higher-level abstractions guide lower-level transitions. Gumbsch et al.~\cite{gumbsch2023learning} show that learning temporal abstractions around meaningful change points can support long-horizon consistency while preserving executability in real-world settings. Dream2Flow~\cite{dharmarajan2025dream2flow} further contributes by constraining transitions through 3D object flow, ensuring that imagined trajectories correspond to physically realizable actions.

\subsubsection{\textbf{Imagination.}}

Imagination refers to the ability of a world model to generate hypothetical futures that need not have been directly observed, but remain plausible enough to support planning, exploration, or policy learning. In video world models, imagination is realized through controllable rollout, \emph{i.e.}, the model simulates how a scene may evolve under actions, prompts, or latent interventions.

Genie~\cite{bruce2024genie} is an early example of this shift from passive video generation to interactive environment modeling. Trained on unlabeled internet videos, Genie demonstrates that frame-level controllability can emerge without ground-truth action labels or domain-specific robotic supervision.
Expanding on this, Dream2Flow~\cite{dharmarajan2025dream2flow} extracts 3D object flow from dreamed videos to serve as a modality-agnostic intermediate representation for open-world manipulation. 
By translating AI-generated video rollouts into executable trajectory tracking actions without task-specific demonstrations, it bridges the gap between hallucinated video generation and physical robotic control.

AdaWorld~\cite{gao2025adaworld} addresses the adaptation problem by pretraining world models with self-supervised continuous latent actions extracted from unlabeled video. At deployment time, real actions can be mapped into the learned latent action space, reducing the need to retrain for each new environment. Finally, EMERALD~\cite{burchi2025accurate} enhances DIAMOND~\cite{alonso2024diffusion} and Dreamer-style agents with MaskGIT-based~\cite{chang2022maskgit} parallel token prediction in a spatio-temporal latent representation. This allows jointly encoding spatial structure alongside temporal dynamics to improve accuracy and efficiency in model-based RL for sim-to-real transfer.
Astra~\cite{zhu2025astra} extends this with an autoregressive denoising framework that supports long video horizons sufficient for future prediction comparable to motor control, addressing the practical requirement of imagined rollouts for real-world environments.

\subsubsection{\textbf{Motivation.}}

Motivation introduces objective-driven constraints into video world models. Rather than generating only visually plausible sequences, motivated world models are guided by rewards, prediction errors, or physical constraints that encourage useful and realistic futures.

Le et al.~\cite{le2025gravity} explicitly align generated videos with physical laws by introducing verifiable, physics-grounded rewards during post-training. These rewards penalize violations of Newtonian dynamics, such as gravity and momentum, steering the model toward physically consistent rollouts.
Similarly, PhyWorld~\cite{zhao2026phyworld} grounds video generation to physical laws with direct policy optimization.
We also see VJEPA-2~\cite{assran2025v} use a latent world model with strong intuitive physics as an inference-time critic, scoring candidate trajectories and favoring predictions that better satisfy physical plausibility.

In terms of spatial and temporal awareness, Cambrian-S~\cite{yang2025cambrian} introduces a complementary form of motivation based on surprise. Its sensing mechanism uses prediction error to guide memory updates and event segmentation in long visual streams, connecting video world models to principles from active inference~\cite{friston2010free,friston2016active,pezzulo2024active,friston2025active}. Here motivation is not limited to external rewards, but can also arise from internal signals that identify when the world has changed in a meaningful way.

% \textcolor{blue}{
\subsection{Trends and limitations.}\label{sec:video:trends}
In summary, video world models are shifting from passive video synthesis toward structured, controllable world simulation. The field increasingly emphasizes geometry-aware perception, persistent latent memory, long-horizon rollout, and action-conditioned generation. Approaches such as ViewRope~\cite{xiang2026geometry}, StereoWorld~\cite{stereoworld}, MosaicMem~\cite{mosaicmem}, and TesserAct~\cite{zhen2025tesseract} illustrate this trend by lifting visual evidence into spatial representations to maintain viewpoint consistency.
% }

% \textcolor{blue}{
Yet key limitations remain: learned dynamics are often implicit and difficult to verify; long-horizon predictions are still subject to drift and compounding error; and 3D consistency often depends on additional depth, pose, or geometric priors. It is noteworthy that, motivation and metacognition also remain weakly developed. For example, Cambrian-S~\cite{yang2025cambrian} uses prediction error as a surprise-like signal, resembling Active Inference. Yet it does not explicitly formulate intrinsic motivation through Active Inference's exploration reward nor optimize a loss that models intrinsic latent dynamics. 
Instead, Cambrian-S relies on pixel-level prediction error as a proxy for surprise, omitting the information-gain term that drives exploratory behavior.
This reduction is valid if the next-latent predictive distribution is taken to be Gaussian with fixed variance, in which case the accuracy (negative log-likelihood) term of the free energy reduces to a scaled squared error between the predicted and observed means, i.e., MSE~\cite{friston2010free,bogacz2017tutorial,ho2020denoising}. 
A predictive objective that additionally captures epistemic uncertainty would more closely realize the full active-inference objective. Addressing these gaps will require video world models that couple scalable generative priors with explicit spatial structure, physically grounded objectives, intrinsic motivation, and mechanisms for evaluating and regulating their own predictions.

\section{Embodied World Models}\label{sec5}
\label{sec:embodied_wm}

\begin{figure}[t!]
    \centering
    \includegraphics[width=\linewidth]{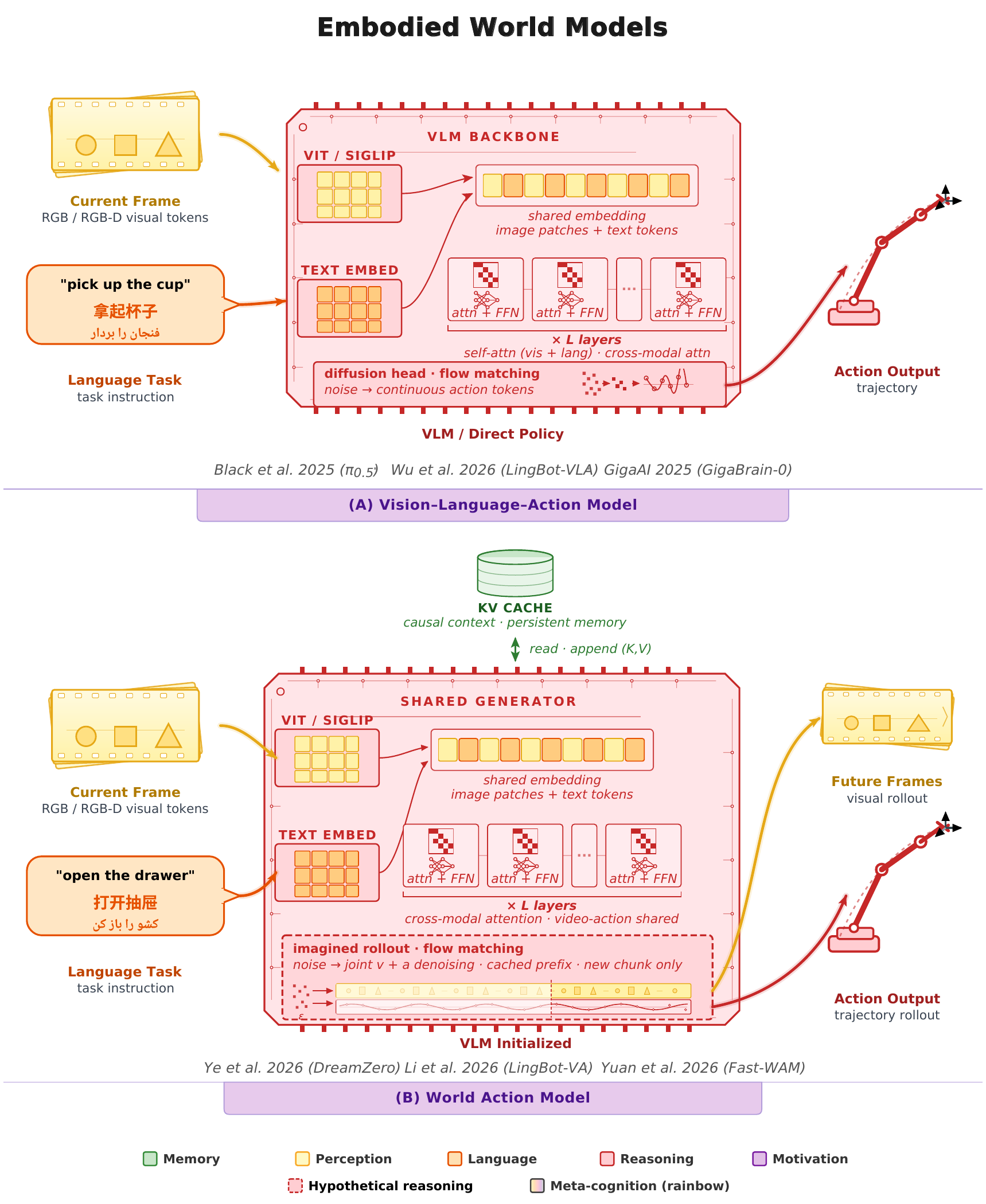}
    \vspace{0.1cm}
    \hrule
    \vspace{0.1cm}
    \caption{
   Above are the typical architectures encountered when reviewing embodied world models.}
    \label{fig:emb_architecturesa}
\end{figure}

\begin{figure}[t!]
    \centering
    \includegraphics[width=\linewidth]{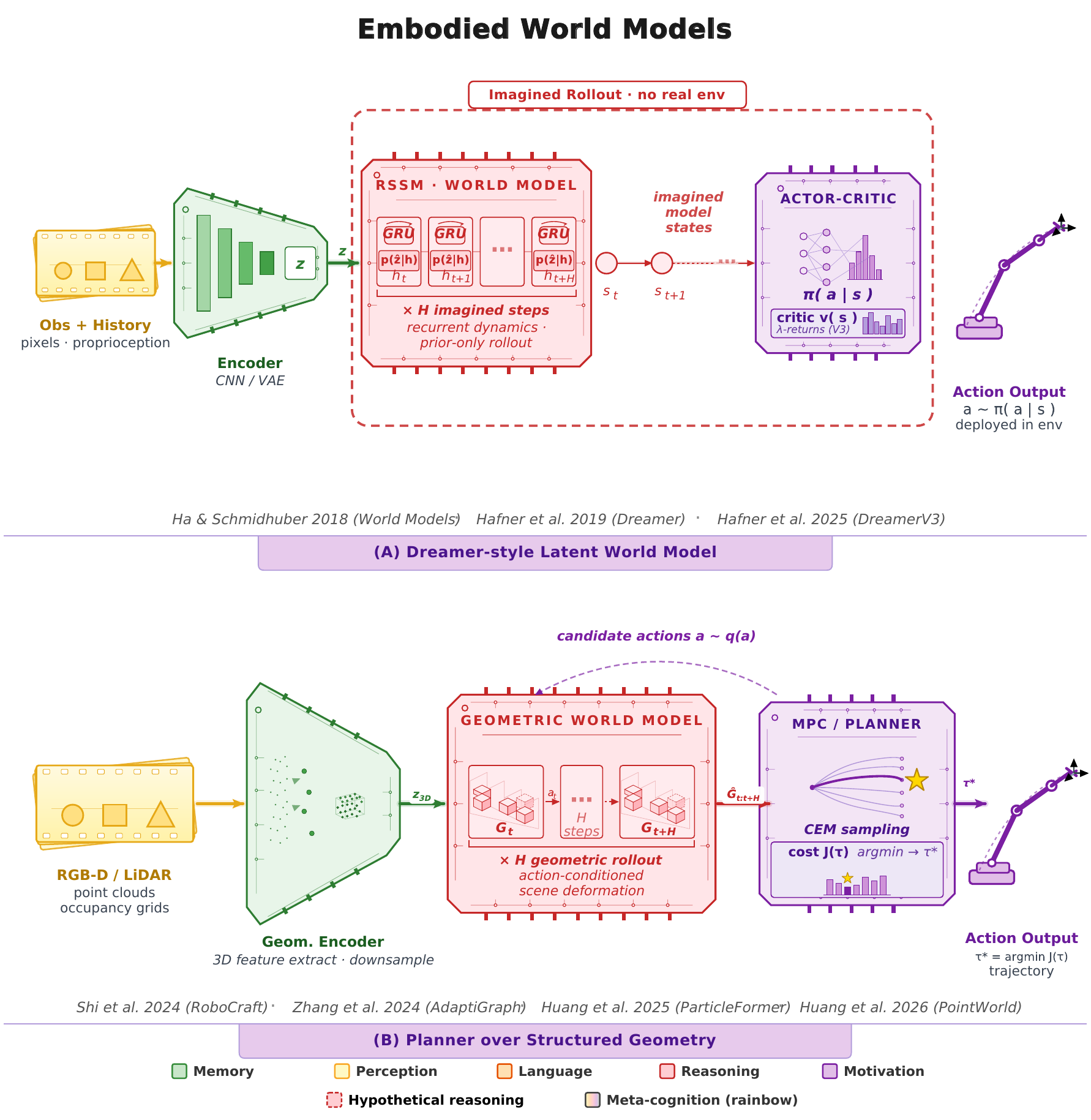}
    \vspace{-0.1cm}
    \hrule
    \vspace{-0.1cm}
    \caption{
   Above are the typical architectures encountered when reviewing embodied world models.}
    \label{fig:emb_architecturesb}
\end{figure}

Embodied world models not only demand visual understanding and linguistic reasoning, but also perceive, act, and anticipate how their actions reshape the physical world. 
This physical grounding constraint distinguishes embodied world models from their video generation counterparts (Section~\ref{sec:video_wm}). 
A video world model succeeds when its generated frames are photorealistic and temporally coherent. 
However, an embodied world model succeeds when its predictions are physically accurate enough to guide a body with mass, kinematics, and contact surfaces toward successful task execution. 
These demands on embodied world models heighten the demands for a Unified World Model, as seen in Figure~\ref{fig:catbrain}, capable of emulating a broad range of cognitive functions.
We explore embodied world models in robotics~\cite{kim2026cosmos,ye2026world,yuan2026fastwamworldactionmodels,team2025gigabrain,ye2026gigaworld,bi2026motus,intelligence2025pi_,wu2026pragmatic,li2026causal,sun2026vla,wang2026vagen,BeingH,luo2026being,huang2026pointworld,lu2024manigaussian,shang2026roboscape,huang2026enerverse,zhu2025unified,wu2024unleashing,cheang2024gr,du2023learning,black2024zero,zhu2025irasim,huang2025ladi,guo2026flowdreamer,seo2023masked,zhang2024pivot,sekar2020planning,wu2023daydreamer,barcellona2025dream,zhou2024robodreamer,mazzaglia2024genrl,jang2025dreamgen,assran2025v,cen2025worldvla,nahrendra2023dreamwaq,wang2025dreamnav,kahn2021badgr,shah2023vint,sridhar2024nomad,wang2026eva}, 
navigation with exploration~\cite{kahn2021badgr,shah2023vint,sridhar2024nomad}, 
and autonomous driving~\cite{zheng2024occworld,zhang2024copilot4d,zhang2024bevworld,dang2026sparseworld,zhang2025epona,liang2026lidarcrafter,chen2025drivinggpt,zheng2024doe,li2025enhancing,chang2021mitigating,bheemaiah2025knowledge,zhao2026forecasting,wang2025adawm,li2024think2drive,yang2026raw2drive,nachkov2025dream,gao2024dream,hu2023gaia,russell2025gaia2,zeng2025rethinking,hao2025neural,chen2025large,qu2025vl,huang2024safedreamer,jiang2025irl,khanzada2025indrive,bosio2025rdar,wang2026latent}.

\textbf{\textit{Perception}} must encode contact geometry and six-degree-of-freedom pose, not merely visual appearance; \textbf{\textit{Memory}} may need to maintain a persistent, spatially-grounded model of the physical world across interactions, not merely temporal coherence across frames. \textbf{\textit{Reasoning}} may need to capture physical causality and force propagation rather than narrative causality; and \textbf{\textit{Imagination}} must generate action conditioned futures that are physically executable, not merely visually plausible. We explore these novelties following the CAT structure of our unified framework: world representation in Sec.~\ref{sec:embodied:world-rep} and world generation in Sec.~\ref{sec:embodied:world-gen}.
Sec.~\ref{sec:embodied:trends} summarizes key trends and limitations of embodied world models.

\begingroup
\centering
\scriptsize
\setlength{\tabcolsep}{3.0pt}
\renewcommand{\arraystretch}{1.05}
\begin{longtable}{l c c c c c c c c c p{4cm}}
\caption{Embodied world model works in Section~5 mapped to Cognitive Architecture Theory (CAT) functions (Memory, Perception, Language, Reasoning, Imagination, Motivation, Metacognition). A check (\cmark) indicates a clear contribution.}
\label{tab:embodied_wm} \\
\toprule
\textbf{Work} & \textbf{Year} & \textbf{App.} & \rotatebox{90}{\textbf{Mem.}} & \rotatebox{90}{\textbf{Perc.}} & \rotatebox{90}{\textbf{Lang.}} & \rotatebox{90}{\textbf{Reas.}} & \rotatebox{90}{\textbf{Imag.}} & \rotatebox{90}{\textbf{Moti.}} & \rotatebox{90}{\textbf{Meta.}} & \textbf{Description} \\
\midrule
\endfirsthead
\multicolumn{11}{c}{\tablename\ \thetable{} -- \textit{continued from previous page}} \\
\toprule
\textbf{Work} & \textbf{Year} & \textbf{App.} & \rotatebox{90}{\textbf{Mem.}} & \rotatebox{90}{\textbf{Perc.}} & \rotatebox{90}{\textbf{Lang.}} & \rotatebox{90}{\textbf{Reas.}} & \rotatebox{90}{\textbf{Imag.}} & \rotatebox{90}{\textbf{Moti.}} & \rotatebox{90}{\textbf{Meta.}} & \textbf{Description} \\
\midrule
\endhead
\midrule
\multicolumn{11}{r}{\textit{continued on next page}} \\
\endfoot
\bottomrule
\endlastfoot

%% -------------------------------------------------------
%% FOUNDATIONAL / PLATFORM MODELS
%% -------------------------------------------------------
% Cosmos-Predict2.5~\cite{ali2025world} & 2025 & MOVE & \cmark &  & \cmark &  &  & \cmark &  & unified flow-matching video generator trained on clips for Physical AI simulation, data augmentation, and policy evaluation. \\

% \midrule
\multicolumn{11}{l}{\textit{Foundational / Platform Models}} \\
\midrule
Being-H0.7~\cite{BeingH} & 2026 & Robot & \cmark & & & \cmark & & & & Action prediction occurs in an action-centric spatio-temporal latent space, not pixel-space, eliminating predictive video \\
Cosmos-Policy~\cite{kim2026cosmos} & 2026 & Robot & \cmark &  &  & \cmark & \cmark & \cmark &  & Post-trains Cosmos-Predict-2 for visuomotor robot control and planning \\
DreamZero ~\cite{ye2026world} & 2026 & Robot & \cmark &  &  & \cmark & \cmark &  &  & World Action Model jointly predicting video and actions; zero-shot policy from video pretraining \\
Fast-WAM~\cite{yuan2026fastwamworldactionmodels} & 2026 & Robot &  &  &  & \cmark & \cmark &  &  & Decouples action prediction from video generation at inference for faster WAM deployment \\
GigaBrain-0~\cite{team2025gigabrain} & 2025 & Robot &  & \cmark &  & \cmark & \cmark &  &  & VLA trained on world-model-generated data with RGBD input and embodied chain-of-thought \\
Motus~\cite{bi2026motus} & 2025 & Robot & \cmark &  &  & \cmark & \cmark &  &  & Unified latent action world model via Mixture-of-Transformers; optical flow as embodiment-agnostic action \\
%% -------------------------------------------------------
%% VLA / LANGUAGE-GROUNDED EMBODIED MODELS
%% -------------------------------------------------------
\midrule
\multicolumn{11}{l}{\textit{VLA / Language-Grounded Embodied Models}} \\
\midrule
$\pi_{0.5}$~\cite{intelligence2025pi_} & 2025 & Robot &  & \cmark & \cmark & \cmark &  &  &  & Flow-matching VLA decomposing language into causally grounded multi-step physical plans \\
LingBot-VLA~\cite{wu2026pragmatic} & 2026 & Robot & & \cmark & \cmark & \cmark &  &  &  & Cross-morphology VLA on 20{,}000h bimanual data with geometry-aware depth distillation \\
LingBot-VA~\cite{li2026causal} & 2026 & Robot & \cmark &  & \cmark & \cmark & \cmark &  &  & Interleaves video and action tokens for joint imagination and action decoding \\
VLA-JEPA~\cite{sun2026vla} & 2026 & Robot & \cmark &  & \cmark & \cmark & \cmark & \cmark  &  & Leakage-free JEPA grounding visual encoder in action-relevant dynamics \\
VAGEN~\cite{wang2026vagen} & 2025 & Robot &  &  & \cmark & \cmark &  & \cmark &  & RL-structured world model reasoning into state estimation and transition modeling for VLM agents \\
Being-H0.5~\cite{luo2026being} & 2026 & Robot &  & \cmark & \cmark & \cmark &  & \cmark &  & A foundational VLA for robust cross-embodiment generalization across diverse robotic platforms \\
%% -------------------------------------------------------
%% MANIPULATION World ModelS
%% -------------------------------------------------------
\midrule
\multicolumn{11}{l}{\textit{World Models for Manipulation}} \\
\midrule
Flash-WAM~\cite{akbari2026flash} & 2026 & Robot & \cmark & \cmark & & & \cmark & & & Step-distillation framework for achieving inference latency under 350 ms.\\
PointWorld~\cite{huang2026pointworld} & 2026 & Robot & \cmark & \cmark &  & \cmark & \cmark &  &  & Unifies state and action as 3D point flows with MPC over imagined scene deformations \\
ManiGaussian~\cite{lu2024manigaussian} & 2024 & Robot &  & \cmark &  & \cmark & \cmark &  &  & Dynamic 3DGS world model predicting future Gaussian scenes under action for manipulation \\
RoboScape~\cite{shang2026roboscape} & 2025 & Robot &  & \cmark &  & \cmark & \cmark &  &  & Physics-informed world model jointly learning video, depth, and keypoint dynamics \\
EnerVerse-AC~\cite{huang2026enerverse} & 2025 & Robot & \cmark & \cmark &  & \cmark & \cmark &  &  & Chunk-wise autoregressive video diffusion with sparse memory and 4DGS for action-conditioned prediction \\
UWM~\cite{zhu2025unified} & 2025 & Robot & \cmark &  &  & \cmark & \cmark &  &  & Couples video and action diffusion in one transformer; pretrained on video-only and video+action data \\
GR-1~\cite{wu2024unleashing} & 2024 & Robot & \cmark & \cmark & \cmark &  & \cmark &  &  & GPT transformer pretrained on 800K Ego4D clips jointly predicting actions and future frames \\
GR-2~\cite{cheang2024gr} & 2024 & Robot & \cmark & \cmark & \cmark & \cmark & \cmark &  &  & Scaled video-language-action model (719M) achieving 97.7\% success across 100+ real tasks \\
UniPi~\cite{du2023learning} & 2023 & Robot &  &  & \cmark &  & \cmark &  &  & Text-conditioned video diffusion as policy; extracts actions via inverse dynamics \\
SuSIE~\cite{black2024zero} & 2024 & Robot &  & \cmark & \cmark &  & \cmark & \cmark &  & Image-editing diffusion synthesizing subgoal images for goal-conditioned manipulation policy \\
IRASim~\cite{zhu2025irasim} & 2024 & Robot &  & \cmark &  &  & \cmark & \cmark &  & Diffusion transformer with frame-level action conditioning for manipulation simulation \\
LaDi-WM~\cite{huang2025ladi} & 2025 & Robot &  & \cmark &  & \cmark & \cmark & \cmark &  & Predicts latent state evolution via diffusion; more generalizable than pixel-level prediction \\
FlowDreamer~\cite{guo2026flowdreamer} & 2025 & Robot &  & \cmark &  & \cmark & \cmark &  &  & RGB-D world model using optical flow as explicit physically interpretable motion supervision \\
MWM~\cite{seo2023masked} & 2023 & Robot & \cmark & \cmark &  & \cmark & \cmark & \cmark &  & Decouples MAE visual representation from RSSM dynamics; 81.7\% on Meta-World tasks \\
PIVOT-R~\cite{zhang2024pivot} & 2024 & Robot & \cmark & \cmark & \cmark & \cmark & &  &  & Waypoint-aware world model focusing prediction on task-relevant key states \\
Plan2Explore~\cite{sekar2020planning} & 2020 & Robot &  &  &  &  & \cmark & \cmark &  & world model exploration via ensemble disagreement for zero-shot task adaptation \\
DayDreamer~\cite{wu2023daydreamer} & 2022 & Robot &  &  &  & \cmark & \cmark & \cmark &  & Dreamer-style world model learning directly on physical robots from sparse rewards \\
Dream to Manipulate~\cite{barcellona2025dream} & 2024 & Robot &  & \cmark &  &  & \cmark &  &  & Compositional 3DGS with object decomposition for imagination-based imitation learning \\
RoboDreamer~\cite{zhou2024robodreamer} & 2024 & Robot &  & \cmark & \cmark &  & \cmark &  &  & Compositional diffusion world model factorizing language instructions into task primitives \\
GenRL~\cite{mazzaglia2024genrl} & 2024 & Robot &  & \cmark & \cmark &  & \cmark & \cmark &  & Foundation world models for generalization in embodied RL via multimodal priors \\
V-JEPA 2~\cite{assran2025v} & 2025 & Robot &  & \cmark &  &  &  & \cmark &  & Representation-space prediction on 1M video hours; zero-shot planning via MPC \\
WorldVLA~\cite{cen2025worldvla} & 2025 & Robot & \cmark &  & \cmark & \cmark & \cmark &  &  & Autoregressive action world model unifying video prediction and VLA action generation \\
AdaptiGraph~\cite{zhang2024adaptigraph} & 2024 & Robot & \cmark & \cmark & & \cmark & & & & Graph Neural Network for world represented as particles for general purpose manipulation \\
RoboCraft~\cite{shi2024robocraft} & 2024 & Robot & \cmark & \cmark & & \cmark & & \cmark & & Graph Neural Network for world represented as particles for elastic materials \\
Particle Former~\cite{huang2025particleformer} & 2025 & Robot & \cmark & \cmark &  & \cmark &  & \cmark &  & Transformer-based reasoning from 3d point cloud scene observations \\
%% -------------------------------------------------------
%% LOCOMOTION / NAVIGATION World ModelS
%% -------------------------------------------------------
\midrule
\multicolumn{11}{l}{\textit{Locomotion / Navigation World Models}} \\
\midrule
DreamWaQ~\cite{nahrendra2023dreamwaq} & 2023 & Robot &  & \cmark &  & \cmark & \cmark & \cmark &  & Implicit terrain imagination from proprioception for quadrupeds \\
DreamerNav~\cite{wang2025dreamnav} & 2025 & Robot & & \cmark & \cmark & \cmark & \cmark &  &  & DreamerV3 for quadruped navigation with depth and occupancy; zero-shot sim-to-real \\
BADGR~\cite{kahn2021badgr} & 2021 & Robot &  & \cmark &  & \cmark &  &  &  & Self-supervised terrain affordance learning for outdoor navigation via MPC \\
ViNT~\cite{shah2023vint} & 2023 & Robot & \cmark & \cmark &  &  & \cmark & \cmark &  & Visual navigation model with diffusion subgoal proposals across robot platforms \\
NoMaD~\cite{sridhar2024nomad} & 2024 & Robot &  & \cmark &  & \cmark & \cmark &  &  & Unifies goal navigation and exploration with diffusion policy and goal-masking \\
EVA~\cite{wang2026eva} & 2026 & Robot &  &  &  &  &  & \cmark &  & Trains a video world model with an inverse dynamics reward signal aligns to physical constraints in world dynamics. \\
VLM-Safe~\cite{qu2025vl} & 2025 & Robot &  &  &  &  & \cmark & \cmark &  & VLM-guided safety rewards steering imagined rollouts for constrained AV policy inspired by human cognition \\
GeNIE~\cite{wang2025genie} & 2025 & AD & \cmark &  &  & \cmark &  &  &  & Traversable state prediction enables generalizing across a wide range of real-world environments. \\
%% -------------------------------------------------------
%% AUTONOMOUS DRIVING World ModelS
%% -------------------------------------------------------

\midrule
\multicolumn{11}{l}{\textit{Autonomous Driving World Models}} \\
\midrule
OccWorld~\cite{zheng2024occworld} & 2023 & AD & \cmark & \cmark &  & \cmark & &  &  & VQVAE tokenizer with GPT transformer for joint 3D occupancy and ego trajectory forecasting \\
Copilot4D~\cite{zhang2024copilot4d} & 2023 & AD &  & \cmark &  & \cmark &  &  &  & Discrete diffusion over BEV LiDAR tokens; 65\% Chamfer distance reduction \\
BEVWorld~\cite{zhang2024bevworld} & 2024 & AD & \cmark & \cmark &  & \cmark &  &  &  & Multimodal tokenizer fusing camera and LiDAR into unified BEV latent for forecasting \\
SparseWorld~\cite{dang2026sparseworld} & 2025 & AD & \cmark & \cmark &  & \cmark &  &  &  & Sparse dynamic queries modulated by ego-vehicle state for adaptive scene memory \\
Epona~\cite{zhang2025epona} & 2025 & AD & \cmark &  &  & \cmark & \cmark &  &  & Decouples temporal memory from spatial generation via causal transformer and twin diffusion \\
LiDARCrafter~\cite{liang2026lidarcrafter} & 2025 & AD & \cmark & \cmark & \cmark & \cmark &  &  &  & Language-conditioned tri-branch diffusion for 4D LiDAR scene generation \\
DrivingGPT~\cite{chen2025drivinggpt} & 2024 & AD & \cmark & \cmark & \cmark & \cmark &  &  &  & Interleaved image--action tokens unifying world modeling and planning as next-token prediction \\
DOE-1~\cite{zheng2024doe} & 2024 & AD & \cmark & \cmark & \cmark & \cmark &  &  &  & Closed-loop end-to-end AV model using free-form text as perceptual interface \\
LAW~\cite{li2025enhancing} & 2024 & AD &  & \cmark &  & \cmark & \cmark &  &  & Self-supervised latent prediction of future scene features from observations and ego trajectories \\
MILO~\cite{chang2021mitigating} & 2021 & AD &  &  &  & \cmark &  & \cmark &  & Model-based imitation learning mitigating covariate shift via offline data \\
KG-based WM~\cite{bheemaiah2025knowledge} & 2025 & AD & \cmark & \cmark &  &  &  &  &  & Knowledge graphs with sensor data for material-aware obstacle reasoning in AVs \\
PWM~\cite{zhao2026forecasting} & 2025 & AD &  &  &  & \cmark &  & \cmark &  & Collaborative state-action prediction for anticipatory planning in AVs \\
AdaWM~\cite{wang2025adawm} & 2025 & AD &  &  &  & \cmark &  &  &  & Adaptive world model planning with dynamic model selection for autonomous driving \\
Think2Drive~\cite{li2024think2drive} & 2024 & AD &  &  &  &  & \cmark & \cmark &  & Efficient RL via latent world model imagination in CARLA-v2 \\
Raw2Drive~\cite{yang2026raw2drive} & 2025 & AD &  &  &  &  & \cmark & \cmark &  & End-to-end AV aligning RL policy with imagination from raw sensors \\
Dream to Drive~\cite{nachkov2025dream} & 2025 & AD & \cmark &  &  & \cmark & \cmark & \cmark &  & Analytic world model for dreamer-style vehicle control without environment interaction \\
Dream2Drive~\cite{gao2024dream} & 2024 & AD &  & \cmark &  &  & \cmark & \cmark &  & RL in predictive world model imagination with intention-aware latent states \\
Large Video Planner~\cite{chen2025large} & 2025 & AD & &  &  & \cmark &  &  &  & Foundation-scale video model for zero-shot robot video plans to actions \\
SafeDreamer~\cite{huang2024safedreamer} & 2023 & AD &  &  &  &  & \cmark & \cmark &  & Dreamer safe RL pairing imagined trajectories with safety estimations \\
IRL-VLA~\cite{jiang2025irl} & 2025 & AD &  &  & \cmark &  &  & \cmark &  & Inverse RL reward world model for efficient closed-loop reward computation in VLA training \\
InDRiVE~\cite{khanzada2025indrive} & 2025 & AD &  &  &  &  &  & \cmark &  & Intrinsic disagreement reward in Dreamer MBRL for curiosity-driven vehicle exploration \\
RDAR~\cite{bosio2025rdar} & 2025 & AD &  &  &  & \cmark &  & \cmark &  & Reward-driven relevance estimation for safety-critical agents in AV planning \\
% NewtonRewards~\cite{le2025gravity} & 2025 & AD & \cmark & \cmark &  & \cmark &  & \cmark &  & Physics-grounded reward functions penalizing violations of Newtonian laws in video generation \\
Latent-WAM~\cite{wang2026latent} & 2026 & AD & \cmark & \cmark  &  & \cmark & & &  & Uses causal transformer to jointly learn visual and motion dynamics from geometry conditioned world representations \\
\midrule
\multicolumn{11}{l}{\textit{Synthetic Training Data}} \\
\midrule
GAIA-1~\cite{hu2023gaia} & 2023 & AD & \cmark &  & & \cmark & \cmark &  &  & Generative world model with discrete latent space for scalable synthetic data  \\
GAIA-2~\cite{russell2025gaia2} & 2025 & AD & \cmark &  &  & \cmark & \cmark &  &  & Controllable multi-view generative world model with continuous latent space for scalable synthetic data \\
Dream4Drive~\cite{zeng2025rethinking} & 2025 & AD &  &  &  &  & \cmark &  &  & Repurposes world model imagination for scalable synthetic data \\
MoSim~\cite{hao2025neural} & 2025 & AD &  &  &  & \cmark & \cmark & \cmark &  & Motion-grounded simulation generating diverse controllable traffic scenarios \\
DreamGen~\cite{jang2025dreamgen} & 2025 & Robot &  & \cmark &  &  & \cmark &  &  & Fine-tunes Cosmos-Predict-2.5 as synthetic data engine for policy training \\
\end{longtable}
\endgroup

\subsection{World Model Representation } \label{sec:embodied:world-rep}
\subsubsection{\textbf{Perception}}
In embodied world models perception must encode the physical structure of the scene rather than visual appearance alone. This makes 3D occupancy a natural representation for world models as it is expressive, efficient, and versatile across both vision and LiDAR inputs~\cite{zheng2024occworld}. OccWorld operationalizes this by learning in 3D semantic occupancy space, using a VQVAE-based scene tokenizer to produce discrete scene tokens that jointly forecast future occupancy and ego trajectory through a GPT-like spatial-temporal transformer, all without requiring instance or map annotations~\cite{zheng2024occworld}. Building on the same tokenize-then-predict philosophy, Copilot4D~\cite{zhang2024copilot4d} further addresses the scalability of this perceptual pipeline by applying discrete diffusion over BEV tokens derived from raw LiDAR point clouds, reducing prior state-of-the-art Chamfer distance by over 65\% for one-second prediction across multiple benchmarks~\cite{zhang2024copilot4d}.

\subsubsection{\textbf{Memory}}
A central challenge in embodied world models is maintaining a compact yet sufficiently rich state space encoding of the scene across time. BEVWorld addresses this by compressing heterogeneous multimodal inputs including camera imagery and LiDAR point clouds into a unified Bird's Eye View latent space through a self-supervised multimodal tokenizer, enabling temporally consistent future scene forecasting via a latent BEV sequence diffusion model conditioned on action tokens~\cite{zhang2024bevworld}. While BEVWorld~\cite{zhang2024bevworld} establishes a shared spatial memory across modalities, it relies on static grid-based representations that struggle to adapt to the dynamic and continuous nature of real driving environments. SparseWorld addresses this limitation directly by replacing fixed grid embeddings with sparse and dynamic queries modulated by the ego vehicle's state, allowing the memory encoding to scale its perception range with vehicle speed and adapt to foreground object dynamics rather than treating all voxels uniformly~\cite{dang2026sparseworld}. 

Epona~\cite{zhang2025epona} takes a complementary approach to the memory problem by identifying that conventional video diffusion models entangle temporal memory with spatial generation, leading to error accumulation in long-horizon rollouts. By decoupling the two through a GPT-style causal transformer that handles temporal context in compressed latent space separately from twin diffusion transformers that handle spatial rendering and trajectory generation, Epona achieves stable long-duration prediction with a 7.4\% FVD improvement over prior works~\cite{zhang2025epona}.
Further advancing perception, LaDi-WM~\cite{huang2025ladi} finds that predicting the evolution of the latent space is easier to learn and substantially more generalizable than directly predicting pixel-level images in diffusion models. 

\subsubsection{\textbf{Language}}
In the CAT framework, language serves as a semantic and symbolic encoding that connects human intent to world state, and in embodied driving this role becomes particularly concrete. LiDARCrafter~\cite{liang2026lidarcrafter} demonstrates this most directly by using free-form natural language instructions as the entry point for 4D LiDAR world modeling, parsing text into ego-centric scene graphs that condition a tri-branch diffusion network to generate object structures, motion trajectories, and geometry, with an autoregressive module then extending the result into temporally coherent LiDAR sequences~\cite{liang2026lidarcrafter}. Where LiDARCrafter~\cite{liang2026lidarcrafter} uses language to control a geometric representation, DrivingGPT~\cite{chen2025drivinggpt} uses it to unify the entire driving pipeline by constructing a multimodal driving language from interleaved image and action tokens, treating world modeling and trajectory planning as a single next-token prediction problem over this shared symbolic vocabulary~\cite{chen2025drivinggpt}. DOE-1~\cite{zheng2024doe} takes this unification further by closing the loop entirely, using free-form text scene descriptions as the perceptual interface and autoregressively generating perception, prediction, and planning tokens within one multimodal transformer, achieving the first closed-loop end-to-end autonomous driving model under this paradigm~\cite{zheng2024doe}.

\subsection{World Model Generation} \label{sec:embodied:world-gen}

\subsubsection{\textbf{Reasoning.}}

Architectures for embodied world models primarily differ along the following axes (with some sampled in Fig.~\ref{fig:emb_architecturesa} and~\ref{fig:emb_architecturesb}): 

\paragraph{Geometric world models.}
PointWorld~\cite{huang2026pointworld} and other geometric world models~\cite{huang2026pointworld,lu2024manigaussian,zhang2024adaptigraph,shi2024robocraft,huang2025particleformer,shang2026roboscape,zheng2024occworld,zhang2024bevworld,liang2026lidarcrafter} represent the opposite extreme, grounding reasoning directly in 3D structure by predicting scene flow
\(
\Delta \mathbf{x} = f_\theta(\mathbf{x}, a_t)
\)
over point clouds. This enforces physically meaningful state transitions and enables cross-embodiment generalization without task-specific heads. However, it depends on strong geometric supervision and may be less flexible for abstract or semantic tasks.

\paragraph{Language-conditioned planning.}
$\pi_{0.5}$~\cite{intelligence2025pi_} treats reasoning as long-horizon planning conditioned on language, implicitly learning
\begin{align*}
    a_{1:T} \sim \pi_\theta(a_{1:T} \mid z_0, \ell)
\end{align*}
where $\ell$ encodes task structure. This enables coherent multi-step behavior across diverse environments, but places significant burden on the latent representation to maintain causal consistency over long horizons.
Vision-Language-Action (VLAs) models~\cite{intelligence2025pi_,wu2026pragmatic,li2026causal,sun2026vla,wang2026vagen,luo2026being,cen2025worldvla,jiang2025irl} excel in reactive and instruction-following settings but rely on either external planners or hybridization with World Action Models (WAMs) when long-horizon physical reasoning or counterfactual simulation is required.
A central limitation of WAMs is the number of denoising steps, which can be as high as 10 in LingBot-VA, leading to higher latencies.

\paragraph{Autoregressive world-action models.}
Because VLAs are reactive and under-supervised and map action directly, with no explicit model of future researchers turn to World Action Models (WAMs)~\cite{chang2021mitigating,bheemaiah2025knowledge,zhao2026forecasting,hao2025neural,ye2026world,yuan2026fastwamworldactionmodels,bi2026motus,li2026causal,cen2025worldvla,wang2026latent,ye2026gigaworld,BeingH}.
Two such works, LingBot-VA~\cite{li2026causal} and DreamZero~\cite{ye2026world} unifies imagination and control by modeling joint sequences
\begin{align*}
    (z_{t+1}, a_t) \sim p_\theta(z_{t+1}, a_t \mid z_{\le t})
\end{align*}
with causal or block-causal attention. These approaches improve temporal consistency and enable efficient rollout, but remain sensitive to representation quality and training stability.

\paragraph{Representation-driven reasoning.}
VLA-JEPA~\cite{sun2026vla} highlights that reasoning quality depends critically on the learned state space, enforcing dynamics-consistent representations $z_t = \phi(o_t)$ that are invariant to irrelevant visual variation. This improves generalization, but shifts complexity into representation learning as recent works show~\cite{seo2023masked,assran2025v,sun2026vla,huang2025ladi,wang2025adawm,li2025enhancing}.

\paragraph{Planning-centric extensions.}
Across paradigms, there is a convergence toward explicit planning over learned dynamics~\cite{huang2026pointworld,sekar2020planning,wang2025adawm,li2025enhancing,gao2024dream,qu2025vl,huang2024safedreamer} to predict actions in the form,
\begin{align*}
    a_t^* = \arg\max_{a_{t:t+H}} \mathbb{E}\left[\sum_{k=0}^{H} r(z_{t+k})\right],
\end{align*}
as seen in model-predictive control (PointWorld~\cite{huang2026pointworld}), self-supervised world models (LAW~\cite{li2025enhancing}), and latent RL pipelines~\cite{wang2025adawm}. These approaches improve controllability and robustness, but depend on accurate forward models.

Figures~\ref{fig:emb_architecturesa} and ~\ref{fig:emb_architecturesb} can thus be interpreted as a spectrum: generative models offer scalability and multimodal unification; geometric models provide strong physical grounding; and planning-based approaches enable controllable long-horizon reasoning. Most recent systems hybridize these axes, suggesting that effective reasoning emerges from combining expressive latent models with structured representations and explicit planning.

\subsubsection{\textbf{Imagining.}}

In embodied world models, imagination manifests as dreaming~\cite{wu2023daydreamer,nahrendra2023dreamwaq,wang2025dreamnav,gao2024dream,nachkov2025dream,huang2024safedreamer,khanzada2025indrive}, as well as action-conditioned hypothetical reasoning, where agents simulate future trajectories in latent space to guide policy learning and planning~\cite{jang2025dreamgen,chen2025large,hu2023gaia,russell2025gaia2,zeng2025rethinking,nachkov2025dream}. 
Works such as Think2Drive~\cite{li2024think2drive} and Raw2Drive~\cite{yang2026raw2drive} leverage world models to generate imagined rollouts for training driving policies, while autoregressive approaches model multiple probabilistic futures to reason under uncertainty~\cite{xiao2025learning}. 
Dream2Drive~\cite{gao2024dream} further demonstrates this paradigm by operating within a learned imagination space (PIWM), using intention-aware latent states to evaluate candidate trajectories for urban navigation~\cite{gao2024dream}. 
More recent systems integrate imagination directly into the training loop, enabling policies to be optimized through interaction with an internal simulator rather than the real environment~\cite{goff2025learning}.

Beyond policy learning, imagination enables safety-aware planning and scalable data generation. 
world models can pair imagined roll outs with safety estimates to guide actor–critic optimization under additional constraints, such as VLM-based safety signals~\cite{qu2025vl}. 
At larger scale, foundation video models generate zero-shot trajectory plans from internet-scale data, which can be converted into executable robot actions~\cite{chen2025large}, while broader dreaming pipelines~\cite{nachkov2025dream}, including Cosmos-Drive-Dreams~\cite{ren2025cosmos}, GAIA-1, GAIA-2~\cite{hu2023gaia,russell2025gaia2}, and Dream4Drive~\cite{zeng2025rethinking} use synthetic rollouts as training data. 
Benchmarks such as WorldLens evaluate the physical fidelity of these imagined trajectories~\cite{liang2026worldlens}, reinforcing imagination as a core mechanism for bridging perception, reasoning, and action in embodied settings.

\subsubsection{\textbf{Motivation.}}

Motivation in embodied world models defines how agents evaluate imagined trajectories during learning and control~\cite{huang2026pointworld,jiang2025irl,khanzada2025indrive,sekar2020planning,wang2025adawm,bosio2025rdar,li2025enhancing,gao2024dream,shi2024robocraft,huang2025particleformer,qu2025vl,huang2024safedreamer}. 
IRL-VLA~\cite{jiang2025irl} learns a lightweight reward world model via Inverse Reinforcement Learning for efficient closed-loop optimization~\cite{jiang2025irl}, while InDRiVE~\cite{khanzada2025indrive} leverages intrinsic disagreement-based rewards within a Dreamer-style MBRL framework to drive exploration~\cite{khanzada2025indrive}. 
Additional approaches refine reward signals for task relevance in autonomous driving settings~\cite{bosio2025rdar}, reflecting a shift toward learned and uncertainty-aware objectives over hand-crafted rewards.
Safe-Dreamer motivates models to conform to safety criteria by incorporating Lagrangian-based methods into planning~\cite{huang2024safedreamer}.

\subsection{Trends and Limitations} \label{sec:embodied:trends}
% \textcolor{blue}{
Typical implementations of World Action Models though capable of robust reasoning lack the efficiency of direct policy networks such as VLAs.
Some World Action Model works~\cite{yuan2026fastwamworldactionmodels,ye2026world,li2026causal} achieve efficiency gains through the practical implementation of \textbf{\textit{memory}} mechanisms such as KV-cache.
KV-cache allows the reuse of previously generated transformer activations allowing the models to avoid recomputation across subsequent forward passes as depicted in Fig.~\ref{fig:emb_architecturesa}
% }

% \textcolor{blue}{
Another recent work combines the utility of both WAMs and VLAs in what they call a latent World Action Model or BeingH-0.7~\cite{BeingH}.
BeingH-0.7 argues that pixel-space WAMs are inefficient (e.g. VLAs can be up to 60x cheaper in compute) and imperfectly predicted pixels can lead to bad downstream actions.
The solution they argue is to combine VLAs and WAMs by bridging direct action prediction and world modeling through a shared latent space.
It is this efficiency in \textbf{\textit{reasoning}} that allows multiple queries of hypothetical reasoning which was previously considered unobtainable in traditional WAM settings.
% }

% \textcolor{blue}{
Many embodied world models suffer from inefficiencies in \textbf{\textit{reasoning}}, require expensive hardware for robust \textbf{\textit{perception}}, or rely on training data that is costly to create.
Efficiency in the Embodied setting arises in \textbf{\textit{memory}}-enabled efficient WAM's and hybrid VLA-WAM models.
The later innovate across all components of cognition, by introducing to VLAs a robust \textbf{\textit{memory}} mechanic for conditioning action selection on latent future world states, while maintaining efficiency enabling hypothetical reasoning. 
These hybrid VLA-WAMs are true Unified world models as depicted in Fig.~\ref{fig:catbrain} only lacking intrinsic motivation and metacognition.
Even more recently, World Action Models are showing future video prediction can be expendable achieving highly efficient runtime latencies~\cite{akbari2026flash,yuan2026fastwamworldactionmodels}.
However, Fast-WAM removes the capability for imagining future states in its model for further performance gains with little quality drop raising the question of the importance of \textbf{\textit{imagining}} in current embodied settings~\cite{yuan2026fastwamworldactionmodels}.
This achieves 190ms inference latency, or over $4\times$ faster than measured states of the art.
Flash-WAM proposes a step distillation framework deployed on Lingbot-VA reducing action denoising steps to two, and video denoising steps to one providing 348ms inference latency or $23\times$ speedup over Lingbot-VA during inference~\cite{akbari2026flash}.
Both works demonstrate the potential cost savings that can come from reducing reliance on computationally burdensome future state prediction.
% }
% \section{Human-AI Collaboration}
% \section{World Models in Scientific Discovery}
\section{Epistemic World Models}
\label{sec:collab}

\begin{figure*}[]
    \centering
    \includegraphics[width=\linewidth]{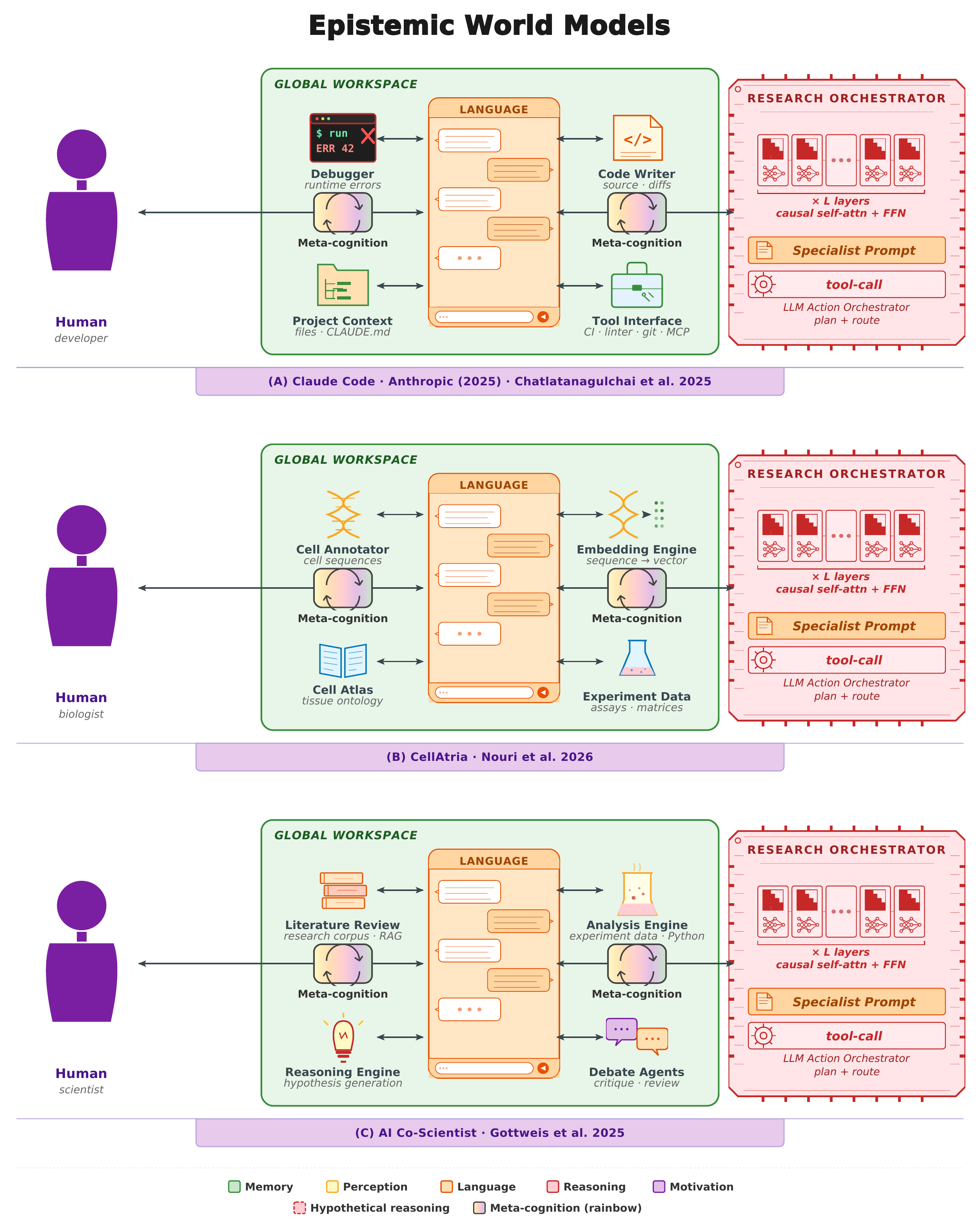}
    \vspace{-0.1cm}
    \hrule
    \vspace{-0.1cm}
    \caption{
   Above are the typical architecture and Global Workspace frameworks encountered when reviewing world models for scientific discovery used with a human-in-the-loop subject-matter expert.}
    \label{fig:collab_architectures}
\end{figure*}

World models in agentic frameworks extend beyond latent dynamics prediction to operate over the scientific process, orchestrating multi-step workflows that integrate perception, memory, reasoning, and tool use in service of discovery. In contrast to latent world models, which encode external environments into learned state representations with explicit transition dynamics, these systems treat structured knowledge as the world itself. In such \textit{epistemic world models}, the environment is a knowledge space defined by literature, databases, and experimental outputs, where domain expertise induces the state space and scientific artifacts act as observations. Rather than evolving an internal latent state, the agent updates this epistemic state through reasoning and tool-mediated operations, effectively controlling externalized cognitive functions.

This distinction highlights two complementary strategies for world modeling: latent compression of external environments versus explicit maintenance and updating of structured knowledge. Importantly, epistemic world models also instantiate a form of global workspace, where intermediate results, tool outputs, and shared context are broadcast across agents and processes. In the sense of Baars' Global Workspace Theory~\cite{baars1993cognitive}, these systems approximate metacognition by enabling self-reflection, self-evaluation, and self-control with selective routing of information across distributed components and even self-evaluation. As introduced in Sec.~\ref{sec:unified:cat}, such workspace-based mechanisms provide a candidate solution to the lack of metacognition in latent world models.
Agentic systems therefore emphasize reasoning, tool coordination, and persistent memory, while introducing early forms of metacognitive control through execution tracing and human-in-the-loop feedback. These capabilities suggest a concrete pathway toward addressing the gaps in \textbf{\textit{metacognition}} and \textbf{\textit{motivation}} outlined in Sec.~\ref{sec:unified:cat}, by externalizing and structuring the global workspace required for self-reflection, self-evaluation and self-control.

In this section, we review world models for Scientific Discovery through the lens of Cognitive Architecture Theory (CAT), emphasizing Human--AI Collaboration within multi-agent frameworks. As in previous sections, Sec.~\ref{sec6:video:representation} covers contributions to world model representation (memory, language, perception), while Sec.~\ref{sec6:video:generation} examines generation and prediction. 
The reviewed works are further subdivided according to the cognitive functions they emulate in their designs.
Figure~\ref{fig:collab_architectures}(a) illustrates common multi-agent architectures, with (b) highlighting Global Workspace-like structures for shared domain knowledge. Table~\ref{tab:human_ai_collab} summarizes all works within the CAT framework. Sec.~\ref{sec:collab:trends} summarizes key trends and limitations of epistemic world models.

% In this section, we survey world models used for Scientific Discovery through the lens of Cognitive Architecture Theory (CAT). 
% Here we will see the centrality of Human--AI Collaboration within these world model multi-agent frameworks.
% % We organize the literature into four application-driven subsections—(i) Trust, Human-alignment, and Interpretability, (ii) Software Co-pilots, (iii) Medical Research and Application, and (iv) Social Science. 
% Like in the previous sections, in Sec.~\ref{sec6:video:representation} we will survey works according to their contribution to world model representation (utilizing memory, language, or perception).
% Then in Sec.~\ref{sec6:video:generation} we will survey works according to their contribution to world model generation and prediction.
% Figure~\ref{fig:collab_architectures}(a) shows us the common multi-agent framework seen in works across the domains of scientific discovery.
% Sub-figure (b) shows examples of the Global Workspaces that multi-agent frameworks and humans represent theeir domain knowledge.
% Table~\ref{tab:human_ai_collab} summarizes all referenced works and provides a compact overview of how each paper contributes across these CAT functions.

\begin{table*}[]
\centering
\caption{Epistemic world model works in this report mapped to Cognitive Architecture Theory (CAT) functions and subtasks. In the \textbf{Subtask} column, \textbf{T} = Trust, Human-alignment, and Interpretability, \textbf{C} = Software Co-pilots, \textbf{M} = Medical Research and Application, and \textbf{S} = Social Science. Composite labels indicate that a work contributes to multiple Human--AI collaboration subtasks. A check (\cmark) indicates the work makes a clear contribution to that CAT function in the context of collaboration.}
\label{tab:human_ai_collab}
\scriptsize
\noindent\makebox[\textwidth]{%
\setlength{\tabcolsep}{2.8pt}
\renewcommand{\arraystretch}{1.05}
\begin{tabular}{p{2.10cm} c c c c c c c c c p{4.25cm}}
\toprule
\textbf{Work} & \textbf{Year} & \textbf{Subtask} & \rotatebox{90}{\textbf{Mem.}} & \rotatebox{90}{\textbf{Perc.}} & \rotatebox{90}{\textbf{Lang.}} & \rotatebox{90}{\textbf{Reas.}} & \rotatebox{90}{\textbf{Imag.}} & \rotatebox{90}{\textbf{Moti.}} & \rotatebox{90}{\textbf{Meta.}} & \textbf{Description} \\

\midrule
\multicolumn{11}{l}{\textit{Global Workspace: language, with human-in-the-loop}} \\
\midrule
Gemini Co-scientist~\cite{gottweis2025towards,gottweis2026accelerating} & 2025 & T+C+M & \cmark & \cmark & \cmark & \cmark & & \cmark & \cmark & multi-agent AI collaborator for hypothesis generation and refinement \\
OpenAI Prism~\cite{OpenAI_Prism_2026} & 2026 & T+C & \cmark & \cmark & \cmark & \cmark &  & \cmark & \cmark & LaTeX-native workspace for scientific writing and collaboration \\
SciSciGPT~\cite{shao2025sciscigpt} & 2025 & T+C+S & \cmark & \cmark & \cmark & \cmark &  & \cmark & \cmark & agentic science-of-science analytics over literature and structured datasets \\
OmniScientist~\cite{shao2025omnisci} & 2025 & T+C+S & \cmark & \cmark & \cmark & \cmark &  & \cmark & \cmark & human--AI scientist ecosystem with community evaluation (ScienceArena) \\
Mehri Shervedani et al.~\cite{Mehri2025multimodal} & 2025 & T+M+S & \cmark & \cmark & \cmark & \cmark & \cmark & \cmark &  & multimodal RL interaction manager for assistive human--robot collaboration \\

CellAtria~\cite{nouri2026agentic} & 2026 & M & \cmark & \cmark & \cmark & \cmark &  & \cmark & \cmark & Using AI for RNA sequencing and analysis\\

Gentile et al.~\cite{gentile2026artificial} & 2026 & M & \cmark & \cmark & \cmark & \cmark &  & \cmark & \cmark & AI Transcriptomics with human-in-the-loop for gene expression analysis \\

\midrule
\multicolumn{11}{l}{\textit{Global Workspace: language, without human-in-the-loop}} \\
\midrule

Generative Agents~\cite{park2023generative} & 2023 & T+S & \cmark & \cmark & \cmark & \cmark & &  & \cmark & LLM agents with memory, reflection, and planning for social simulation \\

SpeechAgents~\cite{zhang2024speechagent} & 2024 & S &  & \cmark & \cmark & \cmark &  &  &  & multi-agent spoken interaction controlled by a speech-centric LLM \\
Kumar et al.~\cite{KumarTanusri2025} & 2025 & T+S & \cmark & \cmark & \cmark & \cmark &  &  &  & unified speech-to-speech model claims for multilingual, emotional interaction \\

Kim et al.~\cite{Kim2025towards} & 2025 & S & \cmark & \cmark & \cmark &  & \cmark &  &  & multimodal conversational agent generating engaging speech from audio-visual cues \\
\midrule
\multicolumn{11}{l}{\textit{Reviews, Benchmarks, and Position Papers}} \\
\midrule

Xie et al.~\cite{xie2025farai} & 2025 & T+C+S &  &  & & \cmark &  &  & \cmark & Roadmap on AI scientists emphasizing verification and falsification \\
Agentic Coding Manifests~\cite{chatlatanagulchai2025use} & 2025 & T+C & &  & \cmark & \cmark &  &  & & Empirical study of Claude Code's repo manifests that externalize project context and rules for coding agents \\
Tsvetkova et al.~\cite{tsvetkova2024anew} & 2024 & T+S &  &  &  & \cmark &  & \cmark & \cmark & framework for sociology of mixed human--machine communities \\
Strachan et al.~\cite{strachan2024testing} & 2024 & T+S &  &  & \cmark & \cmark &  &  &  & empirical theory-of-mind benchmark comparing LLMs and humans \\
Chen et al.~\cite{chen2025exploring} & 2025 & T+S &  &  & \cmark & \cmark &  &  & \cmark & survey of consciousness/metacognition theories, implementations, and risks in LLMs \\
Kwok et al.~\cite{kwok2026explicit} & 2026 & T+S & \cmark & \cmark & \cmark & \cmark &  &  &  & explicit shared world models for reliable human--robot collaboration \\
Fung et al.~\cite{fung2025embodiedaiagentsmodeling} & 2025 & M+S & \cmark & \cmark & \cmark & \cmark &  & \cmark &  & position paper on embodied assistants with memory, world models, and goal inference \\

% Yura et al.~\cite{Yura2025multimodal} & 2025 & C+S &  & \cmark &  & \cmark &  &  &  & multimodal speaker detection and multi-step timing for service-robot interaction \\
% Qu et al.~\cite{qu2024robot} & 2024 & T+C+S & \cmark & \cmark & \cmark & \cmark &  &  &  & Pepper-based multimodal interaction platform integrating ChatGPT \\
% Chat with the Environment~\cite{zhao2023chatenv} & 2023 & T+C+S &  & \cmark & \cmark & \cmark &  &  & \cmark & LLM-driven active multimodal perception via epistemic actions \\
% MILD~\cite{Prasad2022MILD} & 2022 & S &  & \cmark &  &  & \cmark &  &  & latent dynamics model that predicts robot trajectories from human trajectories \\
% Voila~\cite{shi2025voila} & 2025 & C+S &  & \cmark & \cmark & \cmark & \cmark &  &  & real-time, full-duplex voice-language interaction and role-play \\

% MFCIG-CSS~\cite{jia2025multimodal} & 2025 & C+S & \cmark & \cmark & \cmark &  & \cmark &  &  & graph-based modeling of fine-grained semantic/prosody context for conversational TTS \\
% DiffCSS~\cite{wu2025diffcssd} & 2025 & C+S &  & \cmark & \cmark &  & \cmark &  &  & diffusion-based diverse prosody generation for conversational speech synthesis \\
% GPT-Talker~\cite{liu2024generative} & 2024 & C+S & \cmark & \cmark & \cmark &  & \cmark &  &  & expressive conversational speech synthesis conditioned on dialogue context \\

\bottomrule
\end{tabular}
}
\end{table*}

%%%%%%%%%%%%%%%%%%%%%%%%%%%%%%%%%%%%%%
\subsection{World Model Representation}\label{sec6:video:representation}
\subsubsection{\textbf{Language}.}
In human--AI collaboration systems, language is not only a medium for generation (e.g., code or hypotheses), but a substrate for shared world representation, coordination, and self-evaluation, consistent with notions of a global workspace~\cite{baars1993cognitive} as discussed in Sec.~\ref{sec:worldrep}. 
In co-science systems such as Gemini Co-scientist~\cite{gottweis2025towards,gottweis2026accelerating}, OpenAI Prism~\cite{OpenAI_Prism_2026}, SciSciGPT~\cite{shao2025sciscigpt}, and OmniScientist~\cite{shao2025omnisci}, language encodes hypotheses, project state, and intermediate reasoning, enabling iterative critique, revision, and verification across multi-agent interactions. Rather than serving only as input/output, language acts as a persistent interface through which agents construct, refine, and align their shared world model.

Reliable collaboration depends on maintaining explicit and evolving common ground rather than relying on opaque internal state~\cite{kwok2026explicit}. Language enables this by making agent intentions and reasoning processes interpretable and contestable~\cite{park2023generative}. This aligns with theories of human cognition in which language emerges from the capacity to share intentions, attention, and goals~\cite{tomasello2005understanding,deacon1998symbolic,jaynes2013origin}.
Viewed through this lens, \textbf{\textit{language}} functions as an integrative interface linking \textbf{\textit{perception}}, \textbf{\textit{reasoning}}, and \textbf{\textit{metacognition}}, rather than as a standalone token prediction mechanism. It is through language that internal representations become shared, inspected, and coordinated, enabling collaborative intelligence over a common world model.

\subsubsection{\textbf{\textit{Perception}}.}
Many software co-pilots extend perception beyond plain text, treating structured artifacts, speech, and embodied signals as first-class inputs~\cite{OpenAI_Prism_2026,shao2025sciscigpt,shao2025omnisci,KumarTanusri2025,zhang2024speechagent,Kim2025towards,fung2025embodiedaiagentsmodeling,Mehri2025multimodal}. OpenAI Prism~\cite{OpenAI_Prism_2026} exemplifies document-centric perception by operating directly over LaTeX structure (equations, references, figures), grounding edits in the manuscript’s semantics. SciSciGPT~\cite{shao2025sciscigpt} reframes perception as evidence acquisition, retrieving and parsing literature and structured datasets to support downstream reasoning. Similarly, OmniScientist~\cite{shao2025omnisci} organizes perception into structured representations of scientific knowledge. Across these systems, perception shifts from passive input processing to active construction of task-relevant representations.

In embodied and assistive settings, \textbf{\textit{perception}} becomes inherently multimodal and user-centric~\cite{Mehri2025multimodal,fung2025embodiedaiagentsmodeling}. Mehri Shervedani et al.~\cite{Mehri2025multimodal} integrate dialogue acts with multimodal signals to infer user intent in collaborative tasks, while Fung et al.~\cite{fung2025embodiedaiagentsmodeling} emphasize continuous, first-person sensing in wearable or embodied assistants. These approaches enable proactive assistance but introduce challenges in reliability, ambiguity resolution, and privacy, making perception a critical bottleneck for safe deployment.

In social interaction, \textbf{\textit{perception}} expands to include communicative signals such as prosody, emotion, and contextual cues~\cite{zhang2024speechagent,KumarTanusri2025,Kim2025towards,kwok2026explicit}. SpeechAgents~\cite{zhang2024speechagent} treat speech as a primary interaction channel, preserving rhythm and affect rather than reducing communication to text. Kim et al.~\cite{Kim2025towards} further condition interaction on audio-visual signals to improve engagement, while Kwok et al.~\cite{kwok2026explicit} highlight the importance of grounding ambiguous social cues in shared context for reliable collaboration. Other systems approximate perception through dialogue-history encodings or structured observations (e.g., pose or environment state), trading richness for tractability.

Across domains, improvements in trust are less about increasing perceptual bandwidth and more about structuring and grounding perceptual inputs. Systems that expose intermediate representations (e.g., retrieved evidence, structured documents, or shared context) make \textbf{\textit{perception}} more interpretable, whereas purely latent encodings of multimodal input remain difficult to audit. As a result, effective co-pilots treat perception not as raw sensing, but as the construction of verifiable, task-aligned state.

\subsubsection{\textbf{\textit{Memory}}.}

Many co-pilots improve memory by externalizing long-horizon context into persistent artifacts such as project state, corpora, or structured workspaces~\cite{OpenAI_Prism_2026,shao2025sciscigpt,shao2025omnisci,KumarTanusri2025,Kim2025towards,chatlatanagulchai2025use,fung2025embodiedaiagentsmodeling}. OpenAI Prism~\cite{OpenAI_Prism_2026} treats the LaTeX project itself as working memory, enabling consistent revision across documents, while OmniScientist~\cite{shao2025omnisci} extends memory into a research ecosystem via knowledge graphs and evolving ``idea stacks.'' In contrast, Agentic Coding Manifests~\cite{chatlatanagulchai2025use} provide a lightweight approach, storing repository conventions and constraints as human-authored memory. Across these systems, a central trade-off emerges between simple explicit memory (manifests), structured external memory (graphs, databases), and conversational state.

In medical research settings, explicit long-term \textbf{\textit{memory}} remains underdeveloped despite its importance. 
Two works in particular however, provide novel world encoding strategies for their RNA transcription setting~\cite{nouri2026agentic,gentile2026artificial}.
In both, using agentic reasoning for RNA sequence analysis requires representing sequence annotations or researcher meta-data as written language.
Also in both works, they can utilize the sequence of RNA (itself semantically and symbolically salient) as embeddings that LLMs can use to reason with the RNA-text-like representation itself.
This representation of RNA for a global workspace framework is observable in Figure~\ref{fig:collab_architectures}.

Social and interactive systems place stronger demands on \textbf{\textit{memory}} for continuity and shared understanding~\cite{KumarTanusri2025,fung2025embodiedaiagentsmodeling,kwok2026explicit,park2023generative,Mehri2025multimodal}. Generative Agents~\cite{park2023generative} exemplify this by storing episodic experiences and synthesizing higher-level reflections that guide future behavior. Similarly, Kwok et al.~\cite{kwok2026explicit} and Mehri Shervedani et al.~\cite{Mehri2025multimodal} model memory as shared state (common ground) that must be continuously updated during interaction. In speech systems, longer-term personalization is often approximated through dialogue history or user-specific embeddings~\cite{KumarTanusri2025}, while embodied assistants require episodic memory to sustain coherent assistance over time~\cite{fung2025embodiedaiagentsmodeling}. A key distinction is between interpretable shared-memory representations (e.g., common ground) and implicit history encodings that are harder to audit.

Across domains, improvements in trust are closely tied to making \textbf{\textit{memory}} explicit, structured, and inspectable~\cite{gottweis2025towards,gottweis2026accelerating,OpenAI_Prism_2026,chatlatanagulchai2025use,kwok2026explicit,park2023generative,Mehri2025multimodal,shao2025omnisci}. Generative Agents~\cite{park2023generative} provide a clear example by exposing both raw episodic memories and derived reflections, enabling users to trace behavior back to stored experience. Similarly, manifests~\cite{chatlatanagulchai2025use} and shared-state representations~\cite{kwok2026explicit} externalize assumptions and context into artifacts that can be inspected and revised. Larger systems such as Prism~\cite{OpenAI_Prism_2026} and OmniScientist~\cite{shao2025omnisci} extend this idea to document- and ecosystem-level memory. Overall, the dominant pattern is a shift away from opaque latent state toward persistent artifacts that users can inspect, version, and contest.

\subsection{World Model Prediction and Generation}\label{sec6:video:generation}

\subsubsection{\textbf{\textit{Imagination}}.}
Co-pilots employ imagination to generate candidate hypotheses, plans, or expressive outputs that a human or downstream process can select, refine, or test~\cite{gottweis2025towards,gottweis2026accelerating,shao2025omnisci,zhang2024speechagent,Kim2025towards,Mehri2025multimodal}. In scientific collaboration, Gemini Co-scientist~\cite{gottweis2025towards,gottweis2026accelerating} produces research hypotheses and experimental proposals, while OmniScientist~\cite{shao2025omnisci} treats ideation as a first-class module, evolving candidate ideas within a structured knowledge graph. In these settings, imagination functions as controlled hypothesis expansion rather than unconstrained generation.

In medical and scientific domains, \textbf{\textit{imagination}} most directly appears as hypothesis generation under experimental constraints~\cite{gottweis2025towards,gottweis2026accelerating}. Systems such as Gemini Co-scientist propose candidate mechanisms, repurposing strategies subsequently filtered by feasibility and empirical validation, reflecting a generate-then-triage workflow aligned with the scientific method.

In interactive and social systems, \textbf{\textit{imagination}} often takes the form of one-to-many generation. For example, Kim et al.~\cite{Kim2025towards} generate expressive paralinguistic speech, while Mehri Shervedani et al.~\cite{Mehri2025multimodal} employ simulated rollouts via user models to support policy learning. Generative Agents~\cite{park2023generative} extend this paradigm to multi-agent settings, where imagined actions at the individual level produce emergent group behaviors over time. Across these domains, imagination supports diverse objectives, including expressive communication, social simulation, and action planning.

\textbf{\textit{Imagination}} is not used in isolation. Its utility depends on coupling downstream selection mechanisms, such as ranking, constraint satisfaction, or evaluation. In practice, trust is not derived from the generative step itself, but from the processes that filter, verify, and prioritize candidates.

\subsubsection{\textbf{\textit{Reasoning}}.}
Reasoning in world models and co-pilots is typically framed as multi-step problem solving under constraints, and mediated by tool use, workflows, or multi-agent protocols \cite{OpenAI_Prism_2026,shao2025sciscigpt,gottweis2025towards,gottweis2026accelerating,shao2025omnisci,xie2025farai,KumarTanusri2025,chatlatanagulchai2025use,fung2025embodiedaiagentsmodeling,Mehri2025multimodal}. 
Systems span a spectrum from structured, tool-grounded reasoning to deliberative, multi-agent reasoning. OpenAI Prism~\cite{OpenAI_Prism_2026} constrains reasoning over structured artifacts (e.g., LaTeX projects), while SciSciGPT~\cite{shao2025sciscigpt} operationalizes reasoning as an end-to-end empirical pipeline (decomposition, retrieval, computation, and visualization). 
In contrast, Gemini Co-scientist~\cite{gottweis2025towards,gottweis2026accelerating} and OmniScientist~\cite{shao2025omnisci} deliberates through multi-agent protocols such as debate, reflection, and workflow orchestration. 
At the policy level, Gemini Co-scientist~\cite{gottweis2025towards,gottweis2026accelerating} use agents to recognize novel research trajectories by debating hypotheses, ranking ideas with an Elo-style system, filtering out redundant ideas, while Mehri Shervedani et al.~\cite{Mehri2025multimodal} frame reasoning as action selection under uncertainty, learning dialogue and intervention policies for assistive tasks.

In medical and assistive settings, \textbf{\textit{reasoning}} bifurcates into \textit{scientific reasoning} and \textit{decision-making under uncertainty}. Gemini Co-scientist~\cite{gottweis2025towards,gottweis2026accelerating} focuses on evidence-backed hypothesis generation and refinement via multi-agent deliberation, while Mehri Shervedani et al.~\cite{Mehri2025multimodal} optimize real-time interaction policies for successful task completion. This contrast highlights reasoning as either hypothesis validation or policy optimization.
In RNA transcription medical research, models reason from representations of RNA to profile genetic sequences for personalized medicine~\cite{nouri2026agentic,gentile2026artificial}.

In social and collaborative contexts, \textbf{\textit{reasoning}} extends beyond individual cognition to include theory-of-mind inference, experimental methodology, and system-level dynamics \cite{shao2025sciscigpt,fung2025embodiedaiagentsmodeling,kwok2026explicit,park2023generative,Mehri2025multimodal,tsvetkova2024anew,strachan2024testing,chen2025exploring,xie2025farai}. 
SciSciGPT~\cite{shao2025sciscigpt} exemplifies reasoning as transparent scientific workflows, while Kwok et al.~\cite{kwok2026explicit} emphasize explicit shared models for reasoning about human goals. Strachan et al.~\cite{strachan2024testing} evaluates theory-of-mind capabilities, exposing systematic failure modes in social reasoning tasks. At a broader scale, Tsvetkova et al.~\cite{tsvetkova2024anew} model mixed human--machine systems as dynamical processes, and Xie et al.~\cite{xie2025farai} identifies verification and falsification as central bottlenecks for scalable scientific reasoning. Together, these works position reasoning as both an individual capability and an emergent property of socio-technical systems.

Across these domains, \textbf{\textit{reasoning}} is tightly coupled to trust through auditability and verification \cite{gottweis2025towards,gottweis2026accelerating,OpenAI_Prism_2026,shao2025sciscigpt,kwok2026explicit,Mehri2025multimodal,tsvetkova2024anew,strachan2024testing,chen2025exploring,xie2025farai,shao2025omnisci,KumarTanusri2025}. 
Multi-agent deliberation surfaces alternatives and justifications (e.g., Gemini Co-scientist), while procedural pipelines enable reproducibility and inspection (e.g., SciSciGPT). 
Evaluation frameworks, such as those by Strachan et al.~\cite{strachan2024testing}, test whether reasoning capabilities generalize across conditions. More broadly, Xie et al.~\cite{xie2025farai} argues that trustworthiness requires explicit verification loops, and Tsvetkova et al.~\cite{tsvetkova2024anew} situates trust within ecosystem-level dynamics. 
Reasoning in current systems spans deliberative (multi-agent critique), procedural (tool-grounded workflows), and evaluative (verification and benchmarking) paradigms, which together form the basis for reliable and interpretable decision-making.

\subsubsection{\textbf{\textit{Motivation}}.}
Motivation remains underdeveloped in epistemic world models, with most works relying on externally specified objectives rather than intrinsic drives. Existing work primarily operationalizes motivation through rewards, goals, or intent inference \cite{fung2025embodiedaiagentsmodeling,Mehri2025multimodal,tsvetkova2024anew,xie2025farai,shao2025omnisci}. 

A key distinction emerges between \textit{optimization-driven} and \textit{inference-driven} \textbf{\textit{motivation}}. In RL settings, motivation is encoded implicitly as an optimization signal via reward shaping and penalties, as in assistive human--robot interaction systems that incentivize efficient and valid behavior~\cite{Mehri2025multimodal}. In contrast, embodied assistants increasingly model motivation as the inference of user goals or intent, enabling proactive assistance without explicit commands~\cite{fung2025embodiedaiagentsmodeling}. This reframes motivation as alignment to latent human objectives rather than adherence to predefined reward functions.

As discussed in Sec.~\ref{sec:unified:cat}, in practice, motivation is predominantly \textit{extrinsic} and often safety- or task-driven, particularly in medical and assistive domains~\cite{Mehri2025multimodal,fung2025embodiedaiagentsmodeling}. Systems are designed to optimize externally defined criteria such as correctness, efficiency, or user satisfaction, with little evidence of intrinsic or self-directed objective formation.
At a broader scale, motivation is shaped by the surrounding socio-technical system. Incentives, credit assignment, and selection pressures govern agent behavior in collaborative and scientific settings \cite{tsvetkova2024anew,xie2025farai,shao2025omnisci}. Tsvetkova et al.~\cite{tsvetkova2024anew} demonstrate how incentive structures drive emergent phenomena such as cascades and manipulation, while OmniScientist~\cite{shao2025omnisci} embeds motivation in mechanisms such as attribution and peer review to regulate collaboration quality. Xie et al.~\cite{xie2025farai} further argues that automation reshapes which problems are pursued, effectively altering the motivational landscape of the research ecosystem.
Motivation in current world models is not intrinsic but arises from externally imposed objectives, highlighting a key gap between machine systems and human-like cognition.

\subsubsection{\textbf{\textit{Metacognition.}}}
Metacognition is introduced as a reliability scaffold in epistemic world models, through self-reflection, self-evaluation, uncertainty management, and error checking \cite{OpenAI_Prism_2026,shao2025sciscigpt,gottweis2025towards,gottweis2026accelerating,shao2025omnisci,xie2025farai,chatlatanagulchai2025use,park2023generative,tsvetkova2024anew,chen2025exploring}. 
Many designs separate generation from judgment, forming propose--evaluate loops. OpenAI Prism~\cite{OpenAI_Prism_2026} introduces an always-on reviewer layer for proofreading and consistency checking, while Gemini Co-scientist~\cite{gottweis2025towards,gottweis2026accelerating} and OmniScientist~\cite{shao2025omnisci} operationalize critique through dedicated self-reflection roles and ranking mechanisms. 
SciSciGPT~\cite{shao2025sciscigpt} further emphasizes reproducibility and output validation within empirical pipelines.

Metacognitive mechanisms operate at two levels. At the \textit{individual level}, systems employ internal self-reflection to refine outputs or update beliefs, as seen in Generative Agents~\cite{park2023generative}, where reflection transforms experience into higher-level behavioral. At the \textit{system or community level}, metacognition is externalized through evaluation platforms, auditability, and feedback loops, enabling reproducibility and collective validation~\cite{shao2025sciscigpt,shao2025omnisci}. 
This distinction highlights a key design axis: internal self-reflection, self-evaluation, and self-control versus external governance.

These safeguards are particularly critical in high-stakes settings, where unchecked generation can propagate errors or overconfident hypotheses. For instance, Gemini Co-scientist~\cite{gottweis2025towards,gottweis2026accelerating} uses critique and reflection loops to down-select hypotheses prior to wet-lab validation, reducing experimental cost and risk. More broadly, Chen et al.~\cite{chen2025exploring} and Xie et al.~\cite{xie2025farai} argue that without explicit self-evaluation and external oversight, iterative self-improvement can amplify errors. 
Overall, effective metacognition in world models emerges from coupling strong generative capabilities with explicit, testable judgment layers, often combining internal self-reflection, self-evaluation, and self-control with external evaluation and governance.

\subsection{Trends and Limitations}
\label{sec:collab:trends}
% \textcolor{blue}{
% Amir(@Tim, please review):\\
Epistemic world models are emerging as a shift from implicit latent-state modeling toward explicit, externalized knowledge-state modeling. Rather than compressing observations into an opaque latent state, these systems maintain the ``world'' as a shared epistemic \textbf{\textit{language}}-based global workspace consisting of literature, code, documents, tool outputs, retrieved evidence, hypotheses, experimental data, and human feedback. This makes scientific discovery more interpretable than in many video or embodied world models. A recurring trend is therefore the use of global workspaces in which agents coordinate tool use, propose hypotheses, revise artifacts, and expose partial reasoning to humans~\cite{gottweis2025towards,gottweis2026accelerating,OpenAI_Prism_2026,shao2025sciscigpt,shao2025omnisci,chatlatanagulchai2025use}. In this sense, epistemic world models provide one of the clearest current pathways toward metacognitive scaffolding: generation is separated from judgment through reviewer agents, critique loops, debate, ranking, execution traces, and human oversight.
% }

% \textcolor{blue}{
However, these systems remain limited in ways that are easy to obscure behind fluent language and multi-agent workflows. First, their epistemic state transitions are rarely formalized as clearly as latent world models; the state is often distributed across prompts, documents, databases, retrieved passages, logs, and human edits, making reproducibility and causal attribution difficult. Second, motivation remains mostly extrinsic, inherited from user instructions, benchmarks, reward functions, institutional incentives, or peer-review mechanisms rather than intrinsic objective formation. Third, metacognition is still scaffolded rather than autonomous: self-evaluation usually depends on additional agents, external tools, or human validators that may share the same blind spots as the generator. As a result, the central challenge for epistemic world models is not merely adding more agents, but developing reliable mechanisms for provenance tracking, uncertainty-aware verification, and accountable human--AI governance over the evolving knowledge state~\cite{xie2025farai,tsvetkova2024anew,chen2025exploring}.
% }
\section{Conclusion}
% In our review of video, embodied, and newly named epistemic world models, 
We identify in video, embodied, and newly named epistemic world model literature trends for overcoming common problems with common solutions.
We are the first to propose a taxonomy of recent world models rooted in cognitive architecture theory.
% We synthesize our review of the literature into a Unified Framework for world models and provides a conceptual road map that calls on researchers to leverage multi-modal perception, and to design efficient and effective latent spaces for downstream reasoning tasks, including hypothetical reasoning, or imagination.
Our taxonomy reveals a research gap regarding the cognitive functions of motivation (especially intrinsic), and metacognition that may not have been identified otherwise.
Additionally, our taxonomy inspires us to rethink agent frameworks for scientific discovery as epistemic world models with strong metacognitive capabilities from language-based Global Workspaces. 
Adding a human in the loop to promptable video, embodied and epistemic world models, for them to lend their own motivation and metacognitive ability, remains one of the clearest observances of human cognition in machines. 

In answer to what a \textit{human}-like world model would look like, we propose a unified world model framework that holistically incorporates all of the component parts of cognition: memory, perception, language, reasoning, imagining, motivation, and metacognition.
This framework acts as a conceptual roadmap for researchers.
Unified world models encourage researchers to 1) use multi-modal inputs,  2) encode multi-scalar spatio-temporal representations, 3) include tokenized language as an input, intermediate reasoning space, or output to enable human-in-the-loop cooperation, 4) utilize sim-to-real training when data is scarce and hypothetical reasoning at inference, 5) reason with domain-specific state-of-the-art architectures, 6) provide intrinsic reward signals to world models and lastly 7) utilize a global workspace enabling self-reflection, self-evaluation, and self-control. 
Our report provides researchers a vocabulary to debate the merits of equating human and machine cognition.
Our review shows that to begin to fully equate machine and human cognition in function, let alone in any philosophical level, requires a world model that holistically emulates all component parts of cognition, including the under-researched cognitive functions of motivation, and metacognition.

\bibliographystyle{ACM-Reference-Format}
\bibliography{manuscript}

% Representations and Generation

% [x] Move common problems, and unified framework to 3, 4
% Add metrics section (see survey from enfu)
% [x] long context window in LLM (too broad), focus on world model
% 3d space generation, sora 3d video for video, e.g. parabolic views
% Robotics, emphasize wan and ling bot and dream zero use in robotics. Compare to cosmos policy.

% \subsubsection{Acknowledgments} 

%
% ---- Bibliography ----
%
% BibTeX users should specify bibliography style 'splncs04'.
% References will then be sorted and formatted in the correct style.
%
% \bibliographystyle{splncs04}
% \bibliography{mybibliography}

\end{document}